\documentclass[twoside,11pt]{article}
\usepackage[utf8]{inputenc}
\usepackage[T1]{fontenc}
\usepackage{jair}
\usepackage{theapa}
\usepackage{rawfonts}
\usepackage[usenames, dvipsnames]{xcolor}
\usepackage{amsmath,amssymb,amsthm}
\usepackage[noend]{algpseudocode}
\usepackage{algorithm}
\usepackage{fancyvrb,mdframed}
\usepackage{graphicx,overpic,tikz}
\usepackage{float,multirow,makecell,booktabs,tabularx,subfloat}
\usepackage[caption=false]{subfig}
\usepackage{import}

\makeatletter
\let\OldStatex\Statex
\renewcommand{\Statex}[1][3]{%
  \setlength\@tempdima{\algorithmicindent}%
  \OldStatex\hskip\dimexpr#1\@tempdima\relax}
\makeatother

\usepackage[capitalise,noabbrev]{cleveref}

\newcommand{\hidact}[2]{\phi^{{#2}}_{{#1}}}
\newcommand{\hidprop}[2]{{\psi^{{#2}}_{{#1}}}}
\newcommand{\inact}[2]{u^{{#2}}_{{#1}}}
\newcommand{\inprop}[2]{v^{{#2}}_{{#1}}}
\DeclareMathOperator{\getpred}{pred}
\DeclareMathOperator{\getop}{op}
\DeclareMathOperator{\pre}{pre}
\DeclareMathOperator{\eff}{eff}
\DeclareMathOperator{\add}{add}
\DeclareMathOperator{\del}{del}

\DeclareMathOperator{\apply}{apply}

\newcommand{\m}[1]{\ensuremath{\mathcal{#1}}}

\newcommand{\etal}{\textit{et~al.}}
\DeclareMathOperator{\pool}{pool}

\DeclareMathOperator*{\argmax}{arg\,max}

\newcommand{\astar}{A$^\star$}

\makeatletter
\DeclareRobustCommand{\varname}[1]{\begingroup\newmcodes@\mathit{#1}\endgroup}
\makeatother

\newcommand{\mobj}[1]{\varname{#1}}
\newcommand{\mpred}[1]{\text{#1}}
\newcommand{\maschema}[1]{\text{#1}}
\makeatletter
\newcommand{\mobjlist}[1]{%
  \def\nextitem{\def\nextitem{, }}%
  \@for \el:=\mobj{#1}\do{\nextitem\el}%
}
\makeatother
\newcommand{\mactn}[2]{\maschema{#1}(\mobjlist{#2})}
\newcommand{\mpropn}[2]{\mpred{#1}(\mobjlist{#2})}

\newcommand{\wskip}{\phantom{z}}

\newcommand{\Omit}[1]{}

\makeatletter
\newcommand\smallscript{\@setfontsize\smallscript\@viipt\@viiipt}
\makeatother

\input{./code-samples/style.tex}

\jairheading{68}{2020}{1-68}{08/2019}{05/2020}
\ShortHeadings{ASNets: Deep Learning for Generalised Planning}
{Toyer, Thi\'ebaux, Trevizan, \& Xie}
\firstpageno{01}

\begin{document}

\title{ASNets: Deep Learning for Generalised Planning}
\author{\name Sam Toyer \email sdt@berkeley.edu\\
        \addr Department of Electrical Engineering and Computer Sciences\\
        \addr University of California, Berkeley\\
        \addr 2121 Berkeley Way, Berkeley CA 94720-1660, USA
        \AND
        \name Sylvie Thi\'ebaux \email sylvie.thiebaux@anu.edu.au\\
        \name Felipe Trevizan \email felipe.trevizan@anu.edu.au\\
        \name Lexing Xie \email lexing.xie@anu.edu.au\\
        \addr College of Engineering and Computer Science\\
        \addr The Australian National University\\
        \addr 145 Science Road, Canberra ACT 2601, Australia}

\maketitle

\begin{abstract}
  In this paper, we discuss the learning of generalised policies for
  probabilistic and classical planning problems using Action Schema Networks
  (ASNets).
  The ASNet is a neural network architecture that exploits the relational
  structure of (P)PDDL planning problems to learn a common set of weights that
  can be applied to any problem in a domain.
  By mimicking the actions chosen by a traditional, non-learning planner on a
  handful of small problems in a domain, ASNets are able to learn a generalised
  reactive policy that can quickly solve much larger instances from the domain.
  This work extends the ASNet architecture to make it more expressive,
  while still remaining invariant to a range of symmetries that exist in PPDDL
  problems.
  We also present a thorough experimental evaluation of ASNets, including a
  comparison with heuristic search planners on seven probabilistic and
  deterministic domains, an extended evaluation on over 18,000 Blocksworld
  instances, and an ablation study.
  Finally, we show that sparsity-inducing regularisation can produce ASNets that
  are compact enough for humans to understand, yielding insights into how the
  structure of ASNets allows them to generalise across a domain.
\end{abstract}

\section{Introduction}
\label{sec:intro}

Learning and planning are both important ingredients for constructing intelligent agents.
Planning can help an agent choose actions which will achieve its long-term goals
by reasoning about future trajectories, and learning can assist the agent in
using prior experience to more efficiently achieve new goals.
However, the prevalent methods for automated planning in the AI literature make
limited use of learning.
Some planning methods like RTDP are said to ``learn'' in the sense that they use
an iterative algorithm to come to successively better estimates of the value of
a certain state in a problem~\cite{barto1995learning}, but cannot transfer that
learnt knowledge to other problem instances.
Other planners can be used in conjunction with autoselectors and
autoconfigurators that predict which combination of planner and planner
configuration might work best for a given problem instance, based on features of
that instance.
These sorts of learning-based portfolio planners commonly feature in the
International Planning Competition (IPC), as they are able to combine the
disjoint strengths of many algorithms~\cite{coles2012survey,vallati2015ipc}.
However, while an autoselector and autoconfigurator may be able to transfer
knowledge across many instances, they are typically limited to choosing between
a handful of planners or changing only a few planner settings.
Despite some work over the past two decades and a recent uptick in
interest~\cite{martin2000learning,yoon2002inductive,de2011scaling,bonet2018features,bonet2019learning},
how best to learn and transfer deeper forms of knowledge across instances---such
as partial solutions, or knowledge of dead ends, or heuristics---is still an
open problem.

Separate to developments in planning, the broader AI community has recently seen
a resurgence of interest in neural networks.
This interest has been driven by the success of deep learning in tackling
problems ranging from image classification~\cite{krizhevsky2012imagenet} to
video game playing~\cite{mnih2013playing} and machine
translation~\cite{wu2016google}.
\citeA{lecun2015deep} argue that deep learning has met with greater success than
other machine learning techniques in these domains due to its ability to
automatically extract structure from high-level data, thereby obviating the need
for laborious feature engineering.
\citeauthor{lecun2015deep} also stress the importance of having appropriate deep
learning ``architectures'' for processing different modalities of input.
For instance, Convolutional Neural Networks (CNNs) are particularly well-suited to
processing image data, since they naturally capture notions like
translation invariance and hierarchical composition of features, and can be
efficiently applied to images of arbitrary size at test time~\cite{long2015fully}.
Similarly, the ability of bidirectional Recurrent Neural Networks (RNNs) to
(in principle) capture long-range dependencies in sequential data of arbitrary
length makes them a natural choice for text processing tasks.
However, there is not yet a standard neural network architecture that can do for
planning problems what CNNs do for images, or what RNNs do for text.
The absence of appropriate architectures is a barrier that must be overcome
before we see greater adoption of deep learning in automated planning.

Action Schema Networks (ASNets) are one of the first attempts to bridge the
worlds of automated planning and deep learning~\cite{toyer2018action}.
ASNets generalise the notion of a ``convolution'' to match the relational
structure of factored planning problems.
Where 2D CNNs operate on regular grids of features corresponding to locations of
pixels in an image, ASNets instead operate on an abstract graph of features
corresponding to actions and propositions from a planning problem.
The connections between actions and propositions in this graph are derived from
the action schemas for the corresponding planning domain.
This scheme makes it possible to share weights between policy
networks instantiated for different problems in a domain.
Hence, a set of small problems from a given domain can be used to learn a single
set of parameters which can be transferred to all other problems in that domain.
In other words, an appropriately-learnt set of weights can be used to obtain a
generalised policy.
\citeA{geffner2018model} observes that this kind of generalisation is not
possible with fully connected neural networks, which need to have fixed input
and output sizes throughout training and evaluation.
In a sense, learning generalised policies with ASNets represents a much tighter
integration of machine learning with automated planning than other strategies
like autoconfiguration, which only learn to tweak a handful of parameters for a
hand-coded planning algorithm.
Further, the flexible structure and generalisation capacity of ASNets could make
them a suitable tool for other tasks beyond learning generalised policies, such
as guiding tree search~\cite{shen2019guiding} or learning generalised
heuristics~\cite{shen2020learning}.

This paper expands upon the original ASNets paper~\cite{toyer2018action} in several ways.
\cref{sec:asnets} extends the original architecture with a more expressive pooling mechanism, as well as \textit{skip connections} between modules of the same type in different layers.
In \cref{sec:expts}, we perform a more thorough evaluation of ASNets across seven probabilistic and deterministic tasks.
The expanded evaluation includes four new tasks, an extended evaluation on 18,300 Blocksworld instances, and an ablation study identifying which ASNet features are most important for obtaining high coverage on our test domains.
In \cref{sec:understand}, we present a method for interpreting ASNet policies, and apply this to a policy for the Triangle Tireworld domain to better understand the mechanism that allows ASNets to generalise.
Finally, in \cref{sec:related}, we connect this work to the large body of relevant literature on deep learning and automated planning.

\section{Background}\label{sec:background}

This paper considers the task of solving Stochastic Shortest Path Problems
(SSPs)~\cite{bertsekas1996neurodynamic}.
An SSP can be represented by a tuple $(\mathcal S, \mathcal A, \mathcal T,
\mathcal C, \mathcal G, s_0)$ consisting of finite sets of states $\mathcal S$
and actions $\mathcal A$, a transition probability distribution $\mathcal T : \mathcal S
\times \mathcal S \times \mathcal A \to [0, 1]$, a cost function $\mathcal C :
\mathcal S \times \mathcal A \to \mathbb (0, \infty)$, a set of goal states
$\mathcal G \subseteq \mathcal S$, and an initial state $s_0 \in
\mathcal S$.
An agent following a policy $\pi : \mathcal A \times \mathcal S \to [0, 1]$ in
an SSP will start in state $s_0$, then repeatedly choose an action $a \sim
\pi(a|s)$ and execute it to reach a new state $s' \sim \mathcal T(s'|s, a)$
while incurring cost $\mathcal C(s,a)$ along the way.
A policy $\pi^*(a|s)$ is an optimal solution to an SSP if it reaches a goal
state $s_g \in \mathcal G$ with probability 1 while minimising total expected
cost.
In order to consider problems with unavoidable dead ends, we relax the
requirement that policies should reach the goal with probability 1.
Instead, policies are permitted to enter dead-end states, but at the cost of
incurring a large (but finite) dead end penalty~\cite{mausam2012planning}.
Note that the classical (deterministic) planning setting can be viewed as a
special case of this general SSP setting in which the transition probability
distribution $\mathcal T(s'|s,a)$ is deterministic.

Large SSPs are typically specified using a \textit{factored} representation
$(\mathcal P, \mathcal A, s_0, s_\star, \mathcal C)$.
$\mathcal P$ is a finite set of propositions (binary variables).
Each state $s \subseteq \mathcal P$ corresponds to the set of propositions that
are true, with the remaining propositions taken to be false.
The full state space is thus of size $|\mathcal S| = 2^{|\mathcal P|}$.
$s_0 \subseteq \mathcal P$ represents the initial state.
$s_\star \subseteq \mathcal P$ is a subset of propositions that defines the set of goal states $\mathcal G = \{s \in \mathcal S | s_\star \subseteq s\}$ in which these propositions are all true and the remaining propositions can be either true or false.
Each action $a \in \mathcal A$ consists of a precondition $\pre_a$ and a
distribution $\Pr_a$ over a set of deterministic effects $\eff_a$.
The precondition $\pre_a \subseteq \mathcal P$ represents the conditions that must
be satisfied before applying $a$---$a$ can only be applied in a state $s$ such that
$\pre_a \subseteq s$.
Thus, the set of actions {\em applicable} in a state $s$ is $\mathcal A(s) = \{a \in
\mathcal A | \pre_a \subseteq s\}$.
On applying $a \in \mathcal A(s)$, a deterministic effect $e \in \eff_a$ is
sampled from $\Pr_a$.
Each deterministic effect $e$ consists of a set of add-effects $\add(e)\subseteq P$ and a set of delete-effects $\del(e) \subseteq P$; applying $e$ in state $s$
yields a new state $s' = \apply(e,s) = (s\setminus \del(e)) \cup \add(e)$ in which the truth values of
some propositions have changed.
%
%
$\mathcal A$ thus gives rise to a transition probability distribution $\mathcal
T(s'|s,a) = \sum_{e \in \eff_a | s'=\apply(e,s)} \Pr_a(e)$, where $\mathcal
T(s'|s,a) = 0$ when $a \notin \mathcal A(s)$.
Finally, the semantics of the cost function $\mathcal C(s,a) \in (0, \infty)$
and initial state $s_0$ are unchanged from their definitions above.

Formally, we view a \textit{generalised} policy (of the sort learnt by an ASNet)
as a solution to the family of all factored SSPs that can be instantiated from a
certain \textit{lifted SSP}.
A lifted SSP is a tuple $(\mathbb P, \mathbb A, \mathcal C)$ consisting of a
set of predicates $\mathbb P$, a set of action schemas $\mathbb A$, and a cost
function $\mathcal C$.
A predicate can be viewed as a function that produces a concrete proposition
from a tuple of \textit{objects}, which are names that represent entities in an
environment.
Repeating this \textit{grounding} process for each predicate and each applicable
tuple of objects in an object set $\mathcal O$ can thus yield a set of
propositions $\mathcal P$.
For instance, say we are given predicates $\mathbb P = \{\mpred{at}(\mobj{?robot},
\mobj{?place})\}$ and objects $\mathcal O = \{\mobj{shakey}, \mobj{hall},
\mobj{kitchen}\}$.
After grounding, we would end up with a set of propositions $\mathcal P = \{
\mpred{at}(\mobj{shakey}, \mobj{kitchen}), \mpred{at}(\mobj{shakey},
\mobj{hall}), \ldots\}$ which we could use to represent the possible locations
of a robot named $\mobj{shakey}$.
An action schema can likewise be interpreted as a function that maps a tuple of
objects to a concrete action.
%
As an example, we could take a schema $\maschema{drive}(\mobj{?robot},
\mobj{?from}, \mobj{?to})$ and a tuple of objects $(\mobj{shakey},
\mobj{kitchen}, \mobj{hall})$ to instantiate a concrete action
$\maschema{drive}(\mobj{shakey}, \mobj{kitchen}, \mobj{hall})$ which moves
$\mobj{shakey}$ from the $\mobj{kitchen}$ to the $\mobj{hall}$.
In this way, a complete factored SSP can be instantiated from a lifted SSP
$(\mathbb P, \mathbb A, \mathcal C)$, a set of objects $\mathcal O$, an initial
state $s_0$, and a partial state $s_\star$ specifying the goal.
As we will see in later sections, factored SSPs that have been instantiated from
a shared set of predicates and action schemas typically have a similar structure
that can be exploited to learn compact generalised policies.

Lifted SSPs and factored SSPs are often specified using \textit{domain} and \textit{problem} definitions written in the Probabilistic Planning Domain Definition Language (PPDDL) \cite{younes2004ppddl1}.
A PPDDL domain defines a lifted SSP, and a PPDDL problem can be combined with the action schemas and predicates from a domain to specify a factored SSP.
\cref{fig:ppddl-rel-ex} shows an example of such a problem and its corresponding domain.

In addition to the constructs that we mentioned when introducing factored SSPs, PPDDL also supports more complex language features.
Some of those features are supported by the ASNet architecture without any kind of additional compilation or reduction.
Such features include arbitrarily nested conditional and probabilistic effects; nested precondition formulae featuring disjunction, negation, etc.; and stochastic initial state distributions.
As will become clear in \cref{sec:asnets} and \cref{sec:train-exploit}, the precise semantics of the preconditions and effects are only relevant for generating training data.
In contrast, when constructing the network, all that matters is where each proposition appears in the precondition or effect of each action (if at all).
We use this information to structure the network in such a way that it can generalise to different problems from the same domain, but the semantics of preconditions, effects, propositions, and so on are otherwise irrelevant to the network architecture.
For clarity of exposition we will therefore ignore these additional language features, and instead pretend that all PPDDL problems and domains are given in the STRIPS-like form introduced earlier.

Although ASNets support most PPDDL features, there are four PPDDL constructs that ASNets do not yet support: numeric variables, rewards, quantifiers, and arbitrary goal formulae.
The lack of support for numeric variables is simply an implementation omission: neural networks are capable of taking scalar inputs, so in principle these could be handled in the same way that propositions are handled currently.
However, we deemed a full treatment of numeric problems to be beyond the scope of this paper.
Likewise, ASNets do not yet support PPDDL rewards, but adding support would be straightforward given support for numeric variables and an appropriate teacher planner for generating training data.
In contrast to numeric variables and rewards, the remaining two constructs are not supported due to actual structural limitations of ASNets:
\begin{itemize}

    \item \textbf{Universal and existential quantifiers:} Generalisation across problems with ASNets requires a specific invariant to hold: if two actions are instantiated from the same action schema, then there should be a one-to-one correspondence between the propositions appearing in the preconditions/effects of the first action and the preconditions/effects of the second action.
    This should be true even if the two actions are instantiated for different problems from the same domain, and is essential to the mechanism by which ASNets generalise control knowledge across problems.
    The use of universal or existential quantifiers can create pairs of ground actions which are instantiated from the same action schema, but which do not even have the same number of propositions in their respective preconditions or in their respective effects.
    Thus, quantifiers are not supported.
    One way to lift this architectural limitation would be to augment the \textit{action modules} described in \cref{ssec:action-layers} with something akin to the pooling mechanism used for \textit{proposition modules} in \cref{ssec:prop-layers}.
    We did not require this capability for our evaluation domains, and so did not investigate it further.

    \item \textbf{Arbitrary goal formulae:} Each problem associated with a given PPDDL domain could have a different formula describing its goal.
    These formulae might have different structures in different problems: one problem could have a goal expression consisting of a single literal, while another might have a goal expression with deeply nested conjunctions, disjunctions, quantifiers, and so on.
    To train a generalised policy, there needs to be some regular way of representing goal expressions from different problems.
    One could imagine addressing this issue by compiling the goal formula for any given problem into a new or existing action.
    Unfortunately, that would violate ASNets' requirement that all actions for all problems in a domain be instantiated from exactly the same set of action schemas, and so a different solution is required.
    In \cref{ssec:inputs-outputs}, we instead suggest using a vector that indicates, for each proposition in the problem, whether the proposition must be made true in the goal state.
    Propositions that do not need to be made true are assumed to be irrelevant to the goal.
    This representation is only suitable for goals which are conjunctions of positive literals.
    Lifting this restriction would likely require some kind of goal-processing network that generalises to different goal expressions in the same way that ASNets generalise to different planning problems.
    We leave this challenging problem to future work.
    
    The same limitations do \textit{not} apply to action preconditions.
    The ``structure'' of an action's precondition is determined by the corresponding action schema in the domain, and so does not change across different problems from the same domain.
\end{itemize}

Finally, a note on grounding: the internal structure of an ASNet for a particular task is dependent on the number of actions and propositions in the grounded problem, which is in turn dependent on the choice of grounding algorithm.
The grounding algorithm does \textit{not} affect the number and shape of network parameters, which is problem-independent (as described in \cref{sec:asnets}).
However, it does affect the number of ``neurons'' in an ASNet and their connectivity: a naive grounding algorithm could produce a much larger network than a grounding algorithm with sensible optimisations.
Our experiments in \cref{sec:expts} use the grounding code from MDPSim~\cite{younes2005first}, which was introduced for use in IPC-4.
MDPSim's grounding code supports typed parameters for action schemas, and also includes some basic optimisations to avoid instantiating propositions that can never be made true, or actions that can never be enabled.
%
The running example in \cref{sec:asnets} assumes the use of a grounding algorithm with similar optimisations.

\vspace{1.5em} 
\section{Action Schema Networks}%
\label{sec:asnets}

\begin{figure}[t]
  \begin{mdframed}
    \begin{small}
    \input{./code-samples/unreliable-robot.tex}
    \end{small}
  \end{mdframed}
\vspace{-1em}
\caption{
  Part of the PPDDL description of a simple problem that we will use to
  illustrate the structure of ASNets.
  This is a toy navigation domain where a robot, $\mobj{shakey}$, is tasked with
  moving from place to place in a building using movement actions of the form
  $\maschema{drive}(\mobj{shakey}, \mobj{?from}, \mobj{?to})$.
  When invoked, a $\maschema{drive}$ action moves the robot from its initial
  position to its destination successfully with 90\% probability, and does
  nothing the remaining 10\% of the time.
  PPDDL actions combine the add and delete lists into a single
  \texttt{:effect} declaration; here the add list includes
  $\maschema{at}(\mobj{?r}, \mobj{?to})$, while \texttt{(not …)}
  indicates that $\maschema{at}(\mobj{?r}, \mobj{?from})$ belongs on the
  delete list.
  In the specific problem depicted above, the robot must move from the
  $\mobj{kitchen}$---as specified in the initial state declaration,
  \texttt{(:init …)}---to the $\mobj{office}$---which satisfies the goal
  formula, \texttt{(:goal …)}.
}\label{fig:ppddl-rel-ex}
\end{figure}

\begin{figure}
  \begin{center}
    \begin{small}
      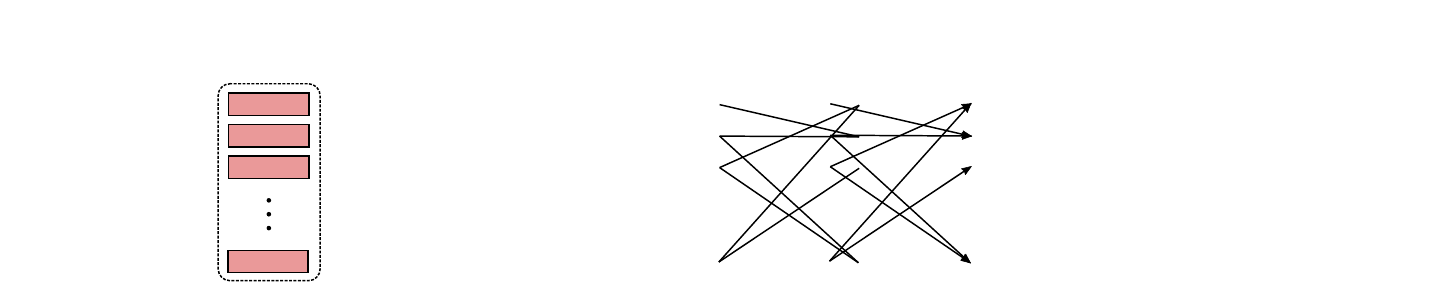
    \end{small}
  \end{center}
  \vspace{-1em}
  \caption{
    High-level overview of an ASNet.
    Each coloured rectangle represents an action module (red) or a proposition module (blue); these modules apply learnt transformations to input feature vectors in order to produce more expressive output feature vectors.
    Information flows from the input (left) to the output (right) along the black lines connecting modules in successive layers.
    For the sake of visual clarity, skip connections (described in the main text) are not depicted.
    Modules are grouped into $L$ proposition layers and $L+1$ action layers.
    Throughout, we refer to such a network as an ``$L$-layer ASNet''.
  }\label{fig:asnet-diagram}
\end{figure}

In this section, we will describe and extend Action Schema Networks (ASNets), which were introduced in past work by \citeA{toyer2018action}.
The approximate structure of an ASNet is illustrated in \cref{fig:asnet-diagram}.
An ASNet transforms a feature representation of the current state $s$ into a policy $\pi^\theta(a|s)$ via an alternating sequence of \textit{action layers} and \textit{proposition layers}.
Each action layer consists of a single \textit{action module} (\cref{ssec:action-layers}) per action.
An action module takes a vector of features from proposition modules in the previous layer and outputs a new vector of features which the network can use to capture relevant properties of the state.
Similarly, each proposition layer consists of a \textit{proposition module} (\cref{ssec:prop-layers}) for each proposition.
Each such module takes a vector of input features from action modules in the previous layer, and produces a new vector of features.
Proposition modules in one layer are connected to action modules in the next layer according to a certain notion of \textit{relatedness} of actions and propositions (\cref{ssec:related}).
This connectivity scheme enables modules to share weights in such a way that the
size and shape of learnt weights is the same for \textit{all} ASNets from a given
domain, even if the ASNets correspond to problems of different sizes.\footnote{
  A note on terminology: throughout this paper, we use ``an ASNet'' to refer to
  a network instantiated for a specific problem instance $\zeta$ from a domain
  $\mathcal D$.
  A specific ASNet is only capable of selecting actions for the corresponding
  instance $\zeta$, but its \textit{weights} will be transferable to an
  ASNet for any other problem $\zeta'$ in the same domain $\mathcal D$.
}
As a result, a policy represented by an ASNet can be applied to any problem from a given domain.

\subsection{Relatedness}%
\label{ssec:related}

The structure of an ASNet is determined by the \textit{relatedness} of actions and propositions in the corresponding PPDDL problem.
To define relatedness, we first need to define the notion of a \textit{lifted proposition}: in the context of a specific PPDDL action schema, a lifted proposition is a specific combination of action parameters applied to a predicate.
In the $\mactn{drive}{?r,?from,?to}$ action schema from the \texttt{unreliable-robot} problem in \cref{fig:ppddl-rel-ex}, we say that $\mpropn{at}{?r,?from}$ is the first unique lifted proposition, $\mpropn{path}{?from,?to}$ is the second unique lifted proposition, and $\mpropn{at}{?r,?to}$ is the third unique lifted proposition.
This notion allows us to define relatedness: a ground action $a$ and ground proposition $p$ are related at position $k$---as denoted by the predicate $R(a, p, k)$---if $p$ corresponds to the $k$th unique lifted proposition in the action schema from which $a$ was instantiated.
As a result, an action and proposition are related at some position when the proposition appears in one of the action's preconditions or effects.
We use this notion to connect action and proposition modules in adjacent layers, as described in \cref{ssec:action-layers} and \cref{ssec:prop-layers}.
Note that this definition of ``relatedness at position $k$'' is distinct from the notion of relatedness used by \citeA{toyer2018action}, which did not draw a distinction between propositions in different positions that were instantiated from the same predicate.
We will see in \cref{ssec:prop-layers} that the concept of positions leads to a slightly more expressive form of pooling at the inputs to the proposition modules.

To see how our notion of relatedness applies to actual ground actions, consider the $\maschema{drive}(\mobj{shakey}, \mobj{kitchen}, \mobj{hall})$ action produced by grounding the \texttt{unreliable-robot} problem.
This action can be executed when $\mpred{at}(\mobj{shakey},$ $\mobj{kitchen})$ $\land$ $\mpred{path}(\mobj{kitchen}$, $\mobj{hall})$ holds, in which case it has the effect of making $\mpred{at}(\mobj{shakey}, \mobj{hall})$ true and $\mpred{at}(\mobj{shakey}, \mobj{kitchen})$ false with probability 90\%, or doing nothing otherwise.
Hence, $\maschema{drive}(\mobj{shakey}, \mobj{kitchen}, \mobj{hall})$ is related to
$\mpred{at}(\mobj{shakey}, \mobj{kitchen})$ at position $k=1$,
$\mpred{path}(\mobj{kitchen}, \mobj{hall})$ at position $k=2$,
and $\mpred{at}(\mobj{shakey}, \mobj{hall})$ at position $k=3$.
Conversely, the set of actions related to $\mpred{at}(\mobj{shakey},$ $\mobj{kitchen})$ will include $\mpred{drive}(\mobj{shakey},$ $\mobj{kitchen},$ $\mobj{hall})$ at position 1 and $\mpred{drive}(\mobj{shakey},$ $\mobj{hall},$ $\mobj{kitchen})$ at position 3.
No other $\maschema{drive}$ actions take $\mobj{shakey}$ to or from the $\mobj{kitchen}$.
It's worth reiterating that these position numbers reflect the order in which propositions first appear in the action schema of \cref{fig:ppddl-rel-ex}.
We do not double-count identical \textit{lifted} propositions: although $\mpred{at}(\mobj{shakey}, \mobj{kitchen})$ appears twice in the action definition (once in the preconditions and once in the effects), we say that it only occurs at one unique position.
Hypothetically, if we had a $\maschema{drive}(\mobj{shakey}, \mobj{kitchen}, \mobj{kitchen})$ action, then it \textit{would} be related to $\mpred{at}(\mobj{shakey}, \mobj{kitchen})$ at two positions ($k=1$ and $k=3$), because the same ground proposition corresponds to two lifted propositions with different arguments in the action schema.

The architecture of an ASNet only depends on aspects of a PPDDL domain and problem that affect the corresponding relatedness graph.
Observe that in the \texttt{unreliable-robot} problem, the relatedness of actions and propositions is only a very coarse encoding of the semantics of those actions and propositions.
For instance, the fact that a $\maschema{drive}$ action can only change the propositions that appear in its (probabilistic) effect 90\% of the time did not change the relatedness graph.
Likewise, whether a proposition appears negated or un-negated in the precondition of an action does not affect is relatedness to that action.
This is why it is straightforward for the ASNets architecture to ``support'' so many of the PPDDL features discussed in \cref{sec:background}: most of them do not affect relatedness, and are consequently irrelevant to the high-level structure of the network.
The missing semantics are of course important to choosing actions at execution time, but the logic for making those decisions is captured by the \textit{weights} of an ASNet during training, rather than being directly encoded in the architecture of the network.

\subsection{Action Layers}\label{ssec:action-layers}

\begin{figure}
  \begin{center}
    \begin{small}
      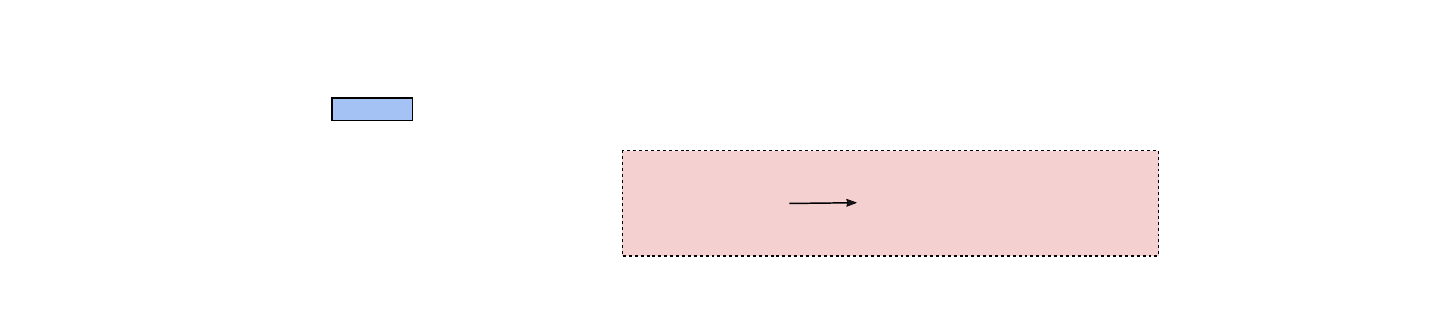
    \end{small}
  \end{center}
  \vspace{-1em}
  \caption{
    Schematic of an action module for the \texttt{unreliable-robot} problem
    (\cref{fig:ppddl-rel-ex}), including skip connection at top.
    The propositions related to $\maschema{drive}(\mobj{shakey}$, $\mobj{kitchen}$, $\mobj{hall})$ are $\mpred{path}(\mobj{kitchen}$, $\mobj{hall}),$ $\mpred{at}(\mobj{shakey},$ $\mobj{kitchen})$ and $\mpred{at}(\mobj{shakey}$, $\mobj{hall})$.
    Thus, the corresponding proposition module representations are concatenated together and then joined with the previous action representation $\hidact{a}{l-1}$ to produce an input vector $\inact{a}{l}$.
    $\inact{a}{l}$ is then passed through a learnt affine transform $W_a^l \inact{a}{l} + b_a^l$ and fixed nonlinearity $f(\cdot)$ to produce an output representation $\hidact{a}{l}$.
  }\label{fig:act-mod}
\end{figure}

Consider an ASNet with $L+1$ action layers, numbered $l = 1, \ldots, L+1$.
In this section, we will examine the structure of the $L-1$ intermediate layers
$2, \ldots, L$.
The structure of first (input) layer and final $(L+1)$th (output) layer is slightly different, and will be deferred to \cref{ssec:inputs-outputs}.
The $l$th intermediate action layer is composed of an action module for each action $a \in \mathcal A$.
Each such module takes as input some vector $\inact{a}{l} \in \mathbb R^{d_a^l}$
and produces as output another vector $\hidact{a}{l} \in \mathbb R^{d_h}$.
We refer to the output size $d_h$ as the \textit{hidden dimension} of the
network.
To construct the input $\inact{a}{l}$ to the action module, we first enumerate all related
propositions $p_1, \ldots, p_M$, then concatenate the corresponding hidden representations $\hidprop{p_1}{l-1}, \ldots, \hidprop{p_M}{l-1}$ from the preceding proposition layer of the network.
Each of these $M$ inputs will themselves be $d_h$-dimensional, for a total input size of $\inact{a}{l} = M \cdot d_h$.
In later layers, we also include a \textit{skip connection} that feeds the
representation $\hidact{a}{l-1}$ for action $a$ from the previous action layer
into the action module for the current layer.
\citeA{toyer2018action} did not include skip connections; we have introduced them to make it easier for the network to propagate information across many layers.
The input vector $\inact{a}{l}$ for the $l$-th layer action module for $a$ is thus the concatenation of $M+1$ earlier output vectors:
\begin{equation}
  \inact{a}{l} = \begin{bmatrix}
    \hidprop{p_1}{l-1}\\ \vdots\\ \hidprop{p_M}{l-1}\\ \hidact{a}{l-1}
  \end{bmatrix}~.
\end{equation}
An output is computed from the input via $\hidact{a}{l} =
f\left(W_a^l\inact{a}{l}+b_a^l\right)$, where $f(\cdot)$ is some fixed
nonlinearity and $W_a^l \in \mathbb R^{d_h \cdot d_a^l}, b_a^l \in \mathbb
R^{d_h}$ are learnt parameters.
An example of such an action module for the \texttt{unreliable-robot} problem is
shown in \cref{fig:act-mod}.

Crucially, the action modules in a given layer are constructed in such a way as
to enable weight sharing, which (as we will see) ultimately allows ASNets to
apply the same set of learnt weights to any problem in a PPDDL domain.
Consider an action $a \in \mathcal A$ and its corresponding action schema $\getop(a) \in \mathbb A$.
We can enumerate the ground propositions related to $a$ by first listing the lifted propositions in the precondition and effect of $\getop(a)$, ignoring ``duplicate'' lifted propositions that apply the same arguments to the same predicate.
Next, we ground those lifted propositions by binding their arguments to the same objects used to instantiate $a$ from $\getop(a)$, thereby yielding all propositions $p_1, \ldots, p_K$ related to $a$ through positions $k=1, 2, \ldots, K$.
Notice that if we repeat this procedure for some $b \in \mathcal A$ with the same schema (so that $\getop(b) = \getop(a)$), then we will obtain another equal-length list of propositions $q_1, \ldots, q_K$ related to $b$.\footnote{
  This assumes there are no quantifiers in the domain description, as noted in \cref{sec:background}.
}
Although $q_1, \ldots, q_K$ may be distinct from $p_1, \ldots, p_K$, the propositions still have a semantic correspondence.
Specifically, if we always enumerate the lifted propositions in $\getop(a)$ in a consistent order, then it will always be the case that $p_j$ and $q_j$ are instantiated from the same predicate, and that they perform a similar ``role'' in $a$ and $b$, respectively.
Further, the equal length of the proposition lists means that $d_a^l = d_b^l$.
Hence, we can tie the weights for the modules for $a$ and $b$ in layer $l$ so that $W_a^l = W_b^l$ and $b_a^l = b_b^l$, and do likewise for all other modules that share the same action schema.

Our weight sharing scheme forces the modules to learn a generic transformation which can be applied by a module for \textit{any} action instantiated from a given schema.
This allows us to generalise across problems: because the number and structure of weights depends only on the action schema, we can re-use the same set of learnt weights for any problem in a domain.
Our weight sharing scheme is reminiscent of the way that convolutional neural networks learn filters which can be applied to an image at any location.
The filter sharing employed by convolutional neural networks improves data
efficiency and introduces useful invariances (e.g.\@ translation invariance)~\cite{lecun1995convolutional}, and we expect similar benefits from
weight sharing in ASNets.

\subsection{Proposition Layers}%
\label{ssec:prop-layers}

Proposition layers operate analogously to action layers.
An $L$-layer ASNet contains $L$ proposition layers numbered $l = 1, \ldots, L$.
For each proposition $p \in \mathcal P$, the $l$th proposition layer contains a
corresponding proposition module which turns some input $\inprop{p}{l} \in
\mathbb R^{d_p^l}$ into a new hidden representation $\hidprop{p}{l} \in
\mathbb R^{d_h}$, where $d_p^l$ is the input dimension of the hidden module.
Again, the input $\inprop{p}{l}$ and output $\hidprop{p}{l}$ are related via a
transformation $\hidprop{p}{l} = f\left(W_p^l \inprop{p}{l} + b_p^l\right)$, for
some fixed nonlinearity $f(\cdot)$ and learnt weights $W_p^l \in \mathbb R^{d_h
\cdot d_p^l}, b_p^l \in \mathbb R^{d_h}$.
The main difference between action and proposition modules lies in the way that
the input $\inprop{p}{l}$ is constructed, which we consider in detail below.

The need for a different mechanism for computing inputs to proposition modules arises from the fact that two propositions instantiated from the same predicate may be related to a \textit{different} number of actions.
As an example, consider the $\mpred{at}(\mobj{shakey},$ $\mobj{hall})$
proposition in the \texttt{unreliable-robot} problem (\cref{fig:ppddl-rel-ex}).
$\mobj{shakey}$ can travel to or from the $\mobj{hall}$ via the
$\mobj{kitchen}$ or $\mobj{office}$.
Hence, the four actions related to $\mpred{at}(\mobj{shakey},$
$\mobj{hall})$ will be
$\maschema{drive}(\mobj{shakey},$ $\mobj{hall}$, $\mobj{kitchen})$ and
$\maschema{drive}(\mobj{shakey},$ $\mobj{hall}$, $\mobj{office})$
at position 1, as well as
$\maschema{drive}(\mobj{shakey},$ $\mobj{kitchen}$, $\mobj{hall})$ and
$\maschema{drive}(\mobj{shakey},$ $\mobj{office}$, $\mobj{hall})$
at position 2.
On the other hand, there is only one path leading to and from the
$\mobj{kitchen}$, which goes straight from the $\mobj{hall}$, and so there will
only be two actions related to the proposition $\mpred{at}(\mobj{shakey}$,
$\mobj{kitchen})$.
It will not suffice to construct the input for an $\mpred{at}(\cdot, \cdot)$ module by
simply concatenating the representations for all related actions.
If we did so then the inputs for different $\mpred{at}(\cdot, \cdot)$ modules would be
of different sizes, and we could not share weights between them.

\begin{figure}
  \begin{center}
    \begin{small}
      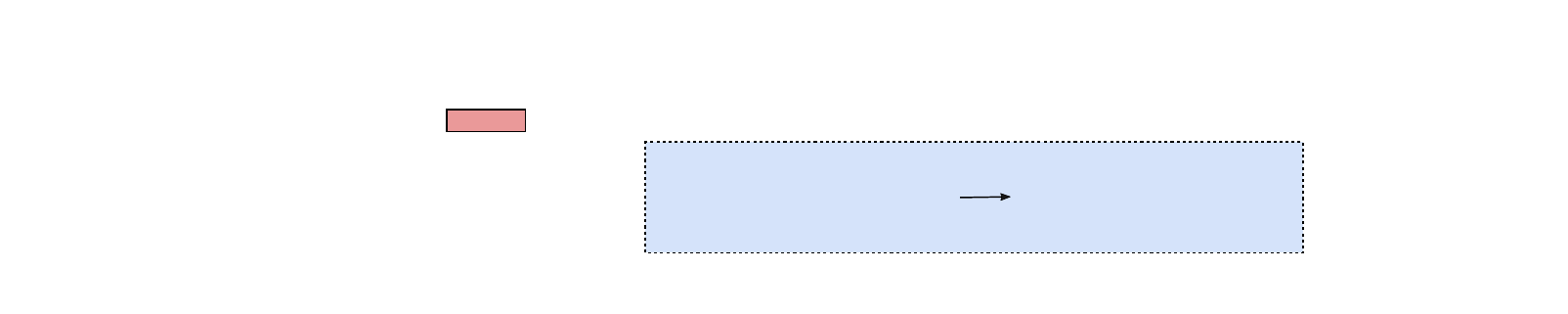
    \end{small}
  \end{center}
  \vspace{-1em}
  \caption{
    A proposition module for the \texttt{unreliable-robot} problem (\cref{fig:ppddl-rel-ex}), again including a skip connection
    Here there are four $\maschema{drive}$ actions related to $\mpred{at}(\mobj{shakey},$
    $\mobj{kitchen})$, including two in position $k=1$ (top two), and two in
    position $k=2$ (bottom two).
    Related actions in the same position are pooled together using two separate
    pooling operations.
    The pooled representations are then concatenated together with the output $\hidprop{p}{l-1}$ from the corresponding module in the previous proposition layer (dashed line) to produce an input $\inprop{p}{l}$.
    Finally, the input is passed through the learnt transformation $\hidprop{p}{l} = f(W_p^l \inprop{p}{l} + b_p^l)$.
  }\label{fig:prop-mod}
\end{figure}

Instead of constructing the input $\inprop{p}{l}$ to the $l$-th layer module for proposition $p$ using concatenation, we choose to pool over related actions.
First, we enumerate all action schemas $A_1, \ldots, A_S$ which refer to the predicate $\getpred(p)$ in a precondition or effect through positions $k_1, \ldots, k_S$.
Clearly, any action related to $p$ must be instantiated from one of these schemas.
Further, an action schema may appear more than once in this list if it is related to $p$ through more than one position.
From the action schema list, we construct the input $\inprop{p}{l}$ using
\begin{equation}
  \inprop{p}{l} = \begin{bmatrix}
    \pool(\{{\hidact{a}{l}} \mid \mathcal \getop(a) = A_1 \land R(a, p, k_1)\})\\
    \vdots\\
    \pool(\{{\hidact{a}{l}} \mid \mathcal \getop(a) = A_S \land R(a, p, k_S)\})\\
  \end{bmatrix}~,
\end{equation}
where $\pool(\{x_1, \ldots, x_R\}) \in \mathbb R^{d_h}$ is a function that takes
an arbitrary-size set of vectors $x_1, \ldots, x_R \in \mathbb R^{d_h}$ and
aggregates their elements into a single vector.
In later proposition layers, we also introduce skip connections between successive proposition layers, so that $\inprop{p}{l}$ will also include the previous-layer output representation $\hidprop{p}{l-1}$ for proposition $p$.
An example proposition module is illustrated in \cref{fig:prop-mod}.

There are many possible implementations of the pooling function $\pool(\cdot)$.
One could perform pooling by averaging corresponding elements (mean pooling), taking an elementwise maximum (max pooling), and so on.
In principle, this pooling process could destroy information that would be relevant to expressing a generalised policy.
In practice, we have found that a simple max pooling strategy is sufficient to solve a range of interesting problems.

Originally, in the work of~\citeA{toyer2018action}, ASNet proposition modules included only a single pooling operation for each related action schema, rather than a separate pooling operation for every related action schema and position.
Thus, inputs corresponding to actions related to a proposition through different positions could not be distinguished by the relevant proposition module.
To see how this might cause problems, imagine an \texttt{unreliable-robot} problem with only two locations: $\mobj{office}$ and $\mobj{hall}$.
In such a problem, the set of $\maschema{drive}$ actions related to $\mpropn{at}{\mobj{shakey}, \mobj{office}}$ would be the same as the set of actions related to $\mpropn{at}{\mobj{shakey}, \mobj{hall}}$, so both $\mpropn{at}{\mobj{shakey}, \cdot}$ proposition modules would share the same representation!
Intuitively, the old pooling scheme could not easily tell whether $\mobj{shakey}$ was leaving a room or entering it.
The pooling scheme here is more expressive, as it enables a proposition module to easily distinguish between inputs for related actions in which the proposition plays a different ``role'', such as appearing in a precondition for one action, and an effect for a different action.

After pooling, the input dimension of a proposition module, $d_p^l = S \cdot d_h$, is the same for all proposition modules instantiated from predicate $\getpred(p)$, so we can again use weight sharing.
Specifically, we define $W_p^l$ to equal $W_q^l$ in the $l$th proposition layer whenever $\getpred(p) = \getpred(q)$.
As with weight sharing in action modules, this allows us to generalise over
different problems drawn from the same domain.
In particular, the complete set of weights for any problem in a given domain
will be
\begin{equation}
\begin{split}
  \theta =
  \{W_a^l, b_a^l \mid 1 \leq l \leq L + 1, a \in \mathbb A\}
  \cup \{W_p^l, b_p^l \mid 1 \leq l \leq L, p \in \mathbb P\}~,
\end{split}
\end{equation}
where we have abused notation slightly by using $a$ and $p$ to refer to action
schemas and predicates instead of actions and propositions.
It is thus possible to learn a generalised policy by acquiring a fixed set of
weights $\theta$ using some small training tasks, and then transferring them to
much larger
problems.

\subsection{Input \& Output Action Layers}%
\label{ssec:inputs-outputs}

The first action layer of an ASNet is the input layer of the network, and thus
has a different input scheme to later layers.
The input to a first-layer module for a given action $a$ is composed of
proposition truth values, binary goal indicators, and a binary indicator to show
whether the action is applicable in the current state.
To make this concrete, consider the propositions $p_1, \ldots, p_M$ which are
related to $a$ through each position $k_1, \ldots, k_M$.
We define the truth value vector $v \in \{0, 1\}^M$ to have $v_j = 1$ if $p_j$
is true in the current state and $v_j = 0$ otherwise, and define the goal
information vector $g \in \{0,1\}^M$ to have $g_j = 1$ if a proposition appears
unnegated in the goal and $g_j = 0$ otherwise.\footnote{
  As noted earlier in \cref{sec:background}, this representation limits us to goals which are conjunctions of positive literals.
}
Further, define $m = 1$ if $a$ is applicable in the current state, and $m = 0$ otherwise.
The input to the first-layer action module corresponding to $a$ is
\begin{equation}\label{eqn:act-mod-input-noheur}
  \inact{a}{1} = \begin{bmatrix}v \\ g \\ m\end{bmatrix}~.
\end{equation}
In \cref{ssec:heur-inputs} we further extend this input representation to use information derived from heuristics evaluated at the current state, which our experimental evaluation shows is critical to allowing ASNets to solve problems in some domains.

The final action layer of an ASNet is also slightly different to the preceding
ones.
At the final layer, we would like an ASNet to give us a probability
$\pi^\theta(a_i|s)$ that a given action $a_i$ is the correct action to take in a
given state $s$.
Hence, we stipulate that the module for a given action $a_i$ in the final
$(L+1)$th action layer of an ASNet should produce a single logit
$\hidact{a_i}{L+1} \in \mathbb R$.
The probability that $a_i$ should be selected in $s$ is thus proportional to $\exp \hidact{a_i}{L+1}$.
To normalise these probabilities, and to ensure that only applicable actions can be
selected, we then pass the logits through a masked softmax activation.
Let $m_1, \ldots, m_N \in \mathbb \{0, 1\}$ be binary indicators of whether each
action $a_i$ is applicable (1) or not (0), and let $\hidact{a_1}{L+1}, \ldots,
\hidact{a_N}{L+1} \in \mathbb R$ be the unscaled log probabilities produced by
the action modules in the final action layer.
The scaled output of the ASNet is then
\begin{equation}
  \pi^\theta(a_i | s) = \frac{m_i \exp\left( \hidact{a_i}{L+1} \right)}
               {\sum_{j=1}^N m_j \exp\left( \hidact{a_j}{L+1} \right)}~.
\end{equation}
Masking out disabled actions ensures that the ASNet is only trained to
distinguish between actions relevant to a given state.
The non-temporal variant of the earlier FPG planner~\cite{buffet2009factored}
uses the same strategy.

\subsection{Heuristic Inputs and the Receptive Field}%
\label{ssec:heur-inputs}

The literature on traditional convolutional neural networks sometimes discusses
the ``receptive field'' (or ``effective receptive field'') of an activation in a
certain layer of the network~\cite{long2015fully}.
This refers to the region of the input image that is able to influence the output of that activation.
Because convolutional neural networks are only locally connected at each layer, it can take many layers to propagate information about the input image to different regions of the network itself, and so the receptive field of an activation in one of the earlier layers of a network will often only be a small sub-window of the input image.
The activations in an ASNet suffer from a similar limitation.
As a result, the maximum length of chains of actions and propositions that the network can reason about is limited by its depth, since each action or proposition layer only propagates information one step along a given chain of related actions and propositions.

To make the receptive field limitation more concrete,
consider a restricted class of problems from the \texttt{unreliable-robot}
domain of \cref{fig:ppddl-rel-ex}.
In this restricted class of problems, the robot starts at some location $m$, and
can move along one of two unidirectional chains: a left chain $m \to l_1 \to
l_2 \to \cdots \to l_K$ and a right chain $m \to r_1 \to r_2 \to \cdots \to r_K$.
If the goal is always either moving to $r_K$ or moving to $l_K$, then the main challenge faced by the agent will be deciding whether to move to $r_1$ or $l_1$ from the initial location $m$.
After it has reached $r_1$ or $l_1$, it will have no choice but to follow the corresponding chain to the end.
The ASNet will be given information about which chain contains the goal via the $g$ vector passed to the $\maschema{drive}(\mobj{shakey}$, $l_{K-1}$, $l_K)$ and $\maschema{drive}(\mobj{shakey}$, $r_{K-1}$, $r_K)$ action modules in the first layer.
Those modules correspond to actions at the end of either chain, which are initially disabled.
Instead, it is the action module for $\maschema{drive}(\mobj{shakey}$, $m$, $l_1)$ in the last layer which governs whether or not the agent chooses to move from $m$ to $l_1$ in the initial state.
If we ignore skip connections, then this module is connected to the module for $\maschema{drive}(\mobj{shakey}$, $l_{K-1}$, $l_K)$ in the input layer via a related action and proposition chain of the form
\begin{align*}
  R(&\maschema{drive}(\mobj{shakey}, m, l_1), \mpred{at}(\mobj{shakey}, l_1), 3)\\
  &\rightarrow R(\maschema{drive}(\mobj{shakey}, l_1, l_2), \mpred{at}(\mobj{shakey}, l_1), 1)\\
  &\rightarrow R(\maschema{drive}(\mobj{shakey}, l_1, l_2), \mpred{at}(\mobj{shakey}, l_2), 3)\\
  &\rightarrow \cdots\\
  &\rightarrow R(\maschema{drive}(\mobj{shakey}, l_{K-1}, l_K), \mpred{at}(\mobj{shakey}, l_{K-1}), 1)~,
\end{align*}
and likewise for the right chain.
Unfortunately, the length of this relatedness chain depends on the size $K$ of the problem.
If a network has fewer than $K$ proposition layers and $K + 1$ action layers then it will be impossible to communicate information about the goal back from $\maschema{drive}(\mobj{shakey}$, $r_{K-1}$, $r_K)$ to $\maschema{drive}(\mobj{shakey}$, $m$, $l_1)$ and $\maschema{drive}(\mobj{shakey}$, $m$, $r_1)$.
It is thus impossible for any fixed-depth ASNet to obtain a generalised policy for this entire subclass of arbitrary-depth problems, as we show experimentally in \cref{app:expts-recept}.

To overcome the receptive field limitation, we supply each action module in the first layer of an ASNet with two kinds of ``heuristic inputs''.
First, we include features derived from the landmarks identified by the LM-cut heuristic in each state~\cite{helmert2009landmarks}.
Second, we add features counting how many times each action has been taken over the current trajectory.

We will begin by describing the LM-cut features.
LM-cut derives a lower bound on the cost-to-go for a problem by identifying a set of disjunctive action landmarks for a delete-relaxed, and possibly determinised, version of that problem.\footnote{
  The \textit{delete relaxation} of a planning problem is one in which all ``negative''
  effects (those which make a proposition false) are removed from each
  action.
  A \textit{determinisation} of an SSP is a deterministic planning problem in
  which each action with stochastic effects is mapped to one or more
  (inequivalent) actions with only deterministic effects.
  These ideas are explained further by \citeA{mausam2012planning}.
}
Each disjunctive action landmark is a set of actions where at least one action
must appear along any path to the goal.
By supplying information about the landmarks recovered by LM-cut directly to the
network, we can improve its ability to reason about which actions will have
helpful long-term consequences.
Specifically, for each action $a$, we create a new indicator vector $c \in \{0,
1\}^3$ to use as an auxiliary input to the corresponding first-layer action
module.
We have $c_1 = 1$ iff $a$ appears 
as the only action in at least one landmark; $c_2 = 1$ iff $a$ appears in a
landmark containing two or more actions; and $c_3 = 1$ iff $a$ does not appear
in any landmark.
These values are concatenated to the input $\inact{a}{1}$ for the first-layer action module for action $a$.

In addition to features derived from LM-cut landmarks, we also include a count $c_a$ of the number of times each action $a$ has been executed over the course of the current trajectory.
$c_a$ is concatenated to the first-layer action module input $\inact{a}{1}$.
We found that this information was useful in domains where the ASNet was unable to distinguish some states from each other even with the help of heuristic information, and would sometimes end up in loops where it would repeatedly switch between two adjacent states.

We note that LM-cut landmarks and action counts are not the only form of heuristic information that could serve to lift the receptive field limitation of ASNets.
For example, one could instead feed ASNets information about helpful actions computed by the FF planner~\cite{hoffmann2001ff}, as done in past work~\cite{de2011scaling}.
In the probabilistic setting, it may be more appropriate to supply the ASNet with operator counts produced by a probability-aware heuristic like $h^{\text{roc}}$ or $h^{\text{pom}}$~\cite{trevizan2017occupation}.
In some domains it is likely possible to remove the need for heuristic information entirely by augmenting ASNets with some combination of recurrent modules~\cite{hochreiter1997long}, attention~\cite{vinyals2015pointer}, and memory~\cite{weston2014memory}.
Of course, the downside of a more powerful architecture is that it would weaken the (very strong) inductive bias inherent in ASNets, as well as increase the total computational cost and parameter count of the network.
All three of these consequences would in turn lead to increased training times and a higher chance of overfitting to the training set.
We leave it to future work to experiment with alternative heuristics and determine the optimal tradeoff between model expressiveness and inductive bias in a planning context.

The receptive field limitation of ASNets is reminiscent of the behaviour of short-sighted probabilistic planners~\cite{trevizan2012short} or receding-horizon Model-Predictive Control (MPC) strategies~\cite{camacho2007model}, which both choose actions by planning over short lookahead windows.
However, the horizon for short-sighted planners is defined in terms of the states reachable in some fixed number of steps, whereas the receptive field of an ASNet is defined by the relatedness of state variables and actions.
In \cref{sec:expts} we compare ASNets against a short-sighted planner on several probabilistic planning benchmarks, and demonstrate that ASNets can easily solve problems that the short-sighted planner cannot solve in a reasonable period of time.

\section{Training and Exploiting Generalised Policies}%
\label{sec:train-exploit}

This section explains our mechanism for training and exploiting ASNet-based
policies.
We note that ASNets could easily be trained in other ways with a variety of
different trade-offs.
The focus here is on \textit{simple} training and exploitation methods to
directly evaluate the quality of ASNets as a class of models, as opposed to
evaluating entire learning-based planning systems in which ASNets only play a
small part.
Readers may refer to the related work in \cref{sec:related} for a thorough
survey of those other mechanisms for learning ML-based policies or control
knowledge (\cref{ssec:know-acq}) and exploiting such knowledge
(\cref{ssec:rel-knowl-exploit}).

\subsection{Training via Imitation Learning}


The training procedure for obtaining an ASNet-based generalised policy is depicted in \cref{alg:training}.
The high-level aim of the training procedure is to optimise a set of ASNet weights $\theta$ so that the corresponding ASNets mimic the actions selected by a heuristic search planner---which we call the \textit{teacher planner}---on a collection $P_{\text{train}}$ of training problems from a given domain.
These problems are assumed to be small enough to quickly solve via heuristic search, while also containing structural elements representative of those found in larger problems.
Learning with the training set $P_{\text{train}}$ proceeds over a series of \textit{epochs}, as depicted in \textsc{ASNet-Train} (\cref{alg:training}).
Throughout the epochs, the algorithm maintains a list $\mathcal M$ of encountered states (initially empty), and a current estimate $\theta$ of the ASNet weights (initialised randomly).
Each epoch is in turn divided into an \textit{exploration phase} and a \textit{training phase}, corresponding to the two outer loops in \textsc{Train-Epoch}.
We will now describe each of those phases separately.


\begin{algorithm}[t]
\begin{algorithmic}[1]
\Procedure{ASNet-Train}{}
  \State{$\m M \gets \textit{empty list}$; $\theta \gets \textsc{Random-Initial-Weights}()$; $n \gets 0$}
  \Repeat{}
    \State{$\textsc{Train-Epoch}(\theta, \m M)$}
    \State{$n \gets n + 1$}
  \Until{$n > T_{\text{max-epochs}}$ or early stopping condition satisfied}
\EndProcedure{}

\Procedure{Train-Epoch}{$\theta$, $\mathcal M$}
  \For{$i=1, \ldots, T_{\text{explore}}$}
    \ForAll{$\zeta \in P_{\text{train}}$}
      \State{$s_0, \ldots, s_N \gets \textsc{Run-Policy}(s_0(\zeta), \pi^\theta)$}

      \State{$\mathcal M\text{.extend}(\{s_0, \ldots, s_N\})$}
      \For{$j = 0, \ldots, N$}
        \State{$s_j' \ldots, s_M' \gets \textsc{Teacher-Rollout}(s_j)$}
        \State{$\mathcal M\text{.extend}(\{s_j' \ldots, s_M'\})$}
      \EndFor{}
    \EndFor{}
  \EndFor{}

  \For{$i = 1, \ldots, T_{\text{train}}$}
    \State{$\mathcal B \gets \textsc{Sample-Minibatch}(\mathcal M)$}
    \State{Update $\theta$ using $\frac{d\mathcal L_\theta(\mathcal
        B)}{d\theta}$ (\cref{eqn:batch-objective})}
  \EndFor{}
\EndProcedure{}

\Function{Run-Policy}{$s$, $\pi$}
  \State{$t \gets 0$; $s_t \gets s$}
  \While{$s \notin \m G \land t < T_{\text{trajectory-limit}}$}
  \State{$a_{t} \sim \pi(a_{t} \mid s_{t})$}
  \State{$s_{t+1} \sim \m T(s_{t+1} \mid s_{t}, a_{t})$}
  \State{$t \gets t + 1$}
  \EndWhile{}
  \State{\Return $s_0, \ldots, s_t$}
\EndFunction{}
\end{algorithmic}
\caption{Learning ASNet weights $\theta$ from a set of training problems $P_{\text{train}}$.}
\label{alg:training}
\end{algorithm}

In the exploration phase, \textsc{Train-Epoch} uses the current ASNet parameters $\theta$ to sample $T_{\text{explore}}$ trajectories on each problem $\zeta \in P_{\text{train}}$.
The code for sampling a trajectory is shown in \textsc{Run-Policy}: the ASNet starts in state $s_0$ and must produce an action $a_t$ distributed according to $\pi^\theta(a|s)$ until a goal state or other terminal state is reached, or the trajectory length limit is exceeded (typically based on the dead end penalty $D$).
After obtaining states $s_0, \ldots, s_N$ from a policy rollout and adding them to the state memory $\mathcal M$, we also extend the state memory with a series of rollouts under the teacher planner's policy, including a separate rollout starting from each of $s_1, \ldots, s_N$.
Including ASNet rollout states in $\mathcal M$ ensures that we continue to optimise $\theta$ to yield good action choices in the states that our ASNets visit more often.
On the other hand, including states from teacher policy rollouts ensures that $\mathcal M$ always contains some goal trajectories, and so $\theta$ can be optimised to perform well on states close to the goal even before the ASNet has trained for long enough to reach those states itself.
The use of a mixture of states generated by $\pi^\theta$ and states generated by the teacher is reminiscent of the way that DAgger imitation learning algorithm~\cite{ross2011reduction} interpolates between expert and novice policies when collecting a training dataset.
Our use of highly non-convex neural networks means that we cannot translate the no-regret guarantees of DAgger to this setting.
However, it's probable that our use of a similar strategy makes it less likely that an ASNet will go ``off-distribution'' and encounter a state for which it cannot select a good action at test time.

After extending the state memory $\mathcal M$, ASNets enters a learning phase in which it updates the weights $\theta$.
The learning phase depends on action labels calculated using the teacher planner during the exploration phase.
In particular, when we add a state $s$ to state memory in the exploration phase, we also invoke the teacher planner to obtain a Q-value $Q^{\text{teach}}(s,a)$ with respect to the teacher planner's policy for each enabled action $a$ in $s$.
This allows us to label actions as ``optimal'' or ``sub-optimal'' with respect to the teacher's value function: action $a$ in state $s$ is given a label $y_{s,a} \in \{0,1\}$ that is set to one if
\[
  Q^{\text{teach}}(s,a) = \max_{a'} Q^{\text{teach}}(s,a') 
\]
and zero otherwise.
During the learning phase of each epoch, we repeatedly sample a fixed-size minibatch of states $\mathcal B$ from state memory $\mathcal M$, then optimise $\theta$ to increase the probability of selecting actions with $y_{s,a}=1$.
Specifically, given a minibatch $\mathcal B$, the batch objective for an ASNet is to maximise the cross-entropy-based loss
\begin{equation}\label{eqn:batch-objective}
  \mathcal L_\theta(\mathcal B) = -\frac{1}{|\mathcal B|} \sum_{s \in \mathcal B} \sum_{a \in A}
    \left[
      (1 - y_{s,a}) \cdot \log (1 - \pi^\theta(a|s))
      + y_{s,a} \log \pi^\theta(a|s)
    \right] + \frac{1}{2} \lambda \|\theta\|^2~.
\end{equation}
The last term is an $\ell_2$ regulariser (with constant coefficient $\lambda > 0$) that ensures $\mathcal L_\theta$ is always bounded below as a function of $\theta$; otherwise it is possible to drive $L_\theta(\mathcal B)$ to $-\infty$ if the data is linearly separable.
We update $\theta$ at the end of a learning step by feeding the gradient $\partial \mathcal L_\theta / \partial \theta$ into any appropriate first-order optimiser.
This process of sampling a minibatch and updating $\theta$ is repeated $T_{\text{train}}$ times in each learning phase.
If the ASNet has sufficient expressive power to imitate the teacher, then the parameter updates should ultimately make it follow similar trajectories to the teacher on the training problems.

The training process typically terminates after a fixed number of epochs or maximum amount of time has elapsed.
However, for domains that are relatively easy for ASNets to solve (e.g.\@ Triangle Tireworld), we have found that early termination conditions can sometimes decrease the time required for training.
Specifically, we terminate early if the success rate of the ASNet on the training problems has been at least $p_{\text{solved}}$ for at least $T_{\text{stop}}$ consecutive epochs.
In experiments, we use the same values of $T_{\text{stop}} = 20$ and $p_{\text{solved}} = 1$ for all domains.

For domains where our chosen training problems varied widely in difficulty, we found that \cref{alg:training} would sometimes spend most of the training period running the teacher planner in order to perform a $\textsc{Teacher-Rollout}(\cdot)$ starting from each visited state.
To avoid this problem, we made three modifications to \cref{alg:training}.
First, we cached the results of calls to the planner so that it would only have to be invoked once for each encountered state.
Second, in the first epoch of training, we skipped rolling out the ASNet policy, and instead simply extended the memory $\mathcal M$ with $\textsc{Teacher-Rollout}(s_0(\zeta))$ for the initial state $s_0(\zeta)$ of each problem $\zeta \in P_{\text{train}}$.
This meant that in the second epoch of training, the ASNet was already following a moderately effective, low-entropy policy, and thus encountered fewer unique states.
Combined with caching, this led to fewer planner calls, and thus limited the impact of very difficult problems in the training set.
Third, we put a timeout of 10s on the teacher planner.
If the teacher planner did not succeed in finding a plan or policy starting from a given state within 10s, then the state was omitted from the state memory $\mathcal M$ and instead recorded elsewhere so that the planner would not be invoked on the same state again.
Together, these changes substantially decreased the cost of planning during the training period.

In addition to supervised learning, we also tried training ASNets with Policy Gradient Reinforcement Learning (PG RL) using a similar strategy to the Factored Policy Gradient (FPG) planner~\cite{buffet2009factored}.
Reinforcement learning has the advantage of enabling us to directly minimise the cost of trajectories produced by our ASNet policy on the problems in $P_{\text{train}}$.\footnote{Recall that in the fSSPUDE framework, the cost of trajectories that fail to reach the goal are set to a high constant $D$; hence, minimising the cost of trajectories is typically sufficient to obtain a high probability of reaching the goal, too.}
In contrast, optimising the supervised objective in \cref{eqn:batch-objective} may not lead to a good ASNet policy if the teacher planner's implicit policy is outside of the ASNet's hypothesis space.
Unfortunately, we found that basic PG RL (in our case: REINFORCE with a state-dependent baseline) was simply too inefficient to train ASNet-based policies in any reasonable amount of time.
We leave exploration of more efficient reinforcement learning strategies to future work.

\subsection{Exploitation}

We exploit our ASNet-based generalised policy directly, by repeatedly picking an action
\[
   a_t \in \argmax_{a \in A} \pi^\theta(a|s_t)
\]
in state $s_t$ (breaking ties arbitrarily), then sampling a successor state $s_{t+1}$ from the transition distribution
\[
   s_{t+1} \sim \mathcal T(s' | s_t, a_t)
\]
until a goal is reached.
It would be equally easy to sample an action directly from the output distribution; that is, replacing the $\argmax$ above with
\[
  a_{t} \sim \pi^\theta(a|s_t)~.
\]
A sampling strategy might be preferable to a direct maximisation strategy on problems where ASNets' learnt control knowledge fails to perfectly solve the problem.
In problems without many dead ends, a degree of randomness during evaluation is sometimes sufficient to push an ASNet out of regions of state space where sampling the ``best'' action could lead to loops.
We compare these strategies empirically in \cref{sec:expts}.

We note that it is also possible to use ASNets to guide a heuristic search planner, instead of relying on an ASNet-based policy to solve all problems in a domain directly.
In the probabilistic planning setting, one particularly promising approach is to incorporate ASNets into Monte Carlo tree search algorithms like UCT, in the style of AlphaGo~\cite{silver2016mastering}.
A recent paper has made a preliminary evaluation of various mechanisms for guiding UCT with ASNets, including the use of ASNet-based generalised policies as rollout policies, and the use of ASNets to bias UCB1 successor selection.
These strategies can alleviate the negative impact of inadequate ASNet training, and help solve problems that would be too complex for ASNets to solve on their own~\cite{shen2019guiding}.
Of course, UCT and other algorithms that can make use of learnt search control knowledge are generally agnostic to the type of learnt model that they are used with.
Thus, it is equally possible to plug ASNets into most existing learning-guided combinatorial search algorithms, which we survey in \cref{ssec:rel-knowl-exploit}.
In order to disentangle the effect of model expressiveness from the quality of heuristic search, our evaluation in the next section will eschew these search-based algorithms in favour of the direct execution approach described previously.

\section{Experimental Evaluation}\label{sec:expts}


In this section, we empirically evaluate the performance of ASNets on a range of
probabilistic and deterministic domains, identify which elements of ASNets
contribute the most to performance, and present an extended evaluation on
deterministic Blocksworld.
Code for our experiments is available on GitHub.\footnote{\texttt{https://github.com/qxcv/asnets}}

\subsection{Time-Based Evaluation}\label{ssec:expts-main}

In practice, we envisage that ASNet-style generalised policies will be most useful for solving problems that are too large for heuristic search, but where there also exists some simple domain-specific trick that makes the problem easy to solve.
For instance, in the venerable Blocksworld domain~\cite{slaney2001blocks}, it's known that optimal planning is NP-hard, but that merely finding a ``reasonable'' plan can be accomplished in linear time with a domain-specific algorithm.
For the user of a planning system, the key question is whether the high fixed cost of training ASNets on a set of small problems from a domain is justified by the time saved when one uses ASNets in place of heuristic search on larger problems.
We answer this question for seven probabilistic and deterministic domains by comparing the number and size of problems that ASNets can solve in a given amount of time against the number and size of problems that can be solved by a range of competitive baseline planners.

\subsubsection{ASNet Hyperparameters}

We use the same architecture and hyperparameters for all ASNet experiments, except where explicitly indicated otherwise.
We arrived at these hyperparameters with a two-stage tuning process.
In the first stage, we applied the Ray Tune automated hyperparameter tuning framework~\cite{liaw2018tune} and the random forest optimiser from scikit-optimize\footnote{\texttt{https://github.com/scikit-optimize/scikit-optimize}} to find domain-specific hyperparameter settings that maximised coverage on the benchmark problems after two hours of training.
In the second stage, we manually interpolated between the automatically-tuned, domain-specific hyperparameters to find a common set of hyperparameters that worked well on all domains.
We report those common hyperparameters below.

\paragraph{Training configuration}
Our networks have two proposition layers and three action layers (i.e.\@ $L=2$), with $d_h=16$ output channels for each action or proposition module.
Training is divided into a series of epochs, each of which begins by sampling up to 70 trajectories from the training problems and adding them to the replay buffer, followed by $T_{\text{train}}=700$ batches of network optimisation.
More specifically, at the beginning of each epoch, up to $T_{\text{explore}} = \lceil 70/|P_{\text{train}}|\rceil$ trajectories are sampled from each of the $|P_{\text{train}}|$ different problems simultaneously, with trajectory sampling terminating as soon as each problem has had at least one trajectory sampled.
Early termination of the sampling process allows for more concentrated sampling from small problems where planning is cheap, while still sampling at least some trajectories from larger problems that require more time to sample each trajectory and perform planning.
For probabilistic problems, the default teacher planner used to label collected trajectories is LRTDP with the h-add heuristic on an all-outcomes determinisation.
For deterministic problems, the default teacher planner is \astar{} with the h-add heuristic.
After data collection, the batches used for training the network each consist of 64 samples drawn equally from across the training problems.
Parameter optimisation itself uses the Adam optimiser ($\beta_1=0.9$, $\beta_2=0.999$, $\epsilon=10^{-8}$) with a learning rate of $10^{-3}$.
We apply an $\ell_2$ regulariser of $2 \cdot 10^{-4}$ to prevent weights from exploding, and dropout probability of 0.1 for all layers.
We arrived at these hyperparameters through a mix of automated hyperparameter search and manual fine-tuning.
Training time is capped at two hours, and we set $T_{\text{stop}} = 20$ with $p_{\text{solved}} = 1$.
That is, training can terminate early if the network obtains a success rate of 100\% on its training rollouts for 20 consecutive epochs.

\paragraph{Testing configuration}
When testing ASNets on probabilistic problems, we perform 30 rollouts per test problem using both deterministic execution (where the agent takes a most-favoured action $a_t$ $\in$ $\argmax_{a} \pi(a | s_t)$, with ties broken using the lexicographic order of action names) and stochastic execution (where the agent samples $a_t \sim \pi(a | s_t)$).
The results of stochastic execution are marked with a ``PE'' (for Probabilistic Execution) in graphs and results tables.
When testing ASNets on deterministic domains, we again test both the strategy of picking the action with highest $\pi(a | s_t)$, and the strategy of sampling $a_t \sim \pi(a | s_t)$.
Again, the runs that use the latter strategy are marked with a ``PE''.
However, when choosing actions deterministically in deterministic domains, we only perform a single rollout instead of 30 rollouts, since all rollout outcomes are identical.
Likewise, when choosing actions stochastically in deterministic problems, we only perform 10 rollouts instead of 30.

In addition to varying the strategy used to pick actions, we also include results for ASNet runs where there are no LM-cut landmarks or past action counts available to the network.
This gives us an indication of how essential those features are in different domains.
In graphs and tables, runs with neither landmarks nor action counts are marked ``no h.'' (for ``no heuristic inputs'').
Further, in probabilistic problems, we perform experiments using a teacher planner with the (admissible) LM-cut heuristic on an all-outcomes determinisation, instead of the default h-add heuristic on an all-outcome determinisation.
In some domains we found that the h-add heuristic led to slightly suboptimal teacher policies (e.g.\@ this was the case in Exploding Blocksworld and Probabilistic Blocksworld, as explained in \cref{ssec:expts-by-domain} and \cref{ssec:other-discuss}).
Substituting in LM-cut ensures the ASNet is \textit{always} trained to mimic optimal action choices on the training problems, albeit at the cost of the teacher planner sometimes taking longer to converge.
Runs with the LM-cut heuristic are marked ``adm.'' (short for ``admissible teacher heuristic'').

\subsubsection{Baselines}

On probabilistic domains, we compare against a total of four different baseline configurations, each composed of one of two different baseline planners and one of two heuristics.
The baseline planner is either LRTDP~\cite{bonet2003labeled} or SSiPP~\cite{trevizan2017occupation}, while the heuristic is either LM-cut~\cite{helmert2009landmarks} or h-add~\cite{geffner2000admissible} heuristic, both with the all-outcomes determinisation.
For each problem, we evaluated each baseline by sequentially performing 30 runs in which planners were re-initialised from scratch with different random seeds, and applied a three-hour cutoff.
Runs that were not evaluated within the cutoff time were marked as failures.
Specific planner configurations were as follows:
\begin{description}
\item[LRTDP]
During each evaluation run, LRTDP was executed until its value estimates converged to within a tolerance of $\epsilon = 10^{-4}$, after which we performed a single rollout of the recovered policy.
If the value estimates did not converge within 60 seconds of the end of the allotted time, we suspended the planning process and executed a greedy policy with respect to the un-converged state values for a maximum of 300 time steps.
We found that this modification fractionally increased cumulative coverage on some domains, relative to counting runs where LRTDP did not converge as failures.
We included LRTDP in our comparison as its performance is representative of the kind of heuristic search planners that are currently popular for solving SSPs.
\item[SSiPP]
SSiPP was repeatedly executed as a re-planner until 60 seconds before the time cut-off, or until it completed 50 successful rollouts (whichever came first).
The short-sighted SSPs encountered by SSiPP were constructed with a horizon of 3 steps, and solved with LRTDP.
After termination, SSiPP's policy was executed once, regardless of whether or not it had converged.
We included SSiPP in our comparison because it makes use of limited lookahead to solve problems.
We can thus compare its results to those of the ASNet policies to evaluate the degree to which trained ASNets behave like limited lookahead planners in practice.
\end{description}

On deterministic domains, we compare against three different types of baseline planners in five different configurations, which we describe below.
For all planners, we used the corresponding implementation from Fast Downward~\cite{helmert2006fast}.
\begin{description}
\item[\astar{} (LM-cut, LM-count)] Two configurations of standard \astar{} search, one with the admissible LM-cut heuristic, and another with the inadmissible LM-count heuristic.

\item[GBF (LM-cut)] Greedy best-first search (i.e.\@ \astar{} with $g=0$ and no
  node re-opening), guided by the LM-cut heuristic.
  This configuration is included to determine how well a planner could perform when using \textit{only} LM-cut heuristic values.

\item[LAMA-2011 and LAMA-first]
  We compare against the state-of-the-art LAMA-2011 portfolio planner~\cite{richter2011lama}, as implemented in Fast Downward.
  We also compare against the first stage of the LAMA-2011 portfolio, which we label ``LAMA-first'' in experiments.
\end{description}

\subsubsection{Hardware}

To provide a fair comparison, all ASNet and baseline planner runs were executed on the same hardware.
Specifically, each run was restricted to a single core of an Intel Xeon Platinum 8175 processor attached to an Amazon AWS \verb|r5.12xlarge| instance, with 16GB of memory available per run.
We opted not to use GPUs or multiple CPU cores because doing so would have allowed ASNets to exploit parallelism that is not available to the sequential baseline planners.
In informal tests, we found that the filters in ASNets were too small to effectively exploit GPU resources.
This stands in contrast to large image CNNs, which are bottlenecked by large matrix multiplication and convolution operations that can be efficiently parallelised on GPU hardware.
However, ASNets could still utilise all available CPU cores during training if not restricted to a single core.

\subsection{Evaluation Domains}\label{ssec:expts-prob-doms}

This section describes the set of training and testing problems for each of our seven evaluation domains.
We will describe the Triangle Tireworld domain in additional detail because it appears again in \cref{sec:understand}.
We will also provide a detailed description of CosaNostra Pizza, which some readers may not be familiar with.
Detailed descriptions of the remaining domains are left to \cref{app:domain-descriptions}.

\begin{figure}[ht!]
    \begin{center}
      \begin{small}
        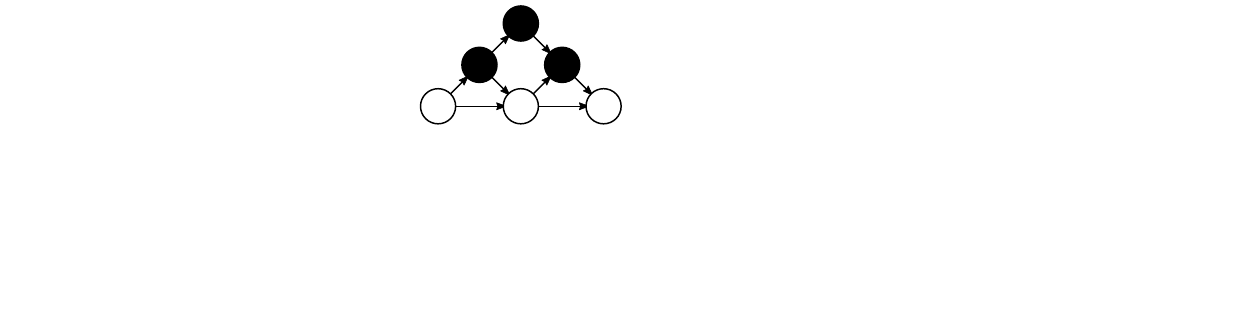
      \end{small}
    \end{center}
    \vspace{-1em}
    \caption{Illustration of the tire placement pattern in the first three Triangle Tireworld problems, adapted from \citeA{little2007probabilistic}.
    Black locations, including those around the outside edge, have spare tires; white locations do not.
    The road structure of larger instances is constructed recursively from that
    of smaller ones.}
    \label{fig:ttw-locs}
\end{figure}

\paragraph{Triangle Tireworld~\cite{little2007probabilistic}}
This is a navigation task in which the agent must move a vehicle from one corner of a triangular road network to another.
Whenever the vehicle moves from location $a$ to location $b$, there is a 50\% chance it will arrive in $b$ with a flat tire, in which case it must replace the damaged tire with a spare before moving again.
Unfortunately, only some locations have spare tires.
In particular, there are spare tires at every location along the outside edges of the triangle, but no tires on the inside edge of the triangle, and only tires at some locations in the interior.
The precise pattern of tire locations is illustrated in \cref{fig:ttw-locs}.
The policy with the highest probability of reaching the goal is to take the longest path from the start to the goal around the outside edges of the triangle, thereby ensuring that all visited locations have a spare tire.
\citeA{little2007probabilistic} introduced this domain to serve as a simple example of a ``probabilistically interesting'' planning task; that is, a task in which planners must intelligently account for risk.
Determinised heuristics ignore the risk of negative action outcomes, and thus encourage heuristic search to look for (risky) paths through the middle of the triangle.
We train Triangle Tireworld policies on three problems of sizes 1--3, and test on 17 problems of sizes 4--20.

\paragraph{CosaNostra Pizza~\cite{toyer2018action}}
CosaNostra Pizza is a simple probabilistic domain that has been designed to be challenging for current heuristic search planners.
The agent's task is to pick up a pizza from a shop, drive a vehicle to a customer's house, deliver the pizza, then drive back to the shop.
The shop and the customer's house both lie at opposite ends of a chain of locations.
At each location is a toll booth: the agent may either spend one time step paying the operator, or save a time step by driving through without paying.
However, if the agent chooses not to pay, then there is a 50\% chance that the toll booth operator will drop the boom gate and crush the agent's vehicle the next time they pass through that toll booth.
The optimal policy thus pays all toll operators on the path from the shop to the customer, thereby ensuring that the agent can return to the shop safely.
Paying operators twice is unnecessary, so an optimal policy will not pay any of them on the path back from the customer to the shop.
Like Triangle Tireworld, this domain poses a significant challenge to heuristic search planners that use determinised heuristics, since they ignore the risk of the toll operator crushing the agent's car.
This domain also poses a challenge to delete-relaxation heuristics: a delete-relaxation heuristic evaluated in the initial state (at the shop) will believe that upon delivering the pizza, the agent is both at the customer's house \textit{and} at the shop.
Thus, the heuristic will not recommend taking actions that could decrease the cost (or risk) of the return path.
We train CosaNostra Pizza policies on five problems with 1--5 toll booths, and test on 17 problems with 6--50 toll booths.

\paragraph{Probabilistic Blocksworld}
A simple probabilistic version of Blocksworld in which blocks may sometimes slip from the gripper onto the table.
Our version differs from the IPPC version~\cite{younes2004ppddl1} in that it lacks actions for moving towers of blocks, and can only move a single block at a time.
The modified version produces $n$-block instances with $O(n^2)$ ground actions, rather than $O(n^3)$ ground actions in the version with tower movement operators.
Without this change, we found that our grounding and network construction code was too memory-intensive to perform experiments on larger instances of 30+ blocks where heuristic search-based planners struggle.
We train Probabilistic Blocksworld policies on 25 problems with 5--9 blocks, and test on 30 problems with 15--40 blocks.

\paragraph{Exploding Blocksworld~\cite{younes2004ppddl1}}
A challenging probabilistic version of Blocksworld in which blocks can ``explode'', which leads to problems with both avoidable and unavoidable dead-ends.
This domain has been modified to remove a bug in which blocks could be stacked on
top of themselves.
We train on 24 problems with 5--9 blocks, and test on 30 problems with 11--20 blocks.

\paragraph{Gold Miner~\cite{fern2011first}}
A deterministic domain in which an agent must navigate through a grid to find gold, destroying obstacles along the way.
We train on 17 problems with sizes from $4 \times 4$ to $6 \times 6$ then test on 21 problems of size $7 \times 7$ to size $19 \times 19$.
Importantly, while we use the instance generator from the IPC 2008 Learning Track, we do \textit{not} use the same train or test instances.
We found that the original training instances did not provide an adequate curriculum for training: they were either too small to learn useful control knowledge (the bootstrap distribution) or too difficult to solve with heuristic search (the target distribution).
Likewise, we found that the test instances did not cover a wide enough range of difficulty to produce an informative comparison with heuristic search.
Instead, we generate a new training set in which problems are an appropriate size for our teacher planner, and a new test set that includes substantially larger instances which cannot all be solved within three hours.

\paragraph{Matching Blocksworld~\cite{fern2011first}}
A more challenging deterministic variant of the traditional Blocksworld domain with \textit{two} grippers, where incorrectly using one gripper instead of the other can lead to a dead end.
We train on 23 problems with 5--9 blocks each, then test on 30 problems with 15--60 blocks each.
Again, we use the instance generator from the IPC 2008 Learning Track, but generate a new training set and a test set with larger problems.

\paragraph{Blocksworld}
The standard deterministic Blocksworld domain.
We have included both deterministic Blocksworld and Probabilistic Blocksworld in our evaluation to demonstrate that ASNets can obtain good coverage on a domain regardless of whether it has any probabilistic elements.
In \cref{ssec:expts-bw}, we also respond to the ``Blocksworld challenge'' of \citeA{geffner2018model} by showing that ASNets can generalise extremely well to large Blocksworld instances.
We train on 25 problems with 8--10 blocks, and test on 30 problems with either 35 or 50 blocks each.

\subsection{Results of Time-Based Evaluation}\label{ssec:expts-main-results}

\begin{figure}
    \centering
    \includegraphics[height=0.48\textwidth]{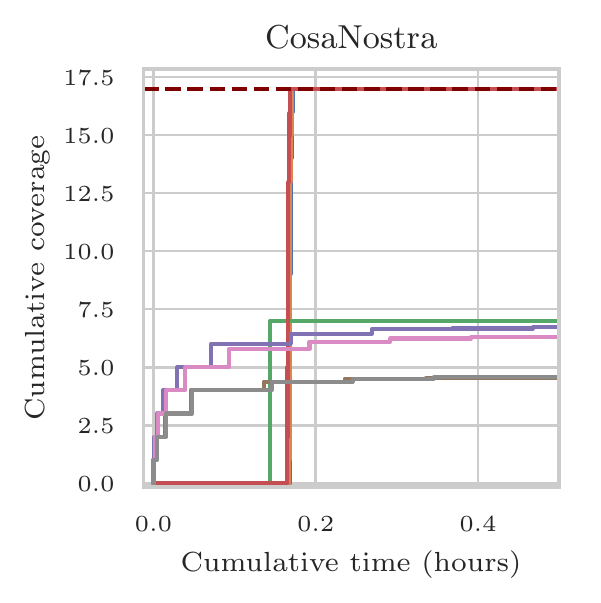}
    \includegraphics[height=0.48\textwidth,trim={6mm 0 0 0},clip]{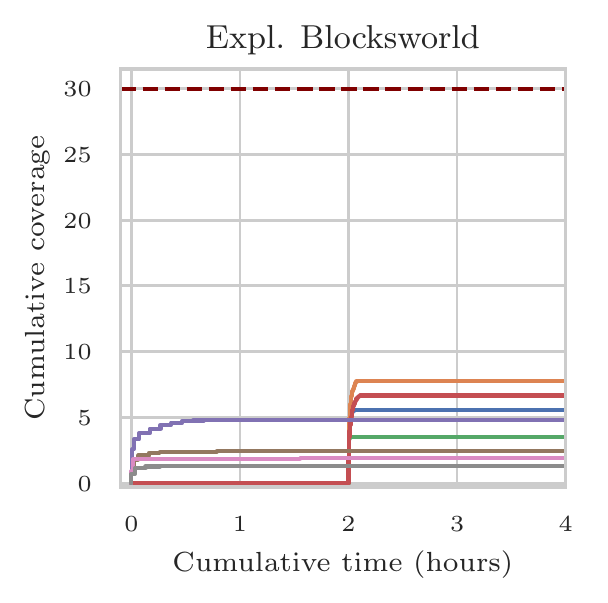}\\
    \includegraphics[height=0.48\textwidth]{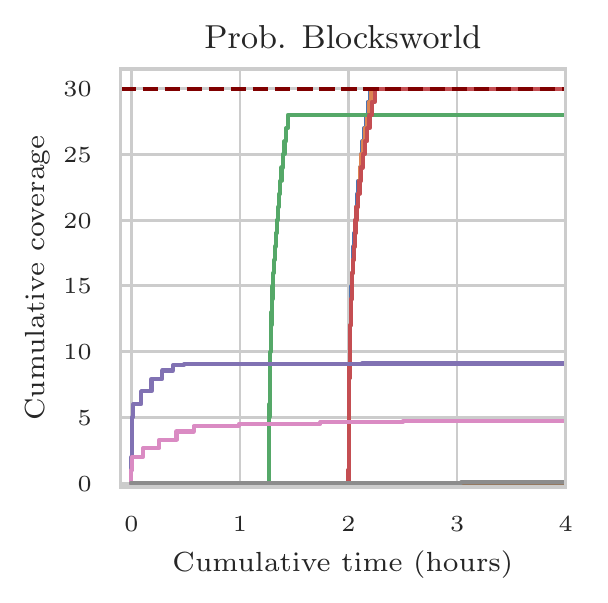}
    \includegraphics[height=0.48\textwidth,trim={6mm 0 0 0},clip]{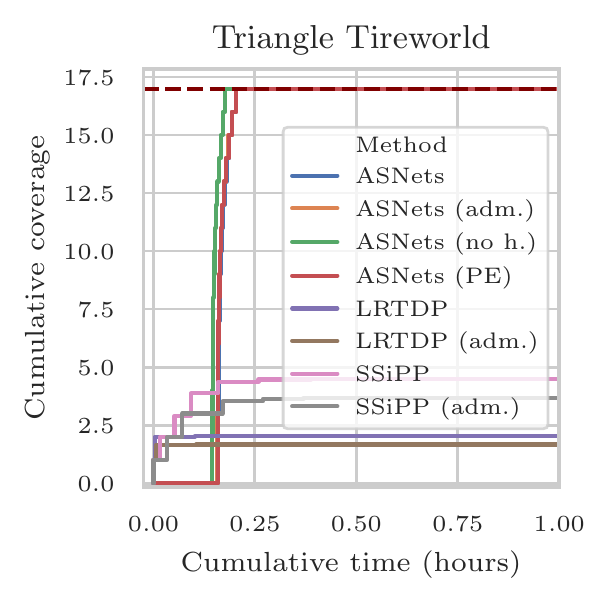}
    \caption{Cumulative success rate of rollouts across a set of test problems for each probabilistic domain.
      All ASNet configurations except the configuration without heuristic input (``no h.'') asymptotically achieve the same (perfect) coverage in CosaNostra, Triangle Tireworld, and Probabilistic Blocksworld.
      Configurations that appear to be missing are simply occluded by other, similar configurations.
      Refer to main text for interpretation of plotted quantities.}
    \label{fig:time-trend-prob}
\end{figure}

\begin{figure}
    \centering
    \includegraphics[height=0.48\textwidth,trim={1mm 0 3mm 0},clip]{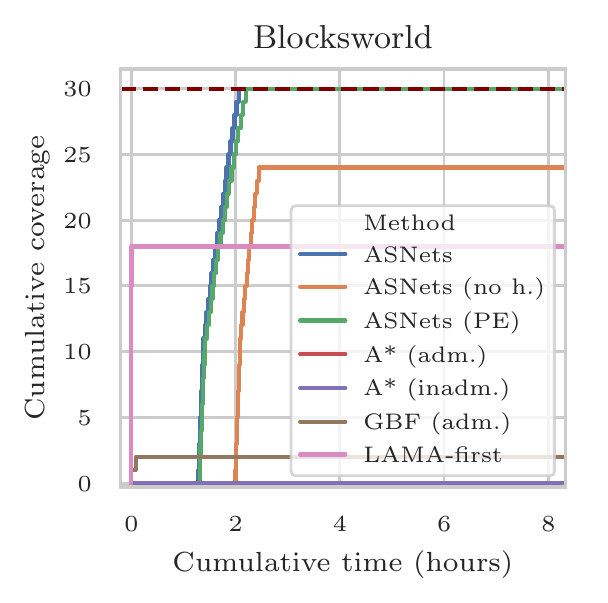}
    \includegraphics[height=0.48\textwidth,trim={6mm 0 3mm 0},clip]{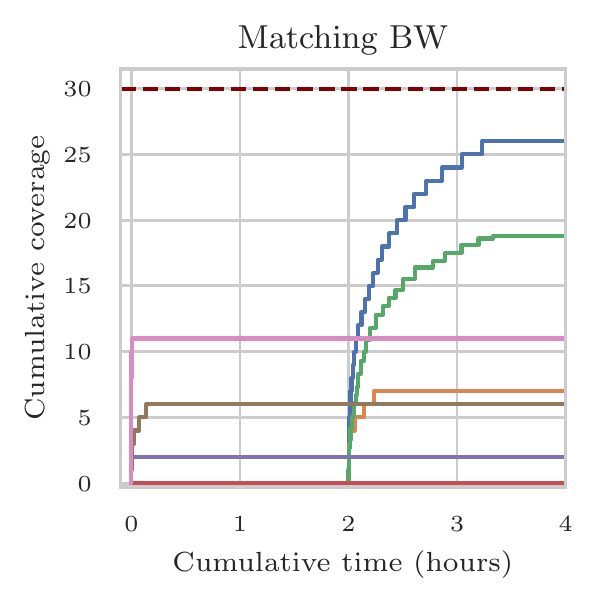}
    \includegraphics[height=0.48\textwidth,trim={6mm 0 3mm 0},clip]{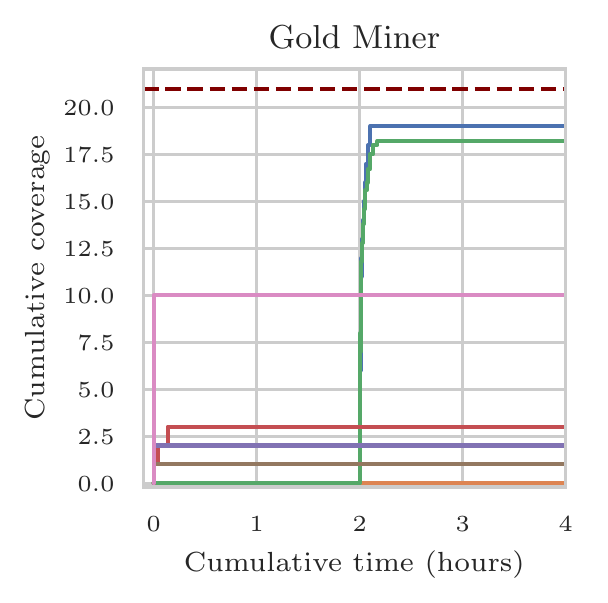}
    \caption{Cumulative number of problems solved over time for deterministic
      test problems.
      Refer to main text for description of plotted quantities.
      LAMA-2011 is omitted: it achieved the same coverage as LAMA-first on our evaluation problems, but took much more time to execute on each problem because it continued to evaluate improved planners after finishing its first planning stage.
      As a result, the LAMA-2011 curve for cumulative coverage over time was dominated by the corresponding curve for LAMA-first.
      LAMA-2011 results are instead presented in tabular form in \cref{app:results-tables}.
     }
    \label{fig:time-trend-det}
\end{figure}

We present our main results as cumulative coverage graphs in
\cref{fig:time-trend-prob} (probabilistic domains) and \cref{fig:time-trend-det}
(deterministic domains).
Specifically, for each domain, we plot the cumulative fraction of evaluation rollouts that have reached a goal state, summed over the total number of problems seen so far.
The fraction of evaluation rollouts that have reached the goal for a specific problem is always a number between 0 and 1, so the sum of those fractions over all $n$ evaluation problems will be a value between 0 and $n$.
For instance, say that after 5 minutes, a given planner has reached the goal on 30/30 runs on one problem, and on 15/30 runs on another problem, but has not finished planning on any other problems.
As a result, its cumulative coverage at 5 minutes would be $30 /30 + 15/30 + 0 = 1.5$.
For deterministic domains, both the ASNets and the baselines generally only do one run per problem, so each problem contributes either 1 (solved) or 0 (not solved) to the total cumulative coverage.
The exceptions are the stochastic rollouts marked ``ASNets (PE)'' where several trajectories are sampled from the same policy.
Those runs are plotted in the same way as for evaluation rollouts on probabilistic domains, where each of the $k$ runs on a single problem contributes $1/k$ to the total cumulative coverage if it is successful, and 0 otherwise.
The number of test instances for each domain is plotted with a red, dashed horizontal line.
Note that we include the time taken to train ASNets in these plots, so all the ASNet runs have zero cumulative coverage during the training period of up to two hours, after which cumulative coverage increases as the ASNets are evaluated on problems.

In addition to cumulative coverage plots, we have included detailed coverage and solution cost tables in \cref{app:results-tables}.
There is one table per domain.
Each row of a table corresponds to a different problem in the test set, while each column corresponds to a different planner or ASNet configuration.
For probabilistic domains, each cell shows the fraction of planner evaluations that reached the goal (e.g.\@ 29/30) within the allotted total time limit for the problem.
Each cell also shows the mean cost of trajectories that reached the goal, along with bounds for a 95\% confidence interval (e.g.\@ ``25.2 $\pm$ 0.4'').
Cells with a ``-'' indicate that the corresponding planner did not produce \textit{any} trajectories that reached the goal within the allotted time.
For deterministic domains, we generally only compute one plan (i.e.\@ rollout) per planner, and so we only report the cost of that plan, or a ``-'' if the planner could not solve the problem within the allotted time.
The exceptions are the ``ASNet (PE)'' runs, where we report the mean cost of stochastic rollouts that did reach the goal, along with the fraction of trajectories that reached the goal in parentheses (e.g. ``7/10'').
Our discussion here will focus on the cumulative coverage plots in \cref{fig:time-trend-prob} and \cref{fig:time-trend-det}, although we will occasionally highlight portions of cost tables in \cref{app:results-tables}.

\subsubsection{The Big Question: When is it Worth Training an ASNet?}

Across the plots for from \cref{fig:time-trend-prob} and \cref{fig:time-trend-det}, we see a common pattern.
Initially, the baseline planners rapidly solve a collection of small instances, while the ASNets remain at zero coverage because they are training.
However, once the ASNets are trained, they quickly surpass the baselines in cumulative coverage because they can be rapidly evaluated on test problems.
Further, the performance of the ASNet runs consistently plateaus at a higher level than the baselines.
From the cost tables in \cref{app:results-tables}, it can be seen that in most domains, this difference in coverage arises from the fact that ASNets can solve some very large problems that the baseline planners cannot solve within the three-hour cutoff.
We will discuss detailed results for each domain in the next section, but the upshot is that it's worthwhile to train an ASNet in most of these domains if you need to solve a single very large problem, or a collection of moderately large problems.

\subsubsection{Results Organised by Domain}\label{ssec:expts-by-domain}

\paragraph{CosaNostra Pizza}
Results for CosaNostra Pizza are presented in \cref{fig:time-trend-prob}, \cref{tab:prob-res-cn-1} and \cref{tab:prob-res-cn-2} (\cref{app:results-tables}).
All four ASNets learn the same (optimal) policy of paying the toll booth operators on the way to the customer's home, then driving through the toll booths without paying on the way back.
The baseline planners can solve some test instances, but cannot solve instances with more than 14 toll booths, since all of them are guided by determinisation-based heuristics that ignore the possibility of the toll booth operator damaging the vehicle on the way back from the customer.
This domain represents a favourable case for ASNets: not only was it constructed to be particularly difficult for planners with determinising heuristics, but it also has a simple trick (pay the operator on the way to the customer) that makes all instances trivial.

\paragraph{Exploding Blocksworld}
Results for Exploding Blocksworld are presented in \cref{fig:time-trend-prob}, \cref{tab:prob-res-exbw-1} and \cref{tab:prob-res-exbw-2} (\cref{app:results-tables}).
Because Exploding Blocksworld problems generally contain unavoidable dead ends,
we do not know the theoretical maximum cumulative coverage for our test set.
The thick, dotted red line at the top of the plot in \cref{fig:time-trend-prob} simply shows the number of test instances, which is a loose upper bound on the maximum cumulative coverage.
Nevertheless, we can see that the ASNet configured to use an optimal teacher (LRTDP with the admissible LM-cut heuristic) obtains substantially higher cumulative coverage than the baselines.
\cref{tab:prob-res-exbw-1} and \cref{tab:prob-res-exbw-2} show how this difference in cumulative coverage arises.
The baselines manage to find reliable solutions for a handful of test problems, and the high proportion of successful rollouts on those problems account for most of their total cumulative coverage.
In contrast, the ASNet policies are generally less reliable on problems where the baselines do well, but still manage to produce a few goal-reaching trajectories on difficult test problems where the baselines fail entirely.
This suggests that ASNets may not have learnt to fully exploit domain-specific tricks (like intentionally detonating explosive blocks in order to prevent them from damaging other blocks), but have still learnt a policy that is good enough to partly solve a wide range of problems.
Given enough planning time, the baseline planners can obviously ``discover'' all the tricks required for a reliable policy on one specific instance, but their inability to transfer knowledge between problems means that they must discover all of those tricks anew on each test instance.



\paragraph{Probabilistic Blocksworld}
Results for Probabilistic Blocksworld are presented in \cref{fig:time-trend-prob}, \cref{tab:prob-res-pbw-1} and \cref{tab:prob-res-pbw-2} (\cref{app:results-tables}).
Here three of the ASNet configurations yield similar policies with perfect coverage on the test set.
The largest test instances have 40 blocks, but the baselines struggle to scale beyond 25 blocks.
On the training instances we found that LRTDP with LM-cut only visited 20\% fewer states on average than LRTDP with the zero heuristic, and actually took more wall time because of the overhead of heuristic evaluation.
It is therefore unsurprising to see that it does not solve any of the test instances.
\cref{tab:prob-res-pbw-2} shows that while SSiPP sometimes produces goal-reaching trajectories on larger problems, it never does so within the three-hour limit on more than 1/30 rollouts, and its solution costs are generally poor (e.g.\@ 583 actions for one of the 35--block instances, where the ASNets require slightly over 120 on average).
The ASNet configuration with no heuristic information does surprisingly well: it only fails on two test instances, where it eventually gets stuck repeatedly picking up and putting down the same block.
We posit that this may be because towers with misplaced blocks usually have pairs of blocks near the top that are not meant to be on top of one another in the goal.\footnote{
  This is not an exotic property of the instance generator, but a consequence of the fact that few blocks in a random initial state will happen to sit on the same block they are meant to sit on in a different, random goal state.
}
Thus, the network can unstack those towers even though the receptive field limitation prevents it from ``seeing'' all of the blocks in the tower.

\paragraph{Triangle Tireworld}
Results for Triangle Tireworld are presented in \cref{fig:time-trend-prob}, \cref{tab:prob-res-ttw-1} and \cref{tab:prob-res-ttw-2} (\cref{app:results-tables}).
As with CosaNostra Pizza, all four ASNet configurations learn the same optimal policy of navigating around the outside edge of the triangle to the goal, changing tires when necessary.
In contrast, the baselines struggle to solve instances beyond size 13 because their determinising heuristics neglect to account for the possibility of losing a tire at a location with no spare.
On the training instances, we found that LRTDP with h-add or LM-cut failed to significantly outperform LRTDP with the zero heuristic in terms of number of states visited.
We also found that LRTDP with LM-cut took slightly more wall time than blind search because of the overhead of heuristic evaluation.
Only SSiPP solves the instances beyond size 5, and that is likely because its fixed-depth lookahead strategy is particularly well-suited to Triangle Tireworld.
Like CosaNostra, this domain is particularly favourable to ASNets, since it is deliberately constructed to be difficult for planners that ignore stochastic transitions, but can easily be solved with a simple domain-specific strategy.
Specifically, a policy can follow the edge of the triangle by repeatedly moving from one location to an adjacent location that has a spare tire, and which is also adjacent to a third location with a spare tire.

\paragraph{Deterministic Blocksworld}
Results for Deterministic Blocksworld are presented in \cref{fig:time-trend-det} and \cref{tab:det-res-bw} (\cref{app:results-tables}).
Again, we see a similar story to Probabilistic Blocksworld: the three ASNet configurations each manage to solve all or most of the test problems, while the baselines struggle with the larger (50-block) problems.
\astar{} does not solve any of the test problems, all of which are relatively large (35+ blocks), while GBF solves only two problems.
LAMA-2011 (in \cref{tab:det-res-bw}) and LAMA-first solve all of the 35-block problems and some of the 50-block problems, but tend to produce much longer plans than the ASNets.
For example, there are five 50-block problems where LAMA baselines both require 200-286 actions to reach the goal, but the ASNets need only 134-164 actions.

\paragraph{Matching Blocksworld}
Results for Matching Blocksworld are presented in \cref{fig:time-trend-det} and \cref{tab:det-res-mbw} (\cref{app:results-tables}).
Matching Blocksworld is much more challenging for the baseline planners: while some of them can quickly solve the small instances (particularly those with 15--25 blocks), they do not solve many of the larger instances.
In contrast, the best ASNet configuration manages to solve most instances, including one of the three instances with 60 blocks.
However, it is still somewhat surprising that ASNets do not achieve higher coverage.
When tuning hyperparameters, we found that performance on Matching Blocksworld was quite sensitive to the choice of $\ell_2$ regularisation coefficient and dropout, and that with the best combination of regularisation (around $10^{-4}$) and dropout (around 0.25) we could generally obtain higher test coverage.
The same hyperparameters did not transfer well to Exploding and Probabilistic Blocksworld, so we did not use them for the final evaluation.
Nevertheless, this suggests that ASNets' limited generalisation on Matching Blocksworld is a consequence of our training strategy and hyperparameters, rather than a fundamental representational limitation.

\paragraph{Gold Miner}
Results for Gold Miner are presented in \cref{fig:time-trend-det} and \cref{tab:det-res-gm} (\cref{app:results-tables}).
As with Matching Blocksworld, we find that ASNets manage to solve almost all test problems when equipped with heuristic input features, while the baselines fail to solve the larger instances featuring 13x13 to 19x19 grids.
The solutions produced by LAMA-2011 and LAMA-first for the larger instances also tend to be quite inefficient; in the case of LAMA-2011, this may be because only the first planner in the portfolio (LAMA-first) manages to finish within the three allotted hours.
We did not find an obvious pattern in the executed trajectories on the instances where ASNets failed to reach the goal.
Further, as with Matching Blocksworld, we found that alternative hyperparameter settings led to policies that would solve all test instances.
Again, this suggests that the limited generalisation of ASNets on this domain is due to a flaw of our training strategy, as opposed to some fundamental representational limitation.

\subsubsection{Other Discussion Questions}\label{ssec:other-discuss}

\paragraph{Are ASNets performing fixed-depth lookahead in state space?}
No.
As noted in \cref{ssec:heur-inputs}, ASNets' receptive field limitation is superficially similar to the limitations of short-sighted probabilistic planning or model-predictive control, both of which choose actions by solving a series of short-horizon sub-problems.
However, the receptive field of an ASNet is a consequence of the relatedness of actions and propositions, rather than of limited lookahead in state space.
Our comparison with SSiPP in the probabilistic domains experimentally
demonstrates the difference.
In all four domains, SSiPP's strategy of repeatedly solving fixed-depth sub-problems is able to solve some instances.
However, its coverage always plateaus well before that of ASNets, as we might expect given the substantial differences between the two strategies.

\paragraph{Do slightly sub-optimal teacher planners lead to worse ASNet policies?}
In two domains (Exploding Blocksworld and Probabilistic Blocksworld), our results show that slight suboptimality of teaching data leads to slight suboptimality of ASNet policies.
In our experiments on probabilistic domains, we performed ASNet runs with both an optimal teacher (LRTDP with the LM-cut heuristic) and a potentially-suboptimal teacher (LRTDP with h-add).
On the training problems for Triangle Tireworld and CosaNostra Pizza, we found that LRTDP with h-add always returned optimal solutions after a similar wall-time duration to LRTDP with LM-cut.
Our results confirm that there is indeed no gap between ASNet policies trained by LRTDP with h-add and ASNet policies trained by LRTDP with LM-cut, either in terms of either coverage or mean solution cost.
On the training problems for Exploding Blocksworld and Probabilistic Blocksworld, we found that LRTDP with h-add returned solutions that were slightly (1\% to 2\%) more costly than those for LM-cut.
Consequently, we see that the cost of trained ASNet policies becomes fractionally higher in these domains when using a h-add teacher planner rather than an LM-cut teacher planner.
It's worth noting that we did obtain better coverage on Exploding Blocksworld when using an optimal planner instead of a sub-optimal planner.
However, when training with different ASNet hyperparameters, we found that this relationship sometimes reversed, and the ASNet with a sub-optimal teacher would occasionally do better than the one with an optimal teacher (and both would do better than the baselines).
For this reason, we suspect that the better coverage of the ASNet trained with a sub-optimal teacher is due to an interaction between choice of planner and other hyperparameters, rather than some intrinsic advantage of training with optimal planners.
For deterministic domains, we simply used a sub-optimal teacher for all runs, as we did not find that an optimal planner lead to increased coverage.

\paragraph{Are heuristic input features always necessary?}
Not always, although they are important in most of the domains that we evaluated on.
For simple domains like Triangle Tireworld, it is straightforward to train correct generalised policies without any heuristic features at all, as we discuss in \cref{sec:understand}.
Heuristic features were helpful for more complex domains, however, as shown by the poor performance of ASNets without heuristic inputs on blocksworld-type domains, on Gold Miner and on CosaNostra.
In all of these domains, it's easy to see why that's the case: for example, in blocksworld-type problems, the blocks at the top and bottom of a long tower will not be in the receptive field for any one enabled action, and so heuristic inputs are necessary to determine whether the tower contains any misplaced blocks.
In CosaNostra, it's not always possible to tell which direction leads to the customer versus to the shop without the use of heuristic inputs, while in Gold Miner the agent cannot ``see'' the location of the gold without heuristic inputs.

\paragraph{Are LM-cut heuristic values sufficient to trivially solve our test tasks?}
No.
The most obvious evidence of this is the limited performance of the GBF planner equipped with LM-cut in the deterministic domains of \cref{fig:time-trend-det}.
Although it can solve some test problems easily, it cannot solve all of them.
Likewise, the \astar{}, LRTDP, and SSiPP baselines equipped with LM-cut never managed to exceed the final coverage of the ASNets.
This shows that LM-cut heuristic values alone are not sufficient to solve these problems, and also suggests that the LM-cut-based input features given to the ASNet are likely not sufficient to solve those problems alone either.

\paragraph{How well do ASNets perform on the remaining IPC 2008 Learning Track domains?}
There are four domains from the IPC 2008 Learning Track that we have not presented results for: $n$-Puzzle, Thoughtful (solitaire), Parking, and Sokoban.
In all four cases, we could not obtain an ASNet policy that both achieved high coverage on the test problems and managed to scale better than classical baseline planners.

In Thoughtful and Parking, ASNets were stymied by the large number of ground actions and propositions in some instances from the original test set.
Both domains include action schemas with relatively high arity: Parking is a blocksworld-like problem with ``direct'' movement actions that take three blocks as parameters, while Thoughtful is a complex card game in which some action schemas have up to seven parameters.
As a result, the number of actions and propositions increases rapidly as a function of the number of objects in a problem.
For example, some Parking instances from the original IPC 2008 test set include over 250,000 ground actions.
As a result, we found that it was too slow to construct and perform forward propagation on an ASNet for the larger problems from either domain.
This limitation is likely to arise in any problem with a very large number of actions and propositions, and is a consequence of the fact that we need to evaluate one or more action and proposition modules for every ground action and ground proposition in the problem.
This is a significant limitation of the ASNet architecture, and we leave investigation of alternative architectures that scale more gracefully with action count to future work. 

We also failed to find a reliable generalised policy for $n$-Puzzle.
In principle, it may be possible to find an efficient generalised policy with ASNets, since tractable suboptimal policies do exist~\cite{ratner1986finding}.
While ASNets had little trouble learning to solve training instances from the ``bootstrap'' training set from IPC 2008, the resulting policies did not generalise to larger problems.
This may simply be a consequence of not using large enough training instances, or sufficiently many instances; we leave further experiments for future work.

In Sokoban, we simply could not train a reliable ASNet policy under the evaluation conditions described in \cref{ssec:expts-main}.
This is not terribly surprising: Sokoban is PSPACE-complete~\cite{culberson1997sokoban}, and so it is unlikely that any polynomial time generalised policy exists.
\citeA{groshev2018learning} show that it is possible to train an ordinary convolutional neural network to solve instances of a limited class of Sokoban problems on up to $18\times 18$ grids.
However, their training set consists of 45,000 instances, including 9,000 different obstacle configurations and five combinations of initial state and goal per obstacle configuration.
Solving all instances in such a large training set would be very expensive, and thus erode the advantage of training a generalised policy instead of repeatedly invoking a heuristic search planner.
We found it difficult to train with more than 10--20 instances while remaining under our self-imposed two-hour training limit.
In future work, it would be interesting to evaluate ASNets on this domain with much more training time and a much larger number of training instances.

\subsection{Additional Ablations}\label{ssec:asnet-ablations}

\begin{table}
  \begin{center}
    \begin{small}
        \begin{tabular}{lccccccc}
          \toprule
          Configuration & CN & ExBW & PBW & TTW & BW & GM & MBW\\
          \midrule
          Default & \textbf{17.0/17} & 5.5/30 & \textbf{30.0/30} & \textbf{17.0/17} & \textbf{30.0/30} & 19.0/21 & 26.0/30\\
          Old-style pooling & \textbf{17.0/17} & 4.5/30 & \textbf{30.0/30} & \textbf{17.0/17} & \textbf{30.0/30} & \textbf{21.0/21} & 28.0/30\\
          No skip conn. & \textbf{17.0/17} & 2.9/30 & \textbf{30.0/30} & \textbf{17.0/17} & \textbf{30.0/30} & 18.0/21 & \textbf{30.0/30}\\
          One layer & \textbf{17.0/17} & 1.1/30 & 17.5/30 & \textbf{17.0/17} & 0.0/30 & \textbf{21.0/21} & 5.0/30\\
          Three layers & \textbf{17.0/17} & \textbf{7.1/30} & \textbf{30.0/30} & \textbf{17.0/17} & \textbf{30.0/30} & 19.0/21 & \textbf{30.0/30}\\
          No history & \textbf{17.0/17} & 4.5/30 & \textbf{30.0/30} & \textbf{17.0/17} & \textbf{30.0/30} & 5.0/21 & 26.0/30\\
          No LM-cut & \textbf{17.0/17} & 5.2/30 & \textbf{30.0/30} & \textbf{17.0/17} & \textbf{30.0/30} & 13.0/21 & 25.0/30\\
          No LM-cut/hist. & 7.0/17 & 3.5/30 & 28.0/30 & \textbf{17.0/17} & 24.0/30 & 0.0/21 & 7.0/30\\
          \bottomrule
        \end{tabular}
    \end{small}
  \end{center}
  \caption{
    Additional ablations for ASNets on our seven test domains.
    Domains are CosaNostra Pizza (CN), Exploding Blocksworld (ExBW),
    Probabilistic Blocksworld (PBW), Triangle Tireworld (TTW), deterministic
    Blocksworld (BW), Gold Miner (GM), Matching Blocksworld (MBW).
    Boldface indicates that corresponding planner achieved (equal) best coverage for the domain in the corresponding column.
  }\label{tab:ablations}
\end{table}

\cref{tab:ablations} shows the coverage achieved by ASNet configurations which have had various features disabled or modified.
Each column corresponds to one domain, and each row to one ASNet configuration.
Each cell shows the total cumulative coverage for one configuration on a given domain, along with the number of problems in the domain.
If we were to plot the results for these ablations in the style of \cref{fig:time-trend-prob} and \cref{fig:time-trend-det}, then the total cumulative coverage would be the value that the cumulative coverage curve plateaus to in the limit.
The default configuration is the one marked ``ASNets'' in the previous subsection: a network with two (proposition) layers, 16 hidden units, and a teacher guided by the h-add heuristic (LRTDP for probabilistic domains, \astar{} for deterministic domains).
The other lines in \cref{tab:ablations} correspond, respectively, to the following modifications over the default:
\begin{itemize}
  \item Replace the new pooling mechanism from \cref{ssec:prop-layers} with the previous pooling mechanism from \citeA{toyer2018action}.
  \item Remove skip connections from the network.
  \item Go from two proposition layers ($L=2$) down to one ($L=1$).
  \item Go from two proposition layers up to three ($L=3$).
  \item Remove ``history'' input features that count the number of times each action has been executed so far (but not LM-cut features).
  \item Remove LM-cut input features (but not history features).
  \item Remove both LM-cut and history features.
\end{itemize}
All networks were trained using the same procedure and on the same hardware as the original time-based evaluation.

We can draw a number of inferences from these ablations.
First, we see that the model is relatively insensitive to the choice of pooling mechanism and use of skip connections.
The exceptions are Exploding Blocksworld (where skip connections and new-style pooling both increase coverage), Gold Miner (where new-style pooling reduces coverage), Matching Blocksworld (where the inclusion of skip connections reduces coverage).
The fact that deeper ASNets do better on Exploding Blocksworld suggests that the domain benefits from increased network capacity, and both skip connections and improved pooling may improve performance by increasing network capacity.
The reason why the new pooling mechanism decreases coverage on Gold Miner is less clear, as is the reason why skip connections decrease performance on Matching Blocksworld.
Our results for the one-layer network indicate that depth is unimportant for the simplest problems (CosaNostra, Triangle Tireworld), but matters a great deal for more complex blocksworld-type problems, where the deepest (three-layer) network achieves equal or better coverage than the default configuration.
Interestingly, the results for Gold Miner \textit{improve} with fewer layers.
We speculate that some form of overfitting or additional optimisation difficulties associated with larger networks might be behind this phenomenon.
It's also worth noting that the three-layer network achieves equal or better coverage than the two-layer network across all domains.
However, the three-layer network is slower to train and evaluate than the two-layer network as it requires around 50\% more time per network evaluation, so we opted to use the two-layer network in the cumulative coverage experiments of the preceding section.
Finally, we see that LM-cut landmarks and history features are each responsible for some improvement in coverage on at least one domain.
It should be emphasised that because \cref{tab:ablations} evaluates over tens of problem instances rather than thousands, it does not reflect changes in coverage on the far tails of the instance distribution.
When performing large-scale experiments on thousands of blocksworld instances in \cref{ssec:expts-bw}, we observed that landmark features were in fact essential to getting from good generalisation (i.e.\@ above 99\%) to near-perfect generalisation.

\subsection{Extended Experiments on Deterministic Blocksworld}\label{ssec:expts-bw}

\begin{table}[t]
  \begin{center}
  {
    \begin{tabular}{@{}l@{}c@{}c@{}c@{}c@{\phantom{i}}c@{\phantom{i}}c@{\phantom{i}}c@{\phantom{i}}c@{}}
    \toprule
    Blocks & \multicolumn{1}{c}{\begin{small} 18 \end{small}} & \phantom{i} & \begin{small}   25 \end{small} & \phantom{i} & \begin{small} 35 \end{small} & \phantom{i} & \multicolumn{2}{c}{\begin{small} 50 \end{small}} \\
    \cmidrule{2-2} \cmidrule{4-4} \cmidrule{6-6} \cmidrule{8-9}
    Towers & \begin{small} 1--18 (6 sizes) \end{small} & & \begin{small} Rand. \end{small} & & \begin{small} Rand. \end{small} & & \begin{small} 1--40 (9 sizes) \end{small} & \begin{small}
Rand. \end{small}\\
    \midrule
    Instances & \begin{small} 9,000 \end{small} & & \begin{small}  100 \end{small} & & \begin{small}  100 \end{small} & & \begin{small}  9,000 \end{small} & \begin{small}  100 \end{small}\\
    \bottomrule
    \end{tabular}
    }
  \end{center}
  \caption{
    Distribution of block and tower counts used for evaluating the trained ASNets.
    The ``Rand.'' instances were sampled uniformly from the space of all instances with the corresponding number of blocks, without fixing a certain tower count.
    Other instances were sampled uniformly from the space of all instances with a specific number of blocks \textit{and} towers.
  }
  \label{tab:problem-sizes}
\end{table}



Recently, \citeA{geffner2018model} has argued that a key limitation of learning-based techniques in AI is their inability to reliably generalise to arbitrary instances of problems from a given class.
Not only are some neural network architectures only capable of processing fixed-size vectors of input, but they also tend to be evaluated in settings where low coverage and small evaluation problems are considered acceptable.
In the context of Blocksworld, this means that learning-based planning systems sometimes demonstrate ``a coverage of 68\% on selected instances with seven blocks'', when they should be able to demonstrate ``near 100\% coverage on arbitrary instances''~\cite{geffner2018ijcai}.
ASNets (and graph convolutional networks more generally) represent one possible way of side-stepping the issue of fixed input size.
In this section, we perform extended experiments on deterministic Blocksworld which demonstrate that ASNets can achieve very high coverage (``near 100\%'') on large (35--50 block) instances, even after training on only a few relatively small instances.

For these experiments we modified the hyperparameters from the previous section, using two-layer networks with 20 channels (instead of 16) in each action and proposition module.
We used 50 training problems with 8--10 blocks instead of 25 problems, and trained our network for six hours instead of two hours.
We also increased dropout to 0.3 and shifted the total number of ASNet rollouts per epoch from 70 to 150, both of which encouraged the ASNet to explore a wider variety of states at training time.
The distribution of block and tower counts in our test set is shown in \cref{tab:problem-sizes}.
Our test set includes 300 instances sampled uniformly at random with 25, 35, or 50 blocks each.
It also includes 9,000 instances of 18 blocks constrained to have 1, 3, 5, 10, 15, or 18 towers, as well as 9,000 instances of 50 blocks constrained to have 1, 4, 7, 10, 15, 20, 25, 30 or 40 towers.
These were generated using BWSTATES and BWKSTATES \cite{slaney2001blocks}.
Since the hardness of optimal planning in Blocksworld is highly correlated with the number of towers in the initial and goal states, constraining the number of towers allows us to consider problems with varying structure and hardness level. 
%

We found that ASNets could solve all 18,300 test instances after training using the hyperparameters described above.
Although this strong empirical result is not proof of the policy's ability to generalise to \textit{all} problems in the domain, it nevertheless underscores the ability of ASNets to solve non-trivial planning tasks with very high reliability when given a modest number of training instances.
It's worth noting that when using the same hyperparameters and training set as \cref{ssec:expts-main}, ASNets would sometimes fail at a small proportion of instances in our extended test set.
For example, due to the receptive field limitation, some trained ASNets would
get stuck in loops when the top few blocks in all towers appeared to be sitting
immediately on top of the correct block, but blocks further down each tower were
not in-position.
The additional instances used to train the ASNets described in this section likely made them more robust to rare edge cases of that kind.

\section{Understanding ASNet Policies}
\label{sec:understand}

Like other kinds of neural networks, ASNets suffer from a lack of interpretability, and in particular a lack of \textit{transparency}: it is difficult for a human to understand what the network is doing internally.
This is because ASNets have far too many parameters and activations for a human to keep track of.
For instance, the main Triangle Tireworld policy described in \cref{ssec:expts-main} has 7,634 parameters, and an instantiated ASNet for a large Triangle Tireworld problem will also have tens of thousands of internal activations.
Ideally, we would instead like trained ASNets to have the property of \textit{simulability}, in the terminology of \citeA{lipton2016mythos}: a person should be able to mentally simulate the expected behaviour of the policy, and anticipate scenarios in which it may or may not work.
In this section, we show that simple sparsity regularisation can yield ASNet policies that have very few parameters, and thus can be simulated in one's head.
Although we do not consider formal verification, we note that the model simplification techniques discussed here would likely make it easier to compile ASNets down into a format that could be automatically verified to be correct with respect to some formal specification of the domain (perhaps expressed as a MILP or SMT problem).
This kind of reasoning is an active area of research in the deep learning community~\cite{katz2017reluplex,tjeng2019evaluating}.

\subsection{Sparsity Through Regularisation}

To train sparse ASNets, we will use two ingredients: an appropriate regulariser, and an appropriate training strategy.
It is well known in the machine learning and statistics communities that an
$\ell_1$ regulariser tends to lead to sparse model weights~\cite{tibshirani1996regression}.
For the purpose of this section, we removed the $\ell_2$ and dropout regularisers, then added an $\ell_1$ penalty $\gamma \|\theta\|_1$ to the batch objective in \cref{eqn:batch-objective} with $\gamma = 10^{-2}$.
We also trained the ASNet for eight hours instead of two, and decayed the learning rate from $10^{-2}$ to $10^{-3}$ (after 30 epochs) to $10^{-4}$ (after 40 epochs), rather than keeping the learning rate fixed at $10^{-3}$.
Together, these changes led to an extremely sparse ASNet for Triangle Tireworld that we will present in the next section.
We also repeated the same experiment for CosaNostra Pizza, but we will defer discussion of the resulting policy to \cref{app:interp-cn}.

\subsection{A Sparse Policy for Triangle Tireworld}

\begin{figure}[ht!]
\centering
    %
    %
    %
    %
    %
%
%
    \begin{footnotesize}
      \begin{tabularx}{0.8\linewidth}{r@{\thinspace}c@{\thinspace}l}
        \toprule
        \multicolumn{3}{X}{\textit{First action layer}: simply gives large
          constant activation to $\maschema{changetire}$ modules.\vspace{4pt}}\\
        \qquad $\hidact{\mactn{changetire}{?loc}}{1}$ & $=$  & $15.76$\\
        \midrule
        \multicolumn{3}{X}{\textit{First proposition layer}: $\mpred{vehicle-at}$ modules assign high value to
          locations where tires can be changed, and low value to others:\vspace{4pt}}\\
        \qquad $\hidprop{\mpropn{vehicle-at}{?loc}}{1}$ & $=$ & $0.81 \cdot \pool\left(\hidact{\mactn{changetire}{?loc}}{1}\right) + 0.02$\\
        \midrule
        \multicolumn{3}{X}{\textit{Second action layer}: $\maschema{move-car}$ modules have high activation
          when the destination has a tire, and low activation otherwise.\vspace{4pt}}\\
        \qquad $\hidact{\mactn{move-car}{?from,?to}}{2}$ & $=$ & $0.46 \cdot \hidprop{\mpropn{vehicle-at}{?to}}{1}$\\
        \midrule
        \multicolumn{3}{X}{\textit{Second proposition layer}: $\maschema{vehicle-at}$
          modules have highest activation when $\mobj{?loc}$ has a tire,
          and there is also a move action to an adjacent location with a
          tire.\vspace{4pt}}\\
          \qquad $\hidprop{\mpropn{vehicle-at}{?loc}}{2}$ & $=$ & $0.35 \cdot \hidprop{\mpropn{vehicle-at}{?loc}}{1} + 0.44 \cdot \pool\left(\hidact{\mactn{move-car}{?loc,\cdot}}{2}\right)+ 6.45$\\
        \midrule
        \multicolumn{3}{X}{\textit{Third action layer}: the most preferred actions are
          those that move to a location that has a tire and is adjacent to
          another location with a tire.
          \maschema{changetire} is only chosen when no other actions are available.\vspace{4pt}}\\
        \qquad $\hidact{\mactn{move-car}{?from,?to}}{3}$ & $=$ & $1.12 \cdot \hidprop{\mpropn{vehicle-at}{?to}}{2}$\\
        \qquad $\hidact{\mactn{changetire}{?loc}}{3}$ & $=$ & $0$\\
        \bottomrule
      \end{tabularx}
    \end{footnotesize}
    \caption{
        An easily-readable depiction of the sparse ASNet trained for the Triangle Tireworld problem, along with comments describing the intuitive meaning of action and proposition layers (demarcated by boxes).
        The ELU activation functions have been omitted because all activations are positive and the ELU is simply the identify function on $[0, \infty)$.
        %
        %
    }\label{fig:ttw-sparse-policy}
\end{figure}

\begin{figure}[ht]
    \centering
    \begin{footnotesize}
    \begin{overpic}[width=0.45\linewidth]{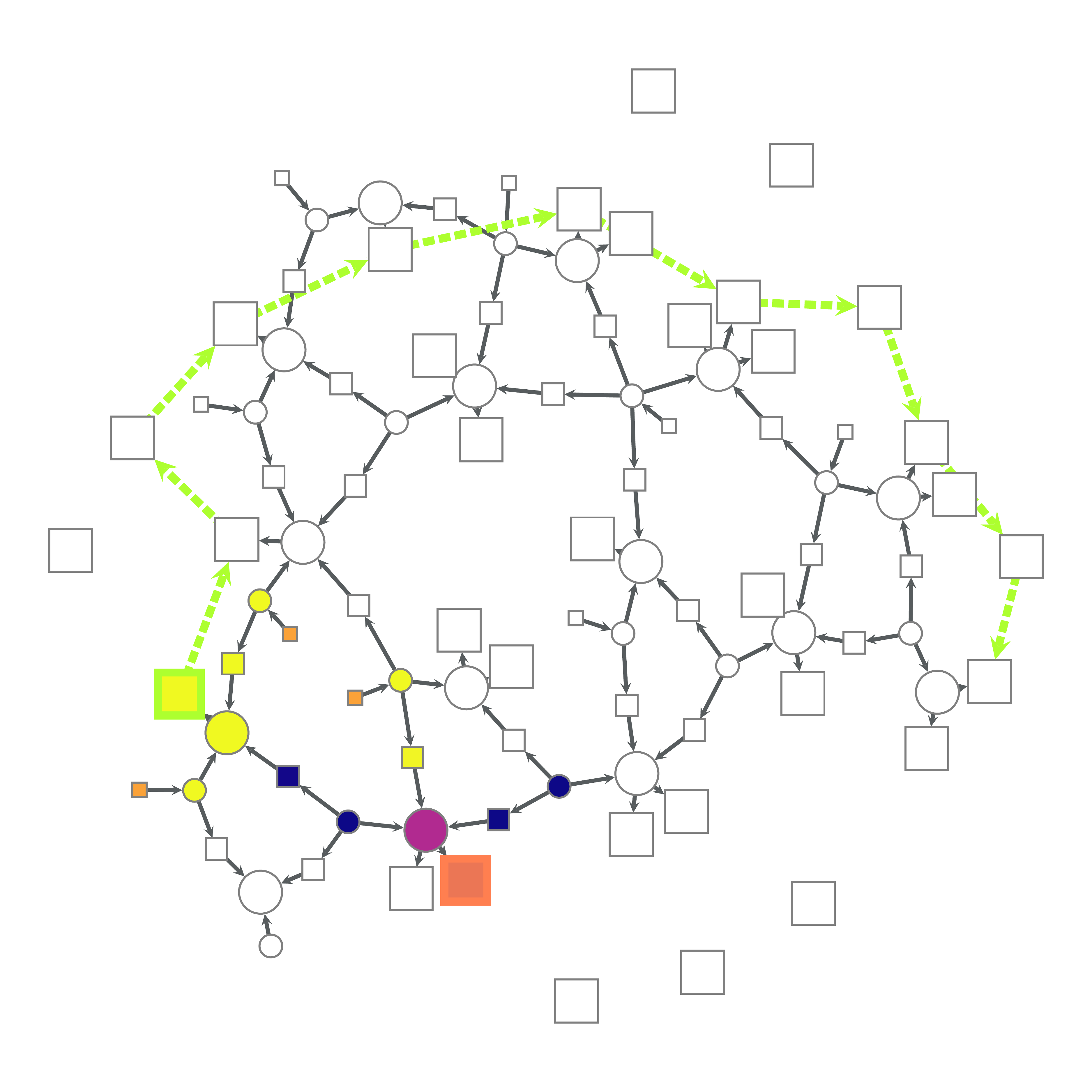}
      \put(0,96){$\mactn{move-car}{l_{1,1},l_{2,1}}$ (step 1/11)}
    \end{overpic}
    \begin{overpic}[width=0.45\linewidth]{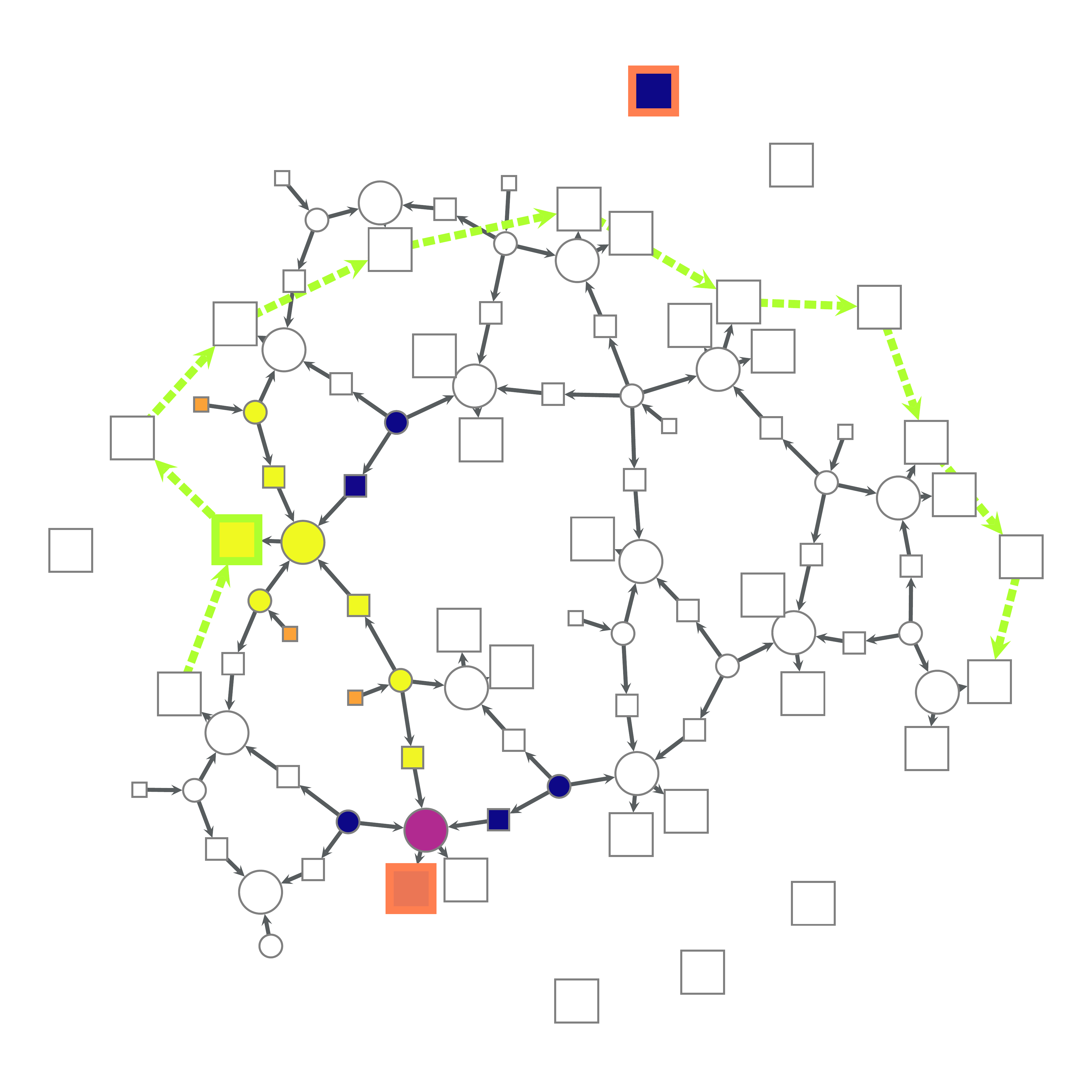}
      \put(0,96){$\mactn{move-car}{l_{2,1},l_{3,1}}$ (step 2/11)}
    \end{overpic}
    \begin{overpic}[width=0.45\linewidth]{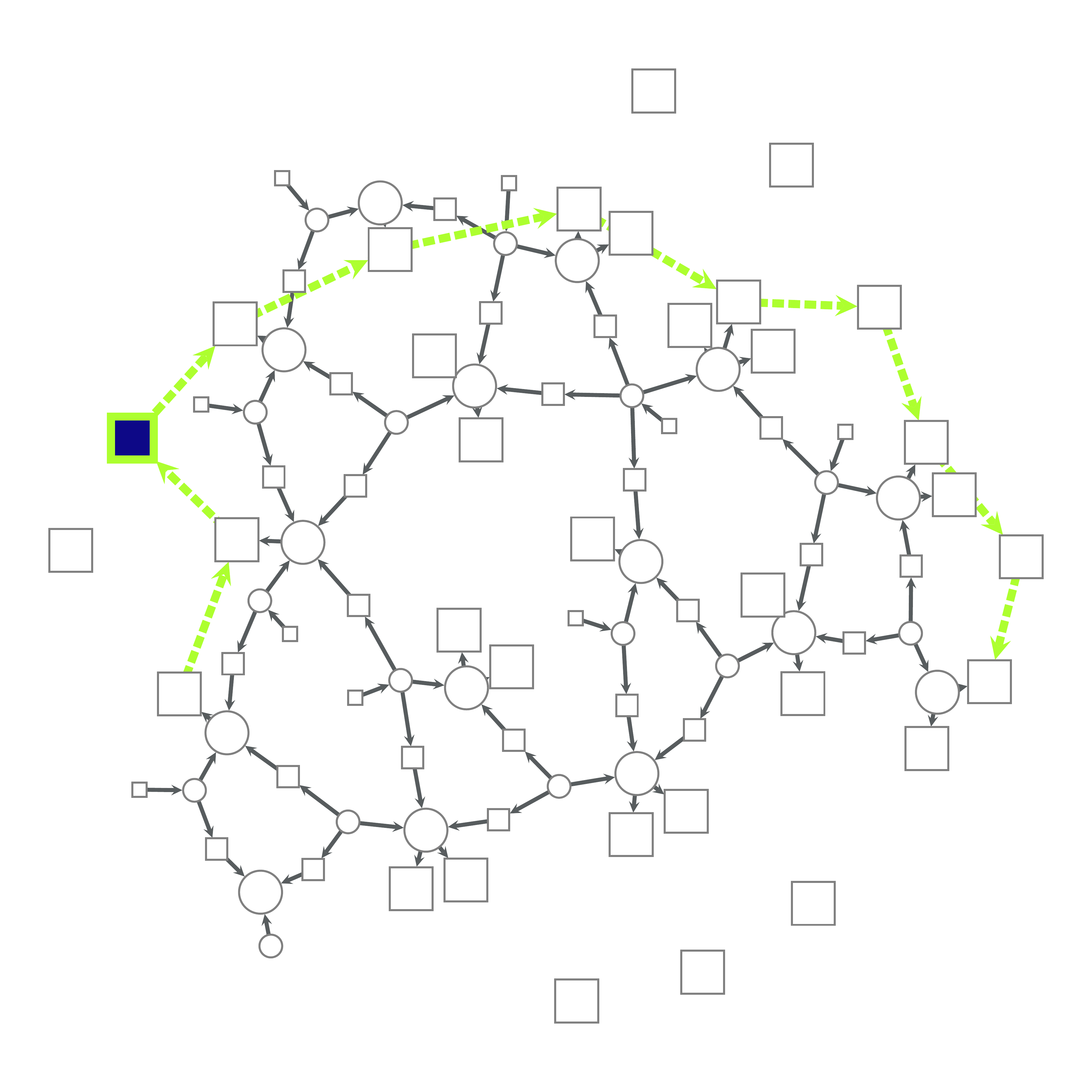}
      \put(0,96){$\mactn{change-tire}{l_{3,1}}$ (step 3/11)}
    \end{overpic}
    \begin{overpic}[width=0.45\linewidth]{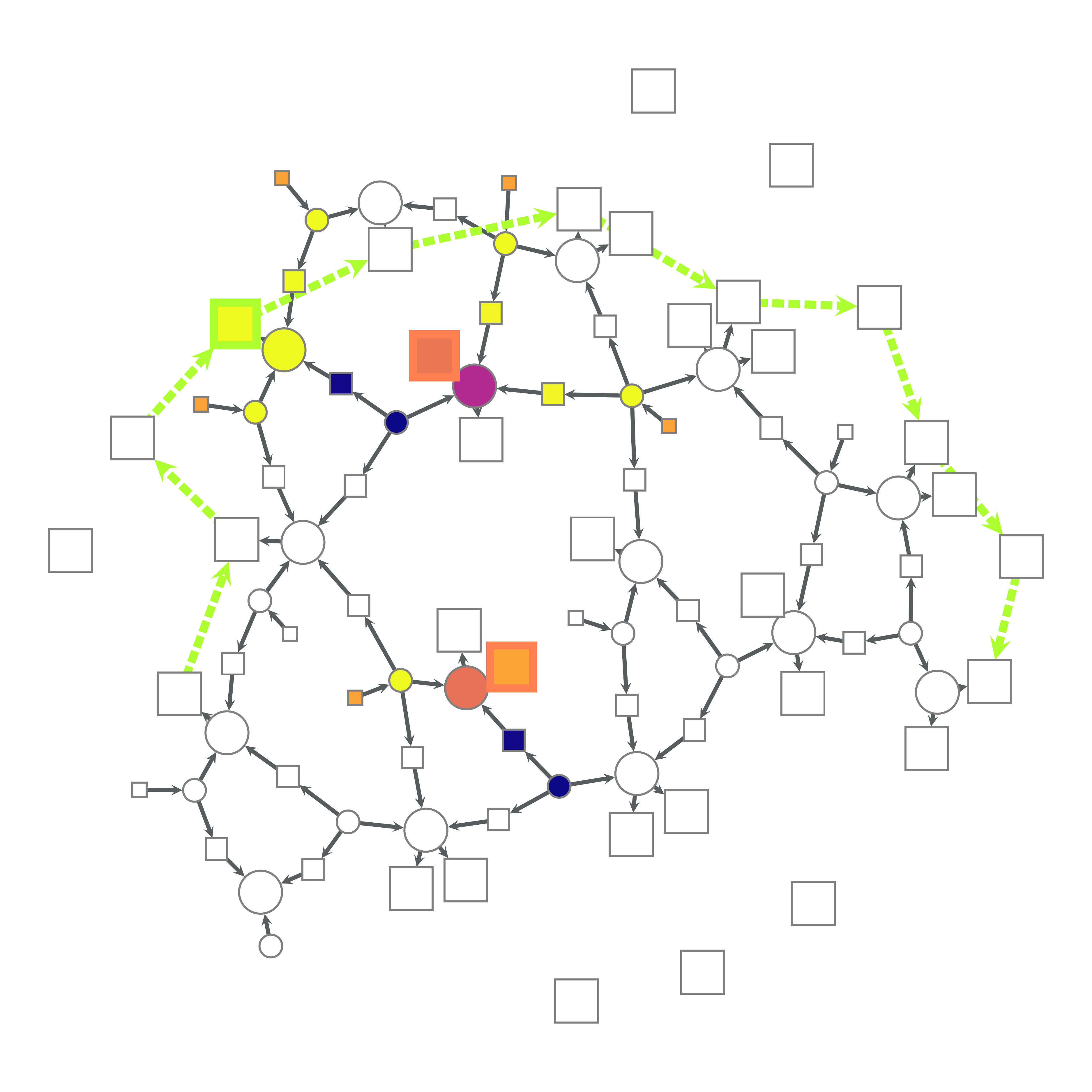}
      \put(0,96){$\mactn{move-car}{l_{3,1},l_{4,1}}$ (step 4/11)}
    \end{overpic}
    \includegraphics[width=0.8\linewidth]{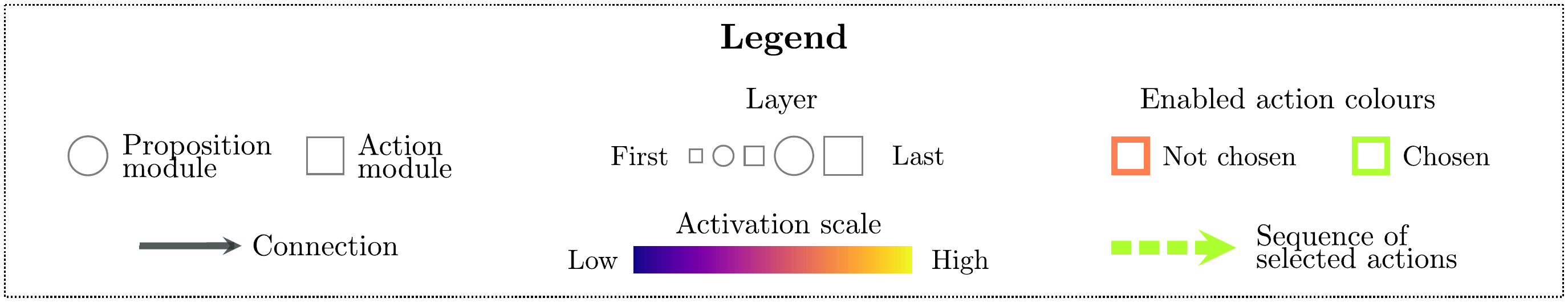}
    \end{footnotesize}
    \caption{
    Visualisation of the activations of an ASNet for the first four actions in a successful rollout on Triangle Tireworld.
    The diagram for ``step $t$'' represents the action chosen by the ASNet and executed at time $t$.
    }\label{fig:ttw-sparse-activations}
\end{figure}

\cref{fig:ttw-sparse-policy} depicts the ``lifted'' equations defining a sparse ASNet policy for Triangle Tireworld, trained using the procedure described in the preceding section.
This policy has a mere eight nonzero parameters.
As a result, all of the modules have either been reduced to a single nonzero output or eliminated (zeroed out) entirely, despite starting with $d_h = 16$ nonzero output channels at initialisation.
It is thus easy to verify that this policy is correct by cross-referencing with the PPDDL domain definition (\cref{fig:ttw-ppddl} in \cref{app:domain-descriptions}).
In the first action layer, we have an action module output $\hidact{\mactn{changetire}{?loc}}{1}$ for each location with a spare tire, which simply ignores the input values and outputs a large positive constant.
The procedure for collecting ground actions only includes $\maschema{changetire}$ actions for locations with a spare tire in the initial state,\footnote{
  Recall that we use the MDPSim grounding code, which instantiates all actions that could possibly be enabled in some state while attempting not to instantiate those actions that will never be applicable.
  In the case of Triangle Tireworld, $\mactn{changetire}{?loc}$ depends on a $\mpropn{spare-in}{?loc}$ proposition which cannot be made true by any action, so if $\mpropn{spare-in}{?loc}$ is not true in the initial state then $\mactn{changetire}{?loc}$ will never be added to the collected list of actions.
} so the $\hidprop{\mpropn{vehicle-at}{?loc}}{1}$ modules at the next layer will pool over zero or one corresponding $\hidact{\mactn{changetire}{?loc}}{1}$ modules to produce an output.
This output will be large if there is a tire at $\mobj{?loc}$ (and thus a corresponding $\maschema{changetire}$ action), and small otherwise.
The $\hidact{\mactn{move-car}{?from,?to}}{2}$ action modules in the next layer simply propagate the $\hidprop{\mpropn{vehicle-at}{?to}}{1}$ value for the destination location up to the next layer.
In the proposition layer that follows, $\hidprop{\mpropn{vehicle-at}{?loc}}{2}$ sums the output of a skip connection back to $\hidprop{\mpropn{vehicle-at}{?loc}}{1}$ (in the previous proposition layer) and a max-pooling operation over modules for $\maschema{move-car}$ locations that start in $\mobj{?loc}$ and end in some other location.
As a result, its value will be moderately high if there is a tire at $\mobj{?loc}$ or if some adjacent location that can be reached with one $\maschema{move-car}$ action, and very high if both conditions are true. 
In the final layer, $\hidact{\mactn{move-car}{?from,?to}}{3}$ attains the highest positive value when the destination $\mobj{?to}$ has a spare tire and also leads to another location with a spare tire.
It's easy to see from the diagrams in \cref{fig:ttw-locs} that following such actions will keep the vehicle on the outside edge of the triangle as it moves towards the goal, as desired.
Further, the final-layer module for $\maschema{changetire}$ always outputs 0, so it will only be chosen when $\maschema{changetire}$ is the only action available (i.e.\@ when the car has a flat tire, as one can see from the domain in \cref{fig:ttw-ppddl}).
We can thus conclude that this policy correctly exploits the structure of Triangle Tireworld instances in order to generalise across the entire domain.

We can gain another perspective on these sparse ASNet weights by using them to instantiate an ASNet for a small Triangle Tireworld problem, and then plotting the activations.
\cref{fig:ttw-sparse-activations} depicts such a visualisation for a Triangle Tireworld problem of size two.
In this visualisation, module outputs are represented by squares (action modules) or circles (proposition modules), where size increases with layer number.
We show activation magnitudes for the first four steps of a successful plan
execution.
Activations for modules that contributed to a decision are filled with an appropriate colour, while modules that could not have influenced the decision (due to network connectivity) are left white.
Additionally, we use a series of thick green arrows between final-layer activation modules to depict the sequence of actions chosen over the entire course of the plan.
Notice that the node positions computed by the force-directed layout algorithm that generated this figure roughly match the triangular road network for problem~(2) in \cref{fig:ttw-locs}.
This a good illustration of how the structure of an ASNet reflects the intuitive structure of the corresponding problem.

The activation diagram in \cref{fig:ttw-sparse-activations} also allows us to check some of the claims we made about the lifted policy.
For example, in the top left plot, we can see that the chosen action (large square, yellow with green outline) has a high activation because of the high activation $\hidprop{\mpropn{vehicle-at}{?loc}}{2}$ of the related $\maschema{vehicle-at}$ module in the preceding proposition layer (large yellow circle).
Further, that proposition module has a high activation because of its skip connection to a $\hidprop{\mpropn{vehicle-at}{?loc}}{1}$ module corresponding to a location with a tire (smaller yellow circle), and because it pools over a $\hidact{\mactn{move-car}{?loc,?next}}{2}$ module that leads to a location with a tire (smaller yellow square).
%
%
Likewise, in the bottom left plot of \cref{fig:ttw-sparse-activations}, we can see that the $\maschema{changetire}$ action is chosen despite its low activation (dark blue) because it is the only available action in states where the agent has a flat tire.
Together, \cref{fig:ttw-sparse-policy} and \cref{fig:ttw-sparse-activations} thus give us confidence that our learnt policy is correct.

\section{Related Work}
\label{sec:related}

To the best of our knowledge, ASNets were the first approach to generalised
probabilistic and deterministic planning with neural networks.
Nevertheless, there is a great deal of related prior work at the intersection of
learning and planning, as well as concurrent and later work that deals with similar themes.
In this section, we compare ASNets to prior work in planning
(\cref{ssec:cont-know} and \cref{ssec:other-related-planning}) and structured
deep learning (\cref{ssec:related-struct-dl}).

\subsection{Learning Generalised Domain-Specific Control Knowledge}\label{ssec:cont-know}

We will begin our survey of prior work by contrasting our work with previous
approaches to learning domain-specific control knowledge.
Although there are many forms of learnable domain-specific
knowledge~\cite{jimenez2012review}, our focus will be on generalised policies,
generalised heuristics, and other forms of knowledge that could conceivably be
learnt by ASNets.
We will decompose prior methods along three axes: first, we consider possible
representations for domain-specific knowledge.
Second, we consider methods used to learn domain-specific knowledge.
Third, and finally, we consider the methods through which the resulting
knowledge is exploited to solve unseen planning problems.

\subsubsection{Knowledge Representations}

Decision lists are one of the oldest and simplest representations for learnt
generalised policies in automatic planning.
They can be viewed as sequences of if--then rules which allow different actions
to be selected when different combinations of logical conditions are satisfied.
For instance, \citeA{khardon1999learning} represents a generalised policy using
a sequence of rules for selecting action schemas based on conjunctions over the
predicates of a domain.
Unfortunately, conditions based on conjunctions over fixed sets of predicates
are not sufficient to solve many planning problems of practical interest.
Thus, Khardon also employs hand-coded, domain-specific \textit{support
predicates} which allow the decision list to encode more powerful action
selection rules.
Later extensions to Khardon's algorithm allow it to make use of \textit{concept
languages}~\cite{martin2000learning} or \textit{taxonomic
syntax}~\cite{yoon2002inductive}.
Both concept languages and taxonomic syntax obviate the need for support
predicates by allowing more complex logical conditions (e.g.\@ those involving
recursion) to be employed in constructing a decision list.
\citeA{de2011scaling} present another extension of Khardon's
approach which also increases its expressive power.
First, they replace the single decision list with a pair of decision
trees:\footnote{
  Decision lists and decision trees have equivalent power in this
  context~\cite{blockeel1998top}, although the two representations do lend
  themselves to different training strategies.
} one for selecting an action schema, and one for binding objects to the action
schema to obtain a ground action.
Second, they introduce new features based on the helpful actions produced by the
FF planner.
Such heuristic features serve a similar purpose to support predicates, but do
not have to be manually coded for each problem; in a sense, use of these
heuristic features can therefore be viewed as an alternative to the use of
concept languages or taxonomic syntax.
\citeA{gretton2004exploiting} show that the rich concepts necessary to represent lifted policies can also be obtained by repeatedly applying logical regression to reward formulae in a first-order description of a domain, then performing inductive logic programming with those concepts.
It's worth noting that all of these methods can be combined in ensembles to produce more-accurate composite models~\cite{dietterich2000ensemble}.
Indeed, past work shows that ensembles can substantially improve the accuracy of
the aforementioned generalised planning
systems~\cite{yoon2002inductive,de2017bagging}.

Our approach fundamentally differs from all the aforementioned techniques in
that it uses a class of continuously-parameterised function approximators based
on graph convolutional neural networks, rather than using
discretely-parameterised decision lists (or trees).
This distinction is particularly relevant for learning, as explained in
\cref{ssec:know-acq}.
Further, unlike \citeA{yoon2002inductive} and \citeA{de2017bagging}, we find
that our models are sufficiently accurate to solve complex problems (e.g.\@
Blocksworld) without resorting to ensembles.
However, our use of heuristic input features is similar to the way that \citeA{de2011scaling} use helpful actions from FF, in that both methods use inputs derived in part from domain-independent heuristics to increase the range of policies that can be expressed without resorting to hand-coded input features.

Concurrent with the original ASNets paper, \citeA{groshev2018learning} also
proposed a novel representation for generalised policies and heuristics based on
neural networks.
They propose using a hand-coded, domain-specific translator to convert states of
a problem into a form that is amenable to processing by neural networks.
For instance, states from Sokoban can be processed by first converting them to
2D images depicting the current and goal positions of all boxes, the layout of
the warehouse walls, and the position of the agent.
The image can then be passed to a 2D CNN which produces an appropriate action or
heuristic value.
A travelling salesman problem can likewise be solved by expressing it as a graph
of locations to be visited and then processing the graph with a graph
convolutional neural network.
This approach is similar to ours insofar as it uses neural networks with
convolution-like operations.
However, it differs from our method in that it requires a manually-engineered input representation for each domain, whereas our method can easily be applied to any planning problem expressed as (P)PDDL.

\citeA{sievers2019deep} have also used traditional (image-based) convolutional neural networks for planning-related tasks.
Instead of learning a generalised policy like the present work or like \citeA{groshev2018learning}, they learn to perform planner
selection in classical planning domains.
Further, their process for converting planning instances into images is
domain-independent.
First they convert an instance into either a problem description graph, which captures the structure of variables and effects in a grounded planning task, or an abstract structure graph, which instead captures the structure of an instance at the level of un-grounded PDDL.
Next, they render the adjacency matrix of that graph as a binary image.
Finally, they dilate the image and resize it to fixed dimensions so that it can be fed to a convolutional neural network.
This work differs from ours in that it performs planner selection instead of action selection, operates on a different graph representation, and uses an image-based convolutional neural network to process the graph, rather than a graph convolutional network.

Closer to our work are the more recent ToRPIDo~\cite{bajpai2018transfer} and TraPSNet~\cite{bajpai2019size} neural network architectures, both of which allow transfer of learnt knowledge between different RDDL problems.
ToRPIDo uses a collection of neural network components, including a graph convolutional network, to represent a policy for a given planning problem.
After training on one problem, some components can be directly transferred to other instances, so long as those instances are of the same size as the original training problem(s).
This enables faster learning of policies for new problems, since only some components need to be re-learned from scratch.
TraPSNet is an improved architecture in which all network components can be transferred between problems of different size, under certain assumptions about domain structure (e.g.\@ that all action templates and fluents are unary).
TraPSNet uses a Graph ATtention network (GAT) and global pooling mechanism to learn an embedding vector for each object in a given RDDL problem.
By assuming that all action templates are unary, TraPSNet can also learn a sub-network that takes a single object embedding and indicates how desirable each of the corresponding actions are.
The weights of the GAT can be transferred between problems with different numbers of objects, and the action selection sub-network only needs to be applied to a single object at a time.
Hence, the weights of a TraPSNet can be transferred between problems with different numbers of objects.
This is analogous to ASNets' ability to transfer policies across problems of arbitrary size, but applied to a subset of RDDL rather than (P)PDDL.

Although ToRPIDo and TraPSNet serve a similar need to ASNets, we do not evaluate against either network in \cref{sec:expts}.
This is because the three network architectures are closely coupled with either RDDL (TorPIDo and TraPSNet) or PPDDL (ASNets).
Automated translations of problems from one language to the other generally produce a separate target-language domain for each input-language problem, which precludes direct comparison of generalisation ability.
Nevertheless, a comparison between the key components of the different network architectures (graph convolutions for ToRPIDo and ASNets, graph attention for TraPSNet) on problems expressed in the same language would make for interesting future work.

Another recent approach which uses graph networks to generalise learnt knowledge across tasks is the STRIPS Hypergraph Network (STRIPS-HGN)~\cite{shen2020learning}.
The hypergraph underlying the STRIPS-HGN architecture is derived from a delete relaxation of an input planning problem.
The vertices of the hypergraph each correspond to a particular proposition, while the hyperedges each correspond to a particular action.
Specifically, for each action $a$, there is a hyperedge linking the set of vertices corresponding to $\pre_a$ to the set of vertices corresponding to $\eff_a$.
The input to the network is an initial set of vertex and hyperedge features derived from the current and goal states; the output is a single heuristic estimate for the state.
Unlike ASNets, STRIPS-HGN architectures do not depend on the specifics of any particular domain, and so a single set of learnt weights can be applied to \textit{any} state of \textit{any} STRIPS problem.
Consequently, \citeauthor{shen2020learning} show that it can be used to learn either domain-specific or domain-independent heuristics.
Further, STRIPS-HGN uses a form of weight tying across different layers that allows the same ``layer'' (i.e. a single set of weights) to be applied multiple times to the same input, in much the same fashion as a recurrent layer in an RNN.
This can in principle overcome the receptive field limitation of ASNets by allowing the same layer to be applied as many times as is necessary to propagate information across the graph.
\citeauthor{shen2020learning} do not apply the architecture to probabilistic problems.
Further, they only investigate how to use the architecture to learn generalised heuristics, rather than generalised policies.

\subsubsection{Knowledge Acquisition}\label{ssec:know-acq}

In addition to an appropriate representation for domain-specific knowledge like
generalised policies, we also need an appropriate learning algorithm to acquire
that knowledge.
Most existing techniques for learning generalised policies incorporate a
strategy for obtaining experience---typically in the form of pairs of states and
``correct'' actions for some small problems---and a strategy for learning a
policy from that experience.
We will now consider these two aspects of knowledge acquisition (obtaining
experience and learning from it) in greater detail.

\paragraph{Acquiring experience}
Much like our work, most existing approaches to learning generalised policies
employ a non-learning ``teacher'' planner which tells the learning-based planner
which actions to take in the states observed at training time.
However, prior work differs in how the set of training states is generated.
One simple approach is to collect training states by running the teacher planner
on randomly-generated problem instances and then labelling and storing all
generated states along goal trajectories produced by the
teacher~\cite{martin2000learning,yoon2002inductive}.
However, the resulting training set would be static, and could not adapt to the
observed weaknesses of the planner during training; indeed,
\citeA{martin2000learning} show that it is helpful to extend the dataset with
states visited during training which were misclassified by the learnt policy.
The training set can be expanded further by including \textit{all} optimal goal trajectories for the given training problems, rather than just a subset.
This can be achieved, for instance, with a branch-and-bound algorithm~\cite{de2011scaling}.
We found it most effective to use a training set composed of all states visited by the agent during rollouts, plus rollouts under the teacher planner's policy starting in each of those states.

Rather than relying solely on high-quality plans from a teacher, some prior
methods use self-supervision or reinforcement learning to learn on problems that
may be too large for other planners to handle.
For instance, the Factored Policy Gradient (FPG) planner learns a policy for a
single problem via policy gradient reinforcement learning, which gradually
tweaks the parameters of a learnt policy so that its probability of success
increases over time~\cite{buffet2009factored}.
We attempted to use the same technique, but found that the random exploration
employed by policy gradient methods was too inefficient to solve our benchmark
problems.
\citeA{groshev2018learning} present an alternative \textit{leapfrogging}
approach that combines self-supervision with supervision from a teacher planner.
It begins by using a teacher policy to acquire experience on small problems.
Later, it uses its partially-learnt control knowledge to guide a search
algorithm on larger problems, and then feeds the actions recommended by that
self-guided search algorithm back into its own training set.
The same method is likely applicable to our model, although we leave it to
future work to investigate this and other techniques for interleaving learning
and planning.

\paragraph{Learning a policy from experience}
Different knowledge representations lend themselves to different learning
algorithms for distilling experience into control knowledge.
For decision list representations, a standard approach is Rivest's
algorithm~\cite{rivest1987learning}, which iteratively builds a decision list by
adding rules with perfect precision or perfect recall until all samples are
classified correctly.
This process requires a search over the space of possible conditions at each
iteration, so its efficiency is dependent on either restricting the size of this
space or having a good search algorithm at hand to find useful conditions.
For instance, \citeA{yoon2002inductive} restrict the size of their taxonomic
syntax expressions, and employ heuristic-guided beam search to find useful
expressions at each iteration.
The different knowledge representation of \citeA{de2011scaling} allows them to
instead use standard methods for learning decision
trees~\cite{blockeel1998top,quinlan1986induction}.
Nevertheless, all these methods must perform a search through a space of
discrete models, and consequently suffer from all of the issues that discrete
search entails (high branching factor, difficulty of guiding the search, etc.).

In contrast to the above approaches, we use a continuously-parameterised
knowledge representation which can be trained via Stochastic Gradient Descent
(SGD).
Training via SGD offers a different set of tradeoffs to search in discrete
spaces, and arguably provides greater flexibility by allowing us to
optimise \textit{any} differentiable loss.
For instance, although this paper only examines the performance of our
model in a classification setting (with a cross-entropy loss), it could just as
easily be used to regress Q-values (e.g. with an $\ell_2$ loss).
Likewise, we could train ASNets with policy gradient reinforcement learning; this could allow ASNets to be used to select planning strategies in the recent framework of \citeA{gomoluch2018learning}, for example.
In contrast, past approaches using discrete representations would need
substantially different optimisation algorithms in order to extend them to
value-learning or reinforcement-learning settings.

\subsubsection{Knowledge Exploitation}\label{ssec:rel-knowl-exploit}

When learnt control knowledge is expressed in the form of a generalised policy,
the most obvious way to exploit it is to simply execute it directly.
However, direct execution is not always possible for other forms of learnt
control knowledge, such as generalised heuristics, and is not always the best
way to make use of a generalised policy.
For instance, \citeA{de2011scaling} note that learnt generalised policies can
sometimes include defective rules, and propose depth-first and breadth-first
search algorithms to ameliorate this problem: the search is guided by the learnt
policy, but is also able to back-track if it reaches a dead end.
Beam search~\cite{xu2007discriminative}, limited discrepancy
search~\cite{yoon2006discrepancy}, and ordinary $\text{A}^\star$ search have all
been used similarly.
For probabilistic problems, it is potentially more appropriate to use
sampling-based strategies to either expand the search space of a heuristic
search algorithm~\cite{yoon2007using} or to estimate the Q-values of actions
using policy rollouts~\cite{fern2004approximate}.
Along the latter lines, it is also possible to apply Monte Carlo Tree Search (MCTS) in conjunction with learnt policies, as is done by AlphaGo~\cite{silver2016mastering}.
As mentioned in \cref{sec:train-exploit}, concurrent work has shown that using ASNets in conjunction with UCT at test time can help prevent mistakes that might occur due to incomplete training of the ASNet~\cite{shen2019guiding}.
In this paper, we are primarily concerned with examining what sort of
generalised policies ASNets can represent directly, so we have not experimented
further with search-based methods for exploiting generalised policies.

\subsection{Other Related Planning Work}\label{ssec:other-related-planning}

Not all work at the intersection of learning and planning fits into
\cref{ssec:cont-know}'s taxonomy for generalised knowledge acquisition.
For instance, the previously-mentioned FPG planner uses reinforcement learning
to train a neural-net-based policy, but the resulting policy is only able to
solve a \textit{single} problem~\cite{buffet2009factored}.
Likewise, \citeA{issakkimuthu2018training} investigate a restricted class of
neural networks for representing single-instance policies, including networks
that use a limited form of weight-sharing.
\citeA{ferber2020neural} similarly investigate the properties that make
fully-connected neural networks well-suited to learning problem-specific
heuristics that generalise only over the initial state of the problem.
The primary difference between these papers and this work is our focus
on generalising across different problems.

There are also several existing approaches to generalised planning that do not
approach the problem from a machine learning perspective.
\citeA{srivastava2011directed} consider how to acquire generalised plans for
domains expressed with a restricted form of classical planning.
Rather than learning from traces produced by a teacher planner on a small, fixed
set of training instances, Srivastava~\etal{} instead assume access to an
algorithm that can automatically generate states which are not assigned an
action by the current (partially-complete) generalised plan.
In this way, they can sometimes acquire generalised plans that are
\textit{guaranteed} to provide an action for all encountered states, and to
terminate eventually.
Continuing on the same theme, \citeA{hu2011generalized} consider the complexity
of generalised planning for finite and infinite environments, while
\citeA{bonet2018features} propose a practical generalised planning algorithm
that can accommodate changes in the number of objects and actions in the
problems of a domain.
\citeA{frances2019generalized} propose an algorithm that can recover generalised
heuristics from linear combinations numeric features derived from simple concept
language expressions.
In some domains, they are able to manually prove that the recovered heuristics
are \textit{descending and dead-end avoiding} across the entire domain, and
can consequently guide a greedy planner to a goal state in polynomial time.
In contrast to these approaches, our approach does not provide any theoretical
guarantees about generalisation of a learnt model to unseen instances.
However, our method also imposes fewer limitations on the sorts of problems that
can (in principle) be solved.
For instance, \citeA{srivastava2011directed} only consider ``generalisation to
$n$'', where problems are identical but for the number of instances of a
certain kind of object.
It remains to be seen whether there is a compromise approach which can offer
reasonable guarantees about generalisation while still remaining applicable to a
wide range of problems.

There have also been a number of related techniques that use deep learning to
acquire models of an environment, and then obtain policies through reinforcement
learning or planning.
For instance, Value Iteration Networks (VINs) are a kind of convolutional neural
network that can learn to formulate an MDP from an observation of an environment,
solve that MDP, and use the result to choose an action~\cite{tamar2016value}.
Generalised VINs extend this approach to MDPs with more general transition
dynamics by employing graph convolutional neural networks instead of ordinary
convolutional neural networks~\cite{niu2017generalized}.
In a similar vein, \textit{schema networks} learn a STRIPS-like environment
model using a specially-structured neural network, then choose actions by
planning on that learnt model~\cite{kansky2017schema}.
\citeA{say2017nonlinear} show that it's possible to learn transition
models for mixed discrete--continuous planning problems using deep learning, and
then plan on the learnt model with a traditional MILP solver, as opposed to
reinforcement learning.
LatPlan~\cite{asai2018classical} likewise shows that discrete autoencoders
(specifically, Gumbel-Softmax VAEs) can learn how to convert image-based
observations of an environment into a PDDL problem description, which can then
be solved using an ordinary classical planner.
Finally, \citeA{zambaldi2018deep} show that adding a form of self-attention~\cite{vaswani2017attention} to convolutional network policies can make it easier for reinforcement learning to solve image-based tasks requiring relational reasoning.
These approaches bear surface similarities to our approach, such as the use of
convolutions, or the use of an internal representation based on the
action-proposition structure of a problem.
However, they differ in that they all aim to learn models for unknown
environments which can then be planned on, as opposed to directly learning a
policy for a known environment.

\subsection{Structured Deep Learning}\label{ssec:related-struct-dl}

ASNets can be interpreted as an extension of convolutional neural networks to
handle a different kind of underlying graph structure.
These sorts of generalisations have been widely studied in recent years,
motivated in part by the desire to apply deep learning to domains where data
cannot be expressed as points sampled on a regular $n$-dimensional spatial
grid~\cite{bronstein2017geometric}.
For instance, graph convolutional neural networks have previously been used to
model molecules~\cite{duvenaud2015convolutional,kearnes2016molecular}, where a
graph structure is induced by the bonds between atoms, and spatial-temporal
action sequences~\cite{jain2016structural}, where interactions between objects
in an environment are modelled as a graph.
These techniques have also been used to reason about relational
databases~\cite{uwents2005classifying}, where the structure of the network is
determined by the database schema, and to first-order logic~\cite{sourek2018lifted}, where
the structure of the network is derived from the structure of a series of first
order logic statements.
So far as we are aware, ASNets are the first application of this technique to
automated planning.
%

\section{Conclusion and Future Work}
\label{sec:conc}

State-of-the-art classical and probabilistic planners currently have limited ability to transfer control knowledge between similar problems or domains.
The most common form of transfer is to learn an autoselector or autoconfigurator that chooses an appropriate planner for new problems based on past planner performance, but such methods only have a limited ability to influence the search for plans and policies.
This paper discussed Action Schema Networks (ASNets), a neural network representation for generalised policies.
An ASNet applies a learnt convolution-like operator to a graph of actions and propositions in order to select appropriate actions for a problem.
Because the number of parameters required by this convolution-like operator is independent of any one problem, the resulting policy can be applied to any instance from a given (P)PDDL domain.
Our experiments across seven probabilistic and deterministic domains show that training ASNets on some small instances from a domain and then executing the resulting policy on a few large instances can be much faster than applying a heuristic search planner directly to those large instances.
Each domain has a simple ``trick'' that the ASNet can learn from small problems which makes planning in larger problems straightforward.
In contrast, the baseline planners cannot transfer any of their experience between problems, and must instead re-discover such tricks anew on each instance.

This paper has also challenged the view that neural-network-based policies must necessarily be unreliable and uninterpretable.
In response to \citeA{geffner2018ijcai}, we showed that ASNets could learn a highly reliable policy on the classical Blocksworld benchmark.
Our policy correctly solved 18,300 test instances with 18--50 blocks after training on just 50 smaller instances with 8--10 blocks.
We have also shown that it is possible to train very sparse ASNet policies for Triangle Tireworld and CosaNostra Pizza which can be interpreted by a human.
There will still be problems for which ASNets cannot learn a reliable policy, or where the policy cannot be made sparse enough for easy interpretation, but our results show that deep learning still can yield interpretable, reliable policies in some settings.

While ASNets have demonstrated impressive performance on some domains, there still remain many directions for future improvement.
One direction is the exploration of search methods that can better exploit the learnt control knowledge in an ASNet, including helping to solve problems that ASNets cannot solve on their own, or for which learnt control knowledge is flawed.
\citeA{shen2019guiding} have already taken some steps towards this goal by using ASNets to guide Monte Carlo tree search.
Another possible direction is to experiment with different methods of training ASNets.
Our current training mechanism simply has ASNets imitate a ``teacher'' planner on a small set of problems, even though it may not be possible to generalise that teacher's strategy to larger problems.
Reinforcement learning is more appealing in this regard, but further advances in RL will be required to make RL-based training reliable and efficient enough to be practical on the benchmarks used in classical and probabilistic planning.
It would also be interesting to investigate alternative architectures that lift the structural limitations of ASNets.
The fixed receptive field of an ASNet is one such limitation, as is its lack of support for quantified action preconditions and arbitrary goal formulae.
The computational overhead of ASNets in problems with many ground actions or propositions is another important limitation.
Indeed, there may be alternative generalised policy architectures that do not require a ground representation at all!
Finally, we note that ASNets could easily be extended to apply to more areas beyond probabilistic and classical planning, including planning problems with numeric state, and problems with concurrent, durative actions.
More generally, we hope that ASNets can serve as both a starting point and inspiration for researchers looking to bring the advances of deep learning to the planning community.

\acks{
  We would like to acknowledge J\"org Hoffmann for suggesting the alternative
  pooling mechanism in \cref{sec:asnets}, and Hector Geffner for
  insightful discussion about network structure and generalisation.
  We would also like to thank our anonymous reviewers for helping to
  strengthen the paper, particularly in the background and evaluation sections.

  Sylvie Thi\'ebaux and Felipe Trevizan are partially funded by Australian
  Research Council Discovery Project grant DP180103446 ``On-line Planning for
  Constrained Autonomous Agents in an Uncertain World.''
}

\appendix
\section{Coverage and Cost Results on Test Domains}\label{app:results-tables}

\begin{subtables}
\begin{table}[H]
  \begin{center}
    \begin{footnotesize}
      \begin{tabular}{@{}lccccc@{}}
      \toprule
      \multirow{2}{*}{Problem}\wskip&
        & \multicolumn{4}{c}{ASNet}
        \\
        \cmidrule{3-6}
        &\wskip & - & PE & Adm. & No h.\\
        \midrule
cosanostra-n6 &\wskip
  & \makecell{30/30 \\ (22.0 $\pm$ 0)} 
  & \makecell{30/30 \\ (22.0 $\pm$ 0)} 
  & \makecell{30/30 \\ (22.0 $\pm$ 0)} 
  & \makecell{30/30 \\ (22.0 $\pm$ 0)} 
  \\
cosanostra-n7 &\wskip
  & \makecell{30/30 \\ (25.0 $\pm$ 0)} 
  & \makecell{30/30 \\ (25.0 $\pm$ 0)} 
  & \makecell{30/30 \\ (25.0 $\pm$ 0)} 
  & \makecell{30/30 \\ (25.0 $\pm$ 0)} 
  \\
cosanostra-n8 &\wskip
  & \makecell{30/30 \\ (28.0 $\pm$ 0)} 
  & \makecell{30/30 \\ (28.0 $\pm$ 0)} 
  & \makecell{30/30 \\ (28.0 $\pm$ 0)} 
  & \makecell{30/30 \\ (28.0 $\pm$ 0)} 
  \\
cosanostra-n9 &\wskip
  & \makecell{30/30 \\ (31.0 $\pm$ 0)} 
  & \makecell{30/30 \\ (31.0 $\pm$ 0)} 
  & \makecell{30/30 \\ (31.0 $\pm$ 0)} 
  & \makecell{30/30 \\ (31.0 $\pm$ 0)} 
  \\
cosanostra-n10 &\wskip
  & \makecell{30/30 \\ (34.0 $\pm$ 0)} 
  & \makecell{30/30 \\ (34.0 $\pm$ 0)} 
  & \makecell{30/30 \\ (34.0 $\pm$ 0)} 
  & \makecell{30/30 \\ (34.0 $\pm$ 0)} 
  \\
cosanostra-n11 &\wskip
  & \makecell{30/30 \\ (37.0 $\pm$ 0)} 
  & \makecell{30/30 \\ (37.0 $\pm$ 0)} 
  & \makecell{30/30 \\ (37.0 $\pm$ 0)} 
  & \makecell{30/30 \\ (37.0 $\pm$ 0)} 
  \\
cosanostra-n12 &\wskip
  & \makecell{30/30 \\ (40.0 $\pm$ 0)} 
  & \makecell{30/30 \\ (40.0 $\pm$ 0)} 
  & \makecell{30/30 \\ (40.0 $\pm$ 0)} 
  & \makecell{30/30 \\ (40.0 $\pm$ 0)} 
  \\
cosanostra-n13 &\wskip
  & \makecell{30/30 \\ (43.0 $\pm$ 0)} 
  & \makecell{30/30 \\ (43.0 $\pm$ 0)} 
  & \makecell{30/30 \\ (43.0 $\pm$ 0)} 
  & - 
  \\
cosanostra-n14 &\wskip
  & \makecell{30/30 \\ (46.0 $\pm$ 0)} 
  & \makecell{30/30 \\ (46.0 $\pm$ 0)} 
  & \makecell{30/30 \\ (46.0 $\pm$ 0)} 
  & - 
  \\
cosanostra-n15 &\wskip
  & \makecell{30/30 \\ (49.0 $\pm$ 0)} 
  & \makecell{30/30 \\ (49.0 $\pm$ 0)} 
  & \makecell{30/30 \\ (49.0 $\pm$ 0)} 
  & - 
  \\
cosanostra-n20 &\wskip
  & \makecell{30/30 \\ (64.0 $\pm$ 0)} 
  & \makecell{30/30 \\ (64.0 $\pm$ 0)} 
  & \makecell{30/30 \\ (64.0 $\pm$ 0)} 
  & - 
  \\
cosanostra-n25 &\wskip
  & \makecell{30/30 \\ (79.0 $\pm$ 0)} 
  & \makecell{30/30 \\ (79.0 $\pm$ 0)} 
  & \makecell{30/30 \\ (79.0 $\pm$ 0)} 
  & - 
  \\
cosanostra-n30 &\wskip
  & \makecell{30/30 \\ (94.0 $\pm$ 0)} 
  & \makecell{29/30 \\ (94.0 $\pm$ 0)} 
  & \makecell{30/30 \\ (94.0 $\pm$ 0)} 
  & - 
  \\
cosanostra-n35 &\wskip
  & \makecell{30/30 \\ (109.0 $\pm$ 0)} 
  & \makecell{30/30 \\ (109.0 $\pm$ 0)} 
  & \makecell{30/30 \\ (109.0 $\pm$ 0)} 
  & - 
  \\
cosanostra-n40 &\wskip
  & \makecell{30/30 \\ (124.0 $\pm$ 0)} 
  & \makecell{30/30 \\ (124.0 $\pm$ 0)} 
  & \makecell{30/30 \\ (124.0 $\pm$ 0)} 
  & - 
  \\
cosanostra-n45 &\wskip
  & \makecell{30/30 \\ (139.0 $\pm$ 0)} 
  & \makecell{30/30 \\ (139.0 $\pm$ 0)} 
  & \makecell{30/30 \\ (139.0 $\pm$ 0)} 
  & - 
  \\
cosanostra-n50 &\wskip
  & \makecell{30/30 \\ (154.0 $\pm$ 0)} 
  & \makecell{30/30 \\ (154.0 $\pm$ 0)} 
  & \makecell{30/30 \\ (154.0 $\pm$ 0)} 
  & - 
  \\

        \bottomrule
      \end{tabular}
    \end{footnotesize}
  \end{center}
  \caption{
  Results for ASNets on CosaNostra Pizza; results for probabilistic baseline
  planners are given in the next sub-table.
  For each planner, we report the fraction of 30 rollouts that reached the
  goal, along with the mean trajectory cost (and 95\% confidence interval
  bounds) for successful rollouts.
  The number in each problem name denotes the number of toll booths between the
  pizza shop and the customer.
}
\label{tab:prob-res-cn-1}
\end{table}

\begin{table}[H]
  \begin{center}
    \begin{footnotesize}
      \begin{tabular}{@{}lcccccc@{\vphantom{\makecell{1 \\ 2}}}}
      \toprule
      \multirow{2}{*}{Problem}\wskip&
        & \multicolumn{2}{c}{LRTDP}\wskip&
        & \multicolumn{2}{c}{SSiPP}
        \\
        \cmidrule{3-4} \cmidrule{6-7}
        &\wskip & h-add & LM-cut &\wskip & h-add & LM-cut\\
        \midrule
cosanostra-n6 &\wskip
  & \makecell{30/30 \\ (22.0 $\pm$ 0)} 
  & \makecell{30/30 \\ (22.0 $\pm$ 0)} &\wskip 
  & \makecell{30/30 \\ (22.0 $\pm$ 0)} 
  & \makecell{30/30 \\ (22.0 $\pm$ 0)} 
  \\
cosanostra-n7 &\wskip
  & \makecell{30/30 \\ (25.0 $\pm$ 0)} 
  & \makecell{30/30 \\ (25.0 $\pm$ 0)} &\wskip 
  & \makecell{30/30 \\ (25.0 $\pm$ 0)} 
  & \makecell{30/30 \\ (25.0 $\pm$ 0)} 
  \\
cosanostra-n8 &\wskip
  & \makecell{30/30 \\ (28.0 $\pm$ 0)} 
  & \makecell{30/30 \\ (28.0 $\pm$ 0)} &\wskip 
  & \makecell{30/30 \\ (28.0 $\pm$ 0)} 
  & \makecell{30/30 \\ (28.0 $\pm$ 0)} 
  \\
cosanostra-n9 &\wskip
  & \makecell{30/30 \\ (31.0 $\pm$ 0)} 
  & \makecell{30/30 \\ (31.0 $\pm$ 0)} &\wskip 
  & \makecell{30/30 \\ (31.0 $\pm$ 0)} 
  & \makecell{30/30 \\ (31.0 $\pm$ 0)} 
  \\
cosanostra-n10 &\wskip
  & \makecell{30/30 \\ (34.0 $\pm$ 0)} 
  & \makecell{11/30 \\ (34.0 $\pm$ 0)} &\wskip 
  & \makecell{30/30 \\ (34.0 $\pm$ 0)} 
  & \makecell{11/30 \\ (34.0 $\pm$ 0)} 
  \\
cosanostra-n11 &\wskip
  & \makecell{30/30 \\ (37.0 $\pm$ 0)} 
  & \makecell{4/30 \\ (37.0 $\pm$ 0)} &\wskip 
  & \makecell{23/30 \\ (37.0 $\pm$ 0)} 
  & \makecell{4/30 \\ (37.0 $\pm$ 0)} 
  \\
cosanostra-n12 &\wskip
  & \makecell{13/30 \\ (40.0 $\pm$ 0)} 
  & \makecell{1/30 \\ (40.0)} &\wskip 
  & \makecell{10/30 \\ (40.0 $\pm$ 0)} 
  & \makecell{2/30 \\ (40.0 $\pm$ 0)} 
  \\
cosanostra-n13 &\wskip
  & \makecell{6/30 \\ (43.0 $\pm$ 0)} 
  & - &\wskip 
  & \makecell{4/30 \\ (43.0 $\pm$ 0)} 
  & - 
  \\
cosanostra-n14 &\wskip
  & \makecell{2/30 \\ (46.0 $\pm$ 0)} 
  & - &\wskip 
  & \makecell{2/30 \\ (46.0 $\pm$ 0)} 
  & - 
  \\
cosanostra-n15 &\wskip
  & \makecell{1/30 \\ (49.0)} 
  & - &\wskip 
  & - 
  & - 
  \\
cosanostra-n20 &\wskip
  & - 
  & - &\wskip 
  & - 
  & - 
  \\
cosanostra-n25 &\wskip
  & - 
  & - &\wskip 
  & - 
  & - 
  \\
cosanostra-n30 &\wskip
  & - 
  & - &\wskip 
  & - 
  & - 
  \\
cosanostra-n35 &\wskip
  & - 
  & - &\wskip 
  & - 
  & - 
  \\
cosanostra-n40 &\wskip
  & - 
  & - &\wskip 
  & - 
  & - 
  \\
cosanostra-n45 &\wskip
  & - 
  & - &\wskip 
  & - 
  & - 
  \\
cosanostra-n50 &\wskip
  & - 
  & - &\wskip 
  & - 
  & - 
  \\

        \bottomrule
      \end{tabular}
    \end{footnotesize}
  \end{center}
  \caption{Results for probabilistic baseline planners on CosaNostra Pizza.}
  \label{tab:prob-res-cn-2}
\end{table}
\end{subtables}

\begin{subtables}
\begin{table}[H]
  \begin{center}
    \smallscript
    \begin{tabular}{@{}lccccc@{}}
    \toprule
    \multirow{2}{*}{Problem}\wskip&
      & \multicolumn{4}{c}{ASNet}
      \\
      \cmidrule{3-6}
      &\wskip & - & PE & Adm. & No h.\\
      \midrule
ex-bw-b11-s0 &\wskip
  & \makecell{15/30 \\ (28.0 $\pm$ 0)} 
  & \makecell{9/30 \\ (36.7 $\pm$ 9.8)} 
  & \makecell{15/30 \\ (28.0 $\pm$ 0)} 
  & \makecell{17/30 \\ (28.9 $\pm$ 0.5)} 
  \\
ex-bw-b11-s1 &\wskip
  & - 
  & \makecell{3/30 \\ (88.0 $\pm$ 86.2)} 
  & - 
  & - 
  \\
ex-bw-b11-s2 &\wskip
  & \makecell{16/30 \\ (26.0 $\pm$ 0)} 
  & \makecell{17/30 \\ (30.2 $\pm$ 1.6)} 
  & \makecell{16/30 \\ (26.0 $\pm$ 0)} 
  & - 
  \\
ex-bw-b12-s0 &\wskip
  & \makecell{29/30 \\ (26.0 $\pm$ 0)} 
  & \makecell{30/30 \\ (26.3 $\pm$ 0.3)} 
  & \makecell{29/30 \\ (26.0 $\pm$ 0)} 
  & \makecell{29/30 \\ (26.0 $\pm$ 0)} 
  \\
ex-bw-b12-s1 &\wskip
  & - 
  & \makecell{4/30 \\ (65.0 $\pm$ 32.9)} 
  & \makecell{3/30 \\ (56.0 $\pm$ 0)} 
  & - 
  \\
ex-bw-b12-s2 &\wskip
  & - 
  & \makecell{5/30 \\ (76.4 $\pm$ 32.1)} 
  & \makecell{5/30 \\ (42.4 $\pm$ 13.8)} 
  & - 
  \\
ex-bw-b13-s0 &\wskip
  & \makecell{26/30 \\ (25.6 $\pm$ 1.1)} 
  & \makecell{24/30 \\ (27.4 $\pm$ 1.5)} 
  & \makecell{29/30 \\ (25.2 $\pm$ 0.4)} 
  & \makecell{27/30 \\ (25.3 $\pm$ 0.4)} 
  \\
ex-bw-b13-s1 &\wskip
  & - 
  & \makecell{2/30 \\ (56.0 $\pm$ 203.3)} 
  & \makecell{7/30 \\ (60.3 $\pm$ 23.1)} 
  & - 
  \\
ex-bw-b13-s2 &\wskip
  & \makecell{11/30 \\ (58.7 $\pm$ 10.2)} 
  & \makecell{2/30 \\ (57.0 $\pm$ 63.5)} 
  & \makecell{11/30 \\ (49.1 $\pm$ 12.9)} 
  & - 
  \\
ex-bw-b14-s0 &\wskip
  & - 
  & \makecell{5/30 \\ (64.8 $\pm$ 17.3)} 
  & - 
  & - 
  \\
ex-bw-b14-s1 &\wskip
  & - 
  & \makecell{6/30 \\ (62.0 $\pm$ 20.7)} 
  & \makecell{15/30 \\ (34.0 $\pm$ 0)} 
  & - 
  \\
ex-bw-b14-s2 &\wskip
  & \makecell{12/30 \\ (38.0 $\pm$ 0)} 
  & \makecell{12/30 \\ (40.7 $\pm$ 1.9)} 
  & \makecell{14/30 \\ (36.0 $\pm$ 0)} 
  & \makecell{12/30 \\ (38.0 $\pm$ 0)} 
  \\
ex-bw-b15-s0 &\wskip
  & \makecell{6/30 \\ (41.3 $\pm$ 1.1)} 
  & \makecell{9/30 \\ (45.8 $\pm$ 3.0)} 
  & \makecell{7/30 \\ (48.3 $\pm$ 4.9)} 
  & - 
  \\
ex-bw-b15-s1 &\wskip
  & \makecell{6/30 \\ (69.3 $\pm$ 18.7)} 
  & \makecell{2/30 \\ (61.0 $\pm$ 38.1)} 
  & \makecell{3/30 \\ (56.0 $\pm$ 9.9)} 
  & - 
  \\
ex-bw-b15-s2 &\wskip
  & \makecell{18/30 \\ (56.7 $\pm$ 5.8)} 
  & \makecell{14/30 \\ (54.1 $\pm$ 16.6)} 
  & \makecell{13/30 \\ (51.2 $\pm$ 3.5)} 
  & \makecell{14/30 \\ (36.0 $\pm$ 0)} 
  \\
ex-bw-b16-s0 &\wskip
  & - 
  & \makecell{13/30 \\ (69.8 $\pm$ 17.7)} 
  & \makecell{14/30 \\ (53.0 $\pm$ 12.2)} 
  & - 
  \\
ex-bw-b16-s1 &\wskip
  & \makecell{4/30 \\ (61.0 $\pm$ 20.2)} 
  & \makecell{4/30 \\ (69.0 $\pm$ 19.7)} 
  & \makecell{6/30 \\ (46.0 $\pm$ 0)} 
  & - 
  \\
ex-bw-b16-s2 &\wskip
  & \makecell{1/30 \\ (78.0)} 
  & \makecell{3/30 \\ (70.7 $\pm$ 30.4)} 
  & \makecell{3/30 \\ (62.0 $\pm$ 8.6)} 
  & - 
  \\
ex-bw-b17-s0 &\wskip
  & \makecell{1/30 \\ (76.0)} 
  & \makecell{5/30 \\ (62.4 $\pm$ 15.4)} 
  & \makecell{3/30 \\ (52.7 $\pm$ 20.1)} 
  & \makecell{3/30 \\ (80.0 $\pm$ 117.5)} 
  \\
ex-bw-b17-s1 &\wskip
  & \makecell{7/30 \\ (67.4 $\pm$ 30.3)} 
  & \makecell{2/30 \\ (66.0 $\pm$ 76.2)} 
  & \makecell{4/30 \\ (101.0 $\pm$ 84.9)} 
  & - 
  \\
ex-bw-b17-s2 &\wskip
  & \makecell{1/30 \\ (64.0)} 
  & \makecell{2/30 \\ (93.0 $\pm$ 165.2)} 
  & - 
  & - 
  \\
ex-bw-b18-s0 &\wskip
  & - 
  & \makecell{2/30 \\ (76.0 $\pm$ 25.4)} 
  & \makecell{4/30 \\ (70.5 $\pm$ 8.0)} 
  & - 
  \\
ex-bw-b18-s1 &\wskip
  & \makecell{2/30 \\ (53.0 $\pm$ 12.7)} 
  & \makecell{4/30 \\ (71.5 $\pm$ 22.0)} 
  & \makecell{2/30 \\ (72.0 $\pm$ 25.4)} 
  & - 
  \\
ex-bw-b18-s2 &\wskip
  & - 
  & \makecell{1/30 \\ (104.0)} 
  & - 
  & - 
  \\
ex-bw-b19-s0 &\wskip
  & - 
  & \makecell{8/30 \\ (99.0 $\pm$ 13.3)} 
  & \makecell{2/30 \\ (83.0 $\pm$ 368.5)} 
  & - 
  \\
ex-bw-b19-s1 &\wskip
  & - 
  & \makecell{1/30 \\ (96.0)} 
  & - 
  & - 
  \\
ex-bw-b19-s2 &\wskip
  & \makecell{10/30 \\ (57.8 $\pm$ 1.3)} 
  & \makecell{8/30 \\ (63.2 $\pm$ 4.6)} 
  & \makecell{17/30 \\ (52.0 $\pm$ 0)} 
  & - 
  \\
ex-bw-b20-s0 &\wskip
  & \makecell{1/30 \\ (70.0)} 
  & \makecell{2/30 \\ (78.0 $\pm$ 254.1)} 
  & \makecell{6/30 \\ (79.7 $\pm$ 27.1)} 
  & \makecell{4/30 \\ (81.5 $\pm$ 18.8)} 
  \\
ex-bw-b20-s1 &\wskip
  & - 
  & - 
  & \makecell{5/30 \\ (76.0 $\pm$ 12.0)} 
  & - 
  \\
ex-bw-b20-s2 &\wskip
  & - 
  & \makecell{1/30 \\ (86.0)} 
  & - 
  & - 
  \\

      \bottomrule
    \end{tabular}
  \end{center}
  \caption{
  Results for ASNets on Exploding Blocksworld.
  The first number in each problem name denotes the number of blocks.
}
\label{tab:prob-res-exbw-1}
\end{table}

\begin{table}[H]
  \begin{center}
    \smallscript
    \begin{tabular}{@{}lcccccc@{\vphantom{\makecell{1 \\ 2}}}}
    \toprule
    \multirow{2}{*}{Problem}\wskip&
      & \multicolumn{2}{c}{LRTDP}\wskip&
      & \multicolumn{2}{c}{SSiPP}
      \\
      \cmidrule{3-4} \cmidrule{6-7}
      &\wskip & h-add & LM-cut &\wskip & h-add & LM-cut\\
      \midrule
ex-bw-b11-s0 &\wskip
  & \makecell{21/30 \\ (44.4 $\pm$ 9.4)} 
  & \makecell{28/30 \\ (45.0 $\pm$ 7.3)} &\wskip 
  & \makecell{1/30 \\ (32.0)} 
  & \makecell{4/30 \\ (66.0 $\pm$ 55.2)} 
  \\
ex-bw-b11-s1 &\wskip
  & \makecell{4/30 \\ (99.5 $\pm$ 29.7)} 
  & - &\wskip 
  & - 
  & \makecell{1/30 \\ (64.0)} 
  \\
ex-bw-b11-s2 &\wskip
  & \makecell{23/30 \\ (28.4 $\pm$ 1.1)} 
  & \makecell{2/30 \\ (24.0 $\pm$ 0)} &\wskip 
  & \makecell{1/30 \\ (28.0)} 
  & \makecell{1/30 \\ (24.0)} 
  \\
ex-bw-b12-s0 &\wskip
  & \makecell{29/30 \\ (28.2 $\pm$ 1.5)} 
  & \makecell{9/30 \\ (26.0 $\pm$ 0)} &\wskip 
  & \makecell{28/30 \\ (27.7 $\pm$ 0.9)} 
  & \makecell{13/30 \\ (26.9 $\pm$ 1.2)} 
  \\
ex-bw-b12-s1 &\wskip
  & \makecell{1/30 \\ (52.0)} 
  & \makecell{6/30 \\ (67.0 $\pm$ 33.0)} &\wskip 
  & - 
  & - 
  \\
ex-bw-b12-s2 &\wskip
  & \makecell{1/30 \\ (42.0)} 
  & - &\wskip 
  & - 
  & - 
  \\
ex-bw-b13-s0 &\wskip
  & \makecell{27/30 \\ (27.9 $\pm$ 0.8)} 
  & \makecell{26/30 \\ (24.0 $\pm$ 0)} &\wskip 
  & \makecell{27/30 \\ (28.1 $\pm$ 1.1)} 
  & \makecell{21/30 \\ (24.0 $\pm$ 0)} 
  \\
ex-bw-b13-s1 &\wskip
  & \makecell{9/30 \\ (79.3 $\pm$ 22.6)} 
  & - &\wskip 
  & - 
  & - 
  \\
ex-bw-b13-s2 &\wskip
  & \makecell{14/30 \\ (63.1 $\pm$ 15.9)} 
  & \makecell{1/30 \\ (82.0)} &\wskip 
  & - 
  & - 
  \\
ex-bw-b14-s0 &\wskip
  & - 
  & - &\wskip 
  & - 
  & - 
  \\
ex-bw-b14-s1 &\wskip
  & - 
  & - &\wskip 
  & - 
  & - 
  \\
ex-bw-b14-s2 &\wskip
  & \makecell{10/30 \\ (55.4 $\pm$ 10.1)} 
  & \makecell{1/30 \\ (46.0)} &\wskip 
  & \makecell{1/30 \\ (80.0)} 
  & \makecell{1/30 \\ (68.0)} 
  \\
ex-bw-b15-s0 &\wskip
  & - 
  & - &\wskip 
  & - 
  & - 
  \\
ex-bw-b15-s1 &\wskip
  & - 
  & - &\wskip 
  & - 
  & - 
  \\
ex-bw-b15-s2 &\wskip
  & \makecell{5/30 \\ (73.6 $\pm$ 28.7)} 
  & - &\wskip 
  & \makecell{1/30 \\ (66.0)} 
  & \makecell{1/30 \\ (96.0)} 
  \\
ex-bw-b16-s0 &\wskip
  & - 
  & - &\wskip 
  & - 
  & - 
  \\
ex-bw-b16-s1 &\wskip
  & - 
  & - &\wskip 
  & \makecell{1/30 \\ (46.0)} 
  & - 
  \\
ex-bw-b16-s2 &\wskip
  & - 
  & - &\wskip 
  & - 
  & - 
  \\
ex-bw-b17-s0 &\wskip
  & \makecell{1/30 \\ (42.0)} 
  & - &\wskip 
  & - 
  & - 
  \\
ex-bw-b17-s1 &\wskip
  & - 
  & - &\wskip 
  & - 
  & - 
  \\
ex-bw-b17-s2 &\wskip
  & - 
  & - &\wskip 
  & - 
  & - 
  \\
ex-bw-b18-s0 &\wskip
  & - 
  & - &\wskip 
  & - 
  & - 
  \\
ex-bw-b18-s1 &\wskip
  & - 
  & - &\wskip 
  & - 
  & - 
  \\
ex-bw-b18-s2 &\wskip
  & - 
  & - &\wskip 
  & - 
  & - 
  \\
ex-bw-b19-s0 &\wskip
  & - 
  & - &\wskip 
  & - 
  & - 
  \\
ex-bw-b19-s1 &\wskip
  & - 
  & - &\wskip 
  & - 
  & - 
  \\
ex-bw-b19-s2 &\wskip
  & - 
  & - &\wskip 
  & - 
  & - 
  \\
ex-bw-b20-s0 &\wskip
  & - 
  & - &\wskip 
  & - 
  & - 
  \\
ex-bw-b20-s1 &\wskip
  & - 
  & - &\wskip 
  & - 
  & - 
  \\
ex-bw-b20-s2 &\wskip
  & - 
  & - &\wskip 
  & - 
  & - 
  \\

      \bottomrule
    \end{tabular}
  \end{center}
  \caption{Results for probabilistic baseline planners on Exploding Blocksworld.}
  \label{tab:prob-res-exbw-2}
\end{table}
\end{subtables}

%

\begin{subtables}
\begin{table}[H]
  \begin{center}
    \smallscript
    \begin{tabular}{@{}lccccc@{}}
    \toprule
    \multirow{2}{*}{Problem}\wskip&
      & \multicolumn{4}{c}{ASNet}
      \\
      \cmidrule{3-6}
      &\wskip & - & PE & Adm. & No h.\\
      \midrule
prob-bw-b15-s1 &\wskip
  & \makecell{30/30 \\ (49.4 $\pm$ 2.2)} 
  & \makecell{30/30 \\ (53.8 $\pm$ 2.7)} 
  & \makecell{30/30 \\ (48.9 $\pm$ 1.8)} 
  & - 
  \\
prob-bw-b15-s2 &\wskip
  & \makecell{30/30 \\ (42.2 $\pm$ 2.1)} 
  & \makecell{30/30 \\ (48.2 $\pm$ 2.8)} 
  & \makecell{30/30 \\ (42.8 $\pm$ 2.0)} 
  & \makecell{30/30 \\ (44.2 $\pm$ 2.1)} 
  \\
prob-bw-b15-s3 &\wskip
  & \makecell{30/30 \\ (66.0 $\pm$ 2.2)} 
  & \makecell{30/30 \\ (67.3 $\pm$ 2.2)} 
  & \makecell{30/30 \\ (63.8 $\pm$ 2.2)} 
  & \makecell{30/30 \\ (66.0 $\pm$ 2.2)} 
  \\
prob-bw-b15-s4 &\wskip
  & \makecell{30/30 \\ (48.6 $\pm$ 2.0)} 
  & \makecell{30/30 \\ (49.2 $\pm$ 2.0)} 
  & \makecell{30/30 \\ (48.4 $\pm$ 2.2)} 
  & \makecell{30/30 \\ (48.6 $\pm$ 2.0)} 
  \\
prob-bw-b15-s5 &\wskip
  & \makecell{30/30 \\ (52.3 $\pm$ 2.5)} 
  & \makecell{30/30 \\ (54.6 $\pm$ 2.5)} 
  & \makecell{30/30 \\ (46.1 $\pm$ 2.2)} 
  & \makecell{30/30 \\ (54.8 $\pm$ 2.4)} 
  \\
prob-bw-b20-s1 &\wskip
  & \makecell{30/30 \\ (73.5 $\pm$ 2.4)} 
  & \makecell{30/30 \\ (77.1 $\pm$ 3.0)} 
  & \makecell{30/30 \\ (73.5 $\pm$ 2.3)} 
  & \makecell{30/30 \\ (72.3 $\pm$ 2.4)} 
  \\
prob-bw-b20-s2 &\wskip
  & \makecell{30/30 \\ (68.5 $\pm$ 2.7)} 
  & \makecell{30/30 \\ (76.1 $\pm$ 2.5)} 
  & \makecell{30/30 \\ (69.2 $\pm$ 2.3)} 
  & \makecell{30/30 \\ (68.4 $\pm$ 2.6)} 
  \\
prob-bw-b20-s3 &\wskip
  & \makecell{30/30 \\ (68.7 $\pm$ 2.0)} 
  & \makecell{30/30 \\ (71.0 $\pm$ 3.2)} 
  & \makecell{30/30 \\ (68.7 $\pm$ 2.5)} 
  & \makecell{30/30 \\ (69.3 $\pm$ 2.5)} 
  \\
prob-bw-b20-s4 &\wskip
  & \makecell{30/30 \\ (76.3 $\pm$ 2.5)} 
  & \makecell{30/30 \\ (80.3 $\pm$ 2.7)} 
  & \makecell{30/30 \\ (73.1 $\pm$ 2.7)} 
  & \makecell{30/30 \\ (76.2 $\pm$ 2.4)} 
  \\
prob-bw-b20-s5 &\wskip
  & \makecell{30/30 \\ (84.2 $\pm$ 2.4)} 
  & \makecell{30/30 \\ (84.7 $\pm$ 2.4)} 
  & \makecell{30/30 \\ (83.0 $\pm$ 2.8)} 
  & \makecell{30/30 \\ (84.2 $\pm$ 2.4)} 
  \\
prob-bw-b25-s1 &\wskip
  & \makecell{30/30 \\ (90.4 $\pm$ 2.7)} 
  & \makecell{30/30 \\ (101.4 $\pm$ 3.4)} 
  & \makecell{30/30 \\ (93.7 $\pm$ 2.5)} 
  & \makecell{30/30 \\ (90.0 $\pm$ 2.7)} 
  \\
prob-bw-b25-s2 &\wskip
  & \makecell{30/30 \\ (90.1 $\pm$ 2.7)} 
  & \makecell{30/30 \\ (92.7 $\pm$ 2.3)} 
  & \makecell{30/30 \\ (88.1 $\pm$ 2.5)} 
  & \makecell{30/30 \\ (89.3 $\pm$ 2.5)} 
  \\
prob-bw-b25-s3 &\wskip
  & \makecell{30/30 \\ (84.5 $\pm$ 2.2)} 
  & \makecell{30/30 \\ (88.8 $\pm$ 3.1)} 
  & \makecell{30/30 \\ (77.8 $\pm$ 2.6)} 
  & - 
  \\
prob-bw-b25-s4 &\wskip
  & \makecell{30/30 \\ (101.0 $\pm$ 3.1)} 
  & \makecell{30/30 \\ (100.9 $\pm$ 3.3)} 
  & \makecell{30/30 \\ (91.9 $\pm$ 2.9)} 
  & \makecell{30/30 \\ (95.6 $\pm$ 2.6)} 
  \\
prob-bw-b25-s5 &\wskip
  & \makecell{30/30 \\ (91.7 $\pm$ 2.3)} 
  & \makecell{30/30 \\ (93.7 $\pm$ 2.5)} 
  & \makecell{30/30 \\ (89.4 $\pm$ 2.6)} 
  & \makecell{30/30 \\ (90.5 $\pm$ 2.8)} 
  \\
prob-bw-b30-s1 &\wskip
  & \makecell{30/30 \\ (106.5 $\pm$ 3.4)} 
  & \makecell{30/30 \\ (116.4 $\pm$ 3.6)} 
  & \makecell{30/30 \\ (102.7 $\pm$ 2.5)} 
  & \makecell{30/30 \\ (109.4 $\pm$ 3.0)} 
  \\
prob-bw-b30-s2 &\wskip
  & \makecell{30/30 \\ (101.4 $\pm$ 2.3)} 
  & \makecell{30/30 \\ (108.5 $\pm$ 3.1)} 
  & \makecell{30/30 \\ (103.7 $\pm$ 3.4)} 
  & \makecell{30/30 \\ (106.2 $\pm$ 2.7)} 
  \\
prob-bw-b30-s3 &\wskip
  & \makecell{30/30 \\ (111.4 $\pm$ 2.6)} 
  & \makecell{30/30 \\ (121.9 $\pm$ 3.4)} 
  & \makecell{30/30 \\ (110.0 $\pm$ 3.3)} 
  & \makecell{30/30 \\ (111.4 $\pm$ 2.8)} 
  \\
prob-bw-b30-s4 &\wskip
  & \makecell{30/30 \\ (111.5 $\pm$ 3.5)} 
  & \makecell{30/30 \\ (115.2 $\pm$ 3.8)} 
  & \makecell{30/30 \\ (108.6 $\pm$ 2.6)} 
  & \makecell{30/30 \\ (113.5 $\pm$ 3.0)} 
  \\
prob-bw-b30-s5 &\wskip
  & \makecell{30/30 \\ (104.6 $\pm$ 3.5)} 
  & \makecell{30/30 \\ (110.4 $\pm$ 3.2)} 
  & \makecell{30/30 \\ (104.8 $\pm$ 3.2)} 
  & \makecell{30/30 \\ (111.3 $\pm$ 2.4)} 
  \\
prob-bw-b35-s1 &\wskip
  & \makecell{30/30 \\ (131.6 $\pm$ 3.8)} 
  & \makecell{30/30 \\ (140.3 $\pm$ 4.5)} 
  & \makecell{30/30 \\ (132.3 $\pm$ 3.4)} 
  & \makecell{30/30 \\ (132.8 $\pm$ 3.1)} 
  \\
prob-bw-b35-s2 &\wskip
  & \makecell{30/30 \\ (137.9 $\pm$ 3.8)} 
  & \makecell{30/30 \\ (147.8 $\pm$ 3.9)} 
  & \makecell{30/30 \\ (135.1 $\pm$ 4.1)} 
  & \makecell{30/30 \\ (143.7 $\pm$ 3.2)} 
  \\
prob-bw-b35-s3 &\wskip
  & \makecell{30/30 \\ (120.2 $\pm$ 3.0)} 
  & \makecell{30/30 \\ (131.4 $\pm$ 4.2)} 
  & \makecell{30/30 \\ (115.7 $\pm$ 3.0)} 
  & \makecell{30/30 \\ (125.1 $\pm$ 3.0)} 
  \\
prob-bw-b35-s4 &\wskip
  & \makecell{30/30 \\ (121.3 $\pm$ 2.8)} 
  & \makecell{30/30 \\ (131.5 $\pm$ 3.7)} 
  & \makecell{30/30 \\ (123.6 $\pm$ 3.4)} 
  & \makecell{30/30 \\ (125.3 $\pm$ 3.2)} 
  \\
prob-bw-b35-s5 &\wskip
  & \makecell{30/30 \\ (133.6 $\pm$ 3.5)} 
  & \makecell{30/30 \\ (138.5 $\pm$ 3.6)} 
  & \makecell{30/30 \\ (134.4 $\pm$ 3.1)} 
  & \makecell{30/30 \\ (130.3 $\pm$ 3.3)} 
  \\
prob-bw-b40-s1 &\wskip
  & \makecell{30/30 \\ (127.0 $\pm$ 3.5)} 
  & \makecell{30/30 \\ (140.9 $\pm$ 5.1)} 
  & \makecell{30/30 \\ (123.6 $\pm$ 2.9)} 
  & \makecell{30/30 \\ (133.8 $\pm$ 3.0)} 
  \\
prob-bw-b40-s2 &\wskip
  & \makecell{30/30 \\ (147.1 $\pm$ 3.8)} 
  & \makecell{30/30 \\ (153.4 $\pm$ 3.1)} 
  & \makecell{30/30 \\ (144.4 $\pm$ 3.3)} 
  & \makecell{30/30 \\ (149.9 $\pm$ 3.7)} 
  \\
prob-bw-b40-s3 &\wskip
  & \makecell{30/30 \\ (146.8 $\pm$ 4.1)} 
  & \makecell{30/30 \\ (160.0 $\pm$ 4.1)} 
  & \makecell{30/30 \\ (142.1 $\pm$ 3.8)} 
  & \makecell{30/30 \\ (154.0 $\pm$ 4.2)} 
  \\
prob-bw-b40-s4 &\wskip
  & \makecell{30/30 \\ (152.2 $\pm$ 3.4)} 
  & \makecell{30/30 \\ (155.2 $\pm$ 4.1)} 
  & \makecell{30/30 \\ (144.8 $\pm$ 3.3)} 
  & \makecell{30/30 \\ (151.8 $\pm$ 4.3)} 
  \\
prob-bw-b40-s5 &\wskip
  & \makecell{30/30 \\ (144.9 $\pm$ 3.5)} 
  & \makecell{30/30 \\ (149.5 $\pm$ 3.9)} 
  & \makecell{30/30 \\ (143.1 $\pm$ 3.0)} 
  & \makecell{30/30 \\ (145.8 $\pm$ 4.2)} 
  \\

      \bottomrule
    \end{tabular}
  \end{center}
  \caption{
  Results for ASNets on Probabilistic Blocksworld.
  The first number in each problem name denotes the number of blocks.
}
\label{tab:prob-res-pbw-1}
\end{table}

\begin{table}[H]
  \begin{center}
    \smallscript
    \begin{tabular}{@{}lcccccc@{\vphantom{\makecell{1 \\ 2}}}}
    \toprule
    \multirow{2}{*}{Problem}\wskip&
      & \multicolumn{2}{c}{LRTDP}\wskip&
      & \multicolumn{2}{c}{SSiPP}
      \\
      \cmidrule{3-4} \cmidrule{6-7}
      &\wskip & h-add & LM-cut &\wskip & h-add & LM-cut\\
      \midrule
prob-bw-b15-s1 &\wskip
  & \makecell{30/30 \\ (52.9 $\pm$ 2.5)} 
  & - &\wskip 
  & \makecell{20/30 \\ (110.4 $\pm$ 23.0)} 
  & \makecell{1/30 \\ (241.0)} 
  \\
prob-bw-b15-s2 &\wskip
  & \makecell{30/30 \\ (48.3 $\pm$ 2.3)} 
  & - &\wskip 
  & \makecell{30/30 \\ (114.1 $\pm$ 21.5)} 
  & - 
  \\
prob-bw-b15-s3 &\wskip
  & \makecell{30/30 \\ (74.6 $\pm$ 2.7)} 
  & - &\wskip 
  & \makecell{30/30 \\ (70.7 $\pm$ 3.0)} 
  & - 
  \\
prob-bw-b15-s4 &\wskip
  & \makecell{30/30 \\ (58.3 $\pm$ 4.2)} 
  & - &\wskip 
  & \makecell{19/30 \\ (107.9 $\pm$ 20.7)} 
  & \makecell{1/30 \\ (168.0)} 
  \\
prob-bw-b15-s5 &\wskip
  & \makecell{30/30 \\ (55.1 $\pm$ 1.5)} 
  & - &\wskip 
  & \makecell{19/30 \\ (140.8 $\pm$ 29.4)} 
  & - 
  \\
prob-bw-b20-s1 &\wskip
  & \makecell{12/30 \\ (81.1 $\pm$ 6.9)} 
  & - &\wskip 
  & \makecell{1/30 \\ (91.0)} 
  & - 
  \\
prob-bw-b20-s2 &\wskip
  & \makecell{30/30 \\ (77.0 $\pm$ 2.9)} 
  & - &\wskip 
  & \makecell{1/30 \\ (79.0)} 
  & \makecell{1/30 \\ (89.0)} 
  \\
prob-bw-b20-s3 &\wskip
  & \makecell{30/30 \\ (78.2 $\pm$ 2.6)} 
  & - &\wskip 
  & \makecell{4/30 \\ (74.2 $\pm$ 8.5)} 
  & - 
  \\
prob-bw-b20-s4 &\wskip
  & \makecell{28/30 \\ (87.6 $\pm$ 3.9)} 
  & - &\wskip 
  & \makecell{1/30 \\ (73.0)} 
  & - 
  \\
prob-bw-b20-s5 &\wskip
  & \makecell{19/30 \\ (85.4 $\pm$ 3.7)} 
  & - &\wskip 
  & \makecell{12/30 \\ (85.1 $\pm$ 3.2)} 
  & - 
  \\
prob-bw-b25-s1 &\wskip
  & \makecell{1/30 \\ (79.0)} 
  & - &\wskip 
  & \makecell{4/30 \\ (135.2 $\pm$ 120.6)} 
  & - 
  \\
prob-bw-b25-s2 &\wskip
  & - 
  & - &\wskip 
  & \makecell{1/30 \\ (88.0)} 
  & - 
  \\
prob-bw-b25-s3 &\wskip
  & - 
  & - &\wskip 
  & \makecell{2/30 \\ (106.5 $\pm$ 184.2)} 
  & - 
  \\
prob-bw-b25-s4 &\wskip
  & \makecell{2/30 \\ (106.0 $\pm$ 76.2)} 
  & - &\wskip 
  & \makecell{1/30 \\ (162.0)} 
  & - 
  \\
prob-bw-b25-s5 &\wskip
  & - 
  & - &\wskip 
  & - 
  & - 
  \\
prob-bw-b30-s1 &\wskip
  & - 
  & - &\wskip 
  & - 
  & - 
  \\
prob-bw-b30-s2 &\wskip
  & - 
  & - &\wskip 
  & \makecell{1/30 \\ (148.0)} 
  & - 
  \\
prob-bw-b30-s3 &\wskip
  & \makecell{1/30 \\ (135.0)} 
  & - &\wskip 
  & - 
  & - 
  \\
prob-bw-b30-s4 &\wskip
  & - 
  & - &\wskip 
  & \makecell{1/30 \\ (190.0)} 
  & - 
  \\
prob-bw-b30-s5 &\wskip
  & - 
  & - &\wskip 
  & - 
  & - 
  \\
prob-bw-b35-s1 &\wskip
  & - 
  & - &\wskip 
  & - 
  & - 
  \\
prob-bw-b35-s2 &\wskip
  & - 
  & - &\wskip 
  & - 
  & - 
  \\
prob-bw-b35-s3 &\wskip
  & - 
  & - &\wskip 
  & - 
  & - 
  \\
prob-bw-b35-s4 &\wskip
  & - 
  & - &\wskip 
  & - 
  & - 
  \\
prob-bw-b35-s5 &\wskip
  & - 
  & - &\wskip 
  & - 
  & - 
  \\
prob-bw-b40-s1 &\wskip
  & - 
  & - &\wskip 
  & - 
  & - 
  \\
prob-bw-b40-s2 &\wskip
  & - 
  & - &\wskip 
  & - 
  & - 
  \\
prob-bw-b40-s3 &\wskip
  & - 
  & - &\wskip 
  & - 
  & - 
  \\
prob-bw-b40-s4 &\wskip
  & - 
  & - &\wskip 
  & - 
  & - 
  \\
prob-bw-b40-s5 &\wskip
  & - 
  & - &\wskip 
  & - 
  & - 
  \\

      \bottomrule
    \end{tabular}
  \end{center}
  \caption{Results for probabilistic baseline planners on Probabilistic Blocksworld.}
  \label{tab:prob-res-pbw-2}
\end{table}
\end{subtables}


\begin{subtables}
\begin{table}[H]
  \begin{center}
    \footnotesize
    \begin{tabular}{@{}lccccc@{}}
    \toprule
    \multirow{2}{*}{Problem}\wskip&
      & \multicolumn{4}{c}{ASNet}
      \\
      \cmidrule{3-6}
      &\wskip & - & PE & Adm. & No h.\\
      \midrule
triangle-tire-4 &\wskip
  & \makecell{30/30 \\ (23.4 $\pm$ 0.7)} 
  & \makecell{30/30 \\ (23.4 $\pm$ 0.7)} 
  & \makecell{30/30 \\ (23.4 $\pm$ 0.7)} 
  & \makecell{30/30 \\ (23.4 $\pm$ 0.7)} 
  \\
triangle-tire-5 &\wskip
  & \makecell{30/30 \\ (28.9 $\pm$ 0.8)} 
  & \makecell{30/30 \\ (28.9 $\pm$ 0.8)} 
  & \makecell{30/30 \\ (28.9 $\pm$ 0.8)} 
  & \makecell{30/30 \\ (28.9 $\pm$ 0.8)} 
  \\
triangle-tire-6 &\wskip
  & \makecell{30/30 \\ (34.9 $\pm$ 0.9)} 
  & \makecell{30/30 \\ (34.9 $\pm$ 0.9)} 
  & \makecell{30/30 \\ (34.9 $\pm$ 0.9)} 
  & \makecell{30/30 \\ (34.9 $\pm$ 0.9)} 
  \\
triangle-tire-7 &\wskip
  & \makecell{30/30 \\ (40.8 $\pm$ 0.9)} 
  & \makecell{30/30 \\ (40.8 $\pm$ 0.9)} 
  & \makecell{30/30 \\ (40.8 $\pm$ 0.9)} 
  & \makecell{30/30 \\ (40.8 $\pm$ 0.9)} 
  \\
triangle-tire-8 &\wskip
  & \makecell{30/30 \\ (46.8 $\pm$ 1.1)} 
  & \makecell{30/30 \\ (46.8 $\pm$ 1.1)} 
  & \makecell{30/30 \\ (46.8 $\pm$ 1.1)} 
  & \makecell{30/30 \\ (46.8 $\pm$ 1.1)} 
  \\
triangle-tire-9 &\wskip
  & \makecell{30/30 \\ (52.9 $\pm$ 1.3)} 
  & \makecell{30/30 \\ (52.9 $\pm$ 1.3)} 
  & \makecell{30/30 \\ (52.9 $\pm$ 1.3)} 
  & \makecell{30/30 \\ (52.9 $\pm$ 1.3)} 
  \\
triangle-tire-10 &\wskip
  & \makecell{30/30 \\ (59.0 $\pm$ 1.1)} 
  & \makecell{30/30 \\ (59.0 $\pm$ 1.1)} 
  & \makecell{30/30 \\ (59.0 $\pm$ 1.1)} 
  & \makecell{30/30 \\ (59.0 $\pm$ 1.1)} 
  \\
triangle-tire-11 &\wskip
  & \makecell{30/30 \\ (64.8 $\pm$ 1.1)} 
  & \makecell{30/30 \\ (64.8 $\pm$ 1.1)} 
  & \makecell{30/30 \\ (64.8 $\pm$ 1.1)} 
  & \makecell{30/30 \\ (64.8 $\pm$ 1.1)} 
  \\
triangle-tire-12 &\wskip
  & \makecell{30/30 \\ (71.1 $\pm$ 1.2)} 
  & \makecell{30/30 \\ (71.1 $\pm$ 1.2)} 
  & \makecell{30/30 \\ (71.1 $\pm$ 1.2)} 
  & \makecell{30/30 \\ (71.1 $\pm$ 1.2)} 
  \\
triangle-tire-13 &\wskip
  & \makecell{30/30 \\ (76.9 $\pm$ 1.2)} 
  & \makecell{30/30 \\ (76.9 $\pm$ 1.2)} 
  & \makecell{30/30 \\ (76.9 $\pm$ 1.2)} 
  & \makecell{30/30 \\ (76.9 $\pm$ 1.2)} 
  \\
triangle-tire-14 &\wskip
  & \makecell{30/30 \\ (82.8 $\pm$ 1.3)} 
  & \makecell{30/30 \\ (82.8 $\pm$ 1.3)} 
  & \makecell{30/30 \\ (82.8 $\pm$ 1.3)} 
  & \makecell{30/30 \\ (82.8 $\pm$ 1.3)} 
  \\
triangle-tire-15 &\wskip
  & \makecell{30/30 \\ (88.7 $\pm$ 1.4)} 
  & \makecell{30/30 \\ (88.7 $\pm$ 1.4)} 
  & \makecell{30/30 \\ (88.7 $\pm$ 1.4)} 
  & \makecell{30/30 \\ (88.7 $\pm$ 1.4)} 
  \\
triangle-tire-16 &\wskip
  & \makecell{30/30 \\ (94.8 $\pm$ 1.3)} 
  & \makecell{30/30 \\ (94.8 $\pm$ 1.3)} 
  & \makecell{30/30 \\ (94.8 $\pm$ 1.3)} 
  & \makecell{30/30 \\ (94.8 $\pm$ 1.3)} 
  \\
triangle-tire-17 &\wskip
  & \makecell{30/30 \\ (100.8 $\pm$ 1.2)} 
  & \makecell{30/30 \\ (100.8 $\pm$ 1.2)} 
  & \makecell{30/30 \\ (100.8 $\pm$ 1.2)} 
  & \makecell{30/30 \\ (100.8 $\pm$ 1.2)} 
  \\
triangle-tire-18 &\wskip
  & \makecell{30/30 \\ (106.5 $\pm$ 1.4)} 
  & \makecell{30/30 \\ (106.5 $\pm$ 1.4)} 
  & \makecell{30/30 \\ (106.5 $\pm$ 1.4)} 
  & \makecell{30/30 \\ (106.5 $\pm$ 1.4)} 
  \\
triangle-tire-19 &\wskip
  & \makecell{30/30 \\ (112.5 $\pm$ 1.6)} 
  & \makecell{30/30 \\ (112.5 $\pm$ 1.6)} 
  & \makecell{30/30 \\ (112.5 $\pm$ 1.6)} 
  & \makecell{30/30 \\ (112.5 $\pm$ 1.6)} 
  \\
triangle-tire-20 &\wskip
  & \makecell{30/30 \\ (118.4 $\pm$ 1.5)} 
  & \makecell{30/30 \\ (118.4 $\pm$ 1.5)} 
  & \makecell{30/30 \\ (118.4 $\pm$ 1.5)} 
  & \makecell{30/30 \\ (118.4 $\pm$ 1.5)} 
  \\

      \bottomrule
    \end{tabular}
  \end{center}
  \caption{
  Results for ASNets on Triangle Tireworld.
  Problems are numbered using the same convention as \citeA{little2007probabilistic}.
}
\label{tab:prob-res-ttw-1}
\end{table}

\begin{table}[H]
  \begin{center}
    \footnotesize
    \begin{tabular}{@{}lcccccc@{\vphantom{\makecell{1 \\ 2}}}}
    \toprule
    \multirow{2}{*}{Problem}\wskip&
      & \multicolumn{2}{c}{LRTDP}\wskip&
      & \multicolumn{2}{c}{SSiPP}
      \\
      \cmidrule{3-4} \cmidrule{6-7}
      &\wskip & h-add & LM-cut &\wskip & h-add & LM-cut\\
      \midrule
triangle-tire-4 &\wskip
  & \makecell{30/30 \\ (23.7 $\pm$ 0.7)} 
  & \makecell{30/30 \\ (23.8 $\pm$ 0.6)} &\wskip 
  & \makecell{30/30 \\ (24.0 $\pm$ 0.7)} 
  & \makecell{30/30 \\ (23.2 $\pm$ 0.7)} 
  \\
triangle-tire-5 &\wskip
  & \makecell{30/30 \\ (30.7 $\pm$ 0.8)} 
  & \makecell{19/30 \\ (29.1 $\pm$ 1.1)} &\wskip 
  & \makecell{29/30 \\ (30.8 $\pm$ 0.9)} 
  & \makecell{30/30 \\ (30.2 $\pm$ 0.9)} 
  \\
triangle-tire-6 &\wskip
  & \makecell{1/30 \\ (34.0)} 
  & \makecell{1/30 \\ (34.0)} &\wskip 
  & \makecell{30/30 \\ (38.0 $\pm$ 0.8)} 
  & \makecell{30/30 \\ (35.7 $\pm$ 0.8)} 
  \\
triangle-tire-7 &\wskip
  & \makecell{1/30 \\ (42.0)} 
  & \makecell{1/30 \\ (45.0)} &\wskip 
  & \makecell{27/30 \\ (45.5 $\pm$ 1.0)} 
  & \makecell{16/30 \\ (41.7 $\pm$ 1.3)} 
  \\
triangle-tire-8 &\wskip
  & \makecell{1/30 \\ (53.0)} 
  & \makecell{1/30 \\ (50.0)} &\wskip 
  & \makecell{15/30 \\ (51.7 $\pm$ 1.2)} 
  & \makecell{3/30 \\ (48.3 $\pm$ 7.6)} 
  \\
triangle-tire-9 &\wskip
  & \makecell{1/30 \\ (59.0)} 
  & - &\wskip 
  & \makecell{3/30 \\ (60.7 $\pm$ 3.8)} 
  & \makecell{1/30 \\ (55.0)} 
  \\
triangle-tire-10 &\wskip
  & \makecell{1/30 \\ (69.0)} 
  & - &\wskip 
  & \makecell{1/30 \\ (67.0)} 
  & - 
  \\
triangle-tire-11 &\wskip
  & \makecell{1/30 \\ (75.0)} 
  & - &\wskip 
  & \makecell{1/30 \\ (77.0)} 
  & - 
  \\
triangle-tire-12 &\wskip
  & - 
  & - &\wskip 
  & \makecell{1/30 \\ (79.0)} 
  & - 
  \\
triangle-tire-13 &\wskip
  & - 
  & - &\wskip 
  & - 
  & - 
  \\
triangle-tire-14 &\wskip
  & - 
  & - &\wskip 
  & - 
  & - 
  \\
triangle-tire-15 &\wskip
  & - 
  & - &\wskip 
  & - 
  & - 
  \\
triangle-tire-16 &\wskip
  & - 
  & - &\wskip 
  & - 
  & - 
  \\
triangle-tire-17 &\wskip
  & - 
  & - &\wskip 
  & - 
  & - 
  \\
triangle-tire-18 &\wskip
  & - 
  & - &\wskip 
  & - 
  & - 
  \\
triangle-tire-19 &\wskip
  & - 
  & - &\wskip 
  & - 
  & - 
  \\
triangle-tire-20 &\wskip
  & - 
  & - &\wskip 
  & - 
  & - 
  \\

      \bottomrule
    \end{tabular}
  \end{center}
  \caption{Results for probabilistic baseline planners on Triangle Tireworld.}
  \label{tab:prob-res-ttw-2}
\end{table}
\end{subtables}


  \begin{table}[htp]
    \begin{center}
      \begin{scriptsize}
        \begin{tabular}{@{}lccccccccccccc@{}}
        \toprule
        \multirow{2}{*}{Problem}\wskip&
          & \multicolumn{3}{c}{ASNet} \wskip&
          & \multicolumn{2}{c}{A*}\wskip&
          & \multicolumn{1}{c}{GBF}\wskip&
          & \multicolumn{2}{c}{LAMA}
          \\
          \cmidrule{3-5} \cmidrule{7-8} \cmidrule{10-10} \cmidrule{12-13}
          &\wskip & - & PE & No h.
          &\wskip & LM-cut & LM-count
          &\wskip & LM-cut
          &\wskip & -2011 & -first
          \\
          \midrule
blocks-n35-s1 &\wskip
  & 110 
  & 119 (10/10) 
  & 108 &\wskip 
  & - 
  & - &\wskip 
  & - &\wskip 
  & 138 
  & 138 
  \\
blocks-n35-s2 &\wskip
  & 108 
  & 120.2 (10/10) 
  & 112 &\wskip 
  & - 
  & - &\wskip 
  & - &\wskip 
  & 108 
  & 162 
  \\
blocks-n35-s3 &\wskip
  & 98 
  & 111.6 (10/10) 
  & 100 &\wskip 
  & - 
  & - &\wskip 
  & - &\wskip 
  & 108 
  & 170 
  \\
blocks-n35-s4 &\wskip
  & 110 
  & 125.2 (10/10) 
  & 120 &\wskip 
  & - 
  & - &\wskip 
  & - &\wskip 
  & 134 
  & 228 
  \\
blocks-n35-s5 &\wskip
  & 114 
  & 133.6 (10/10) 
  & 108 &\wskip 
  & - 
  & - &\wskip 
  & - &\wskip 
  & 146 
  & 166 
  \\
blocks-n35-s6 &\wskip
  & 102 
  & 111 (10/10) 
  & 102 &\wskip 
  & - 
  & - &\wskip 
  & - &\wskip 
  & 110 
  & 110 
  \\
blocks-n35-s7 &\wskip
  & 96 
  & 97 (10/10) 
  & 96 &\wskip 
  & - 
  & - &\wskip 
  & 262 &\wskip 
  & 102 
  & 118 
  \\
blocks-n35-s8 &\wskip
  & 108 
  & 114.8 (10/10) 
  & 106 &\wskip 
  & - 
  & - &\wskip 
  & 290 &\wskip 
  & 114 
  & 160 
  \\
blocks-n35-s9 &\wskip
  & 106 
  & 125.8 (10/10) 
  & 114 &\wskip 
  & - 
  & - &\wskip 
  & - &\wskip 
  & 114 
  & 116 
  \\
blocks-n35-s10 &\wskip
  & 106 
  & 113.6 (10/10) 
  & 104 &\wskip 
  & - 
  & - &\wskip 
  & - &\wskip 
  & 116 
  & 116 
  \\
blocks-n50-s1 &\wskip
  & 158 
  & 181 (10/10) 
  & - &\wskip 
  & - 
  & - &\wskip 
  & - &\wskip 
  & - 
  & - 
  \\
blocks-n50-s2 &\wskip
  & 166 
  & 199.8 (10/10) 
  & - &\wskip 
  & - 
  & - &\wskip 
  & - &\wskip 
  & - 
  & - 
  \\
blocks-n50-s3 &\wskip
  & 164 
  & 198 (10/10) 
  & - &\wskip 
  & - 
  & - &\wskip 
  & - &\wskip 
  & 216 
  & 240 
  \\
blocks-n50-s4 &\wskip
  & 160 
  & 185 (10/10) 
  & 162 &\wskip 
  & - 
  & - &\wskip 
  & - &\wskip 
  & - 
  & - 
  \\
blocks-n50-s5 &\wskip
  & 158 
  & 182.6 (10/10) 
  & 170 &\wskip 
  & - 
  & - &\wskip 
  & - &\wskip 
  & 240 
  & 278 
  \\
blocks-n50-s6 &\wskip
  & 134 
  & 155.6 (10/10) 
  & - &\wskip 
  & - 
  & - &\wskip 
  & - &\wskip 
  & 202 
  & 202 
  \\
blocks-n50-s7 &\wskip
  & 162 
  & 182.6 (10/10) 
  & 160 &\wskip 
  & - 
  & - &\wskip 
  & - &\wskip 
  & - 
  & - 
  \\
blocks-n50-s8 &\wskip
  & 154 
  & 182.2 (10/10) 
  & 158 &\wskip 
  & - 
  & - &\wskip 
  & - &\wskip 
  & - 
  & - 
  \\
blocks-n50-s9 &\wskip
  & 184 
  & 237 (10/10) 
  & 188 &\wskip 
  & - 
  & - &\wskip 
  & - &\wskip 
  & - 
  & - 
  \\
blocks-n50-s10 &\wskip
  & 146 
  & 162.4 (10/10) 
  & 154 &\wskip 
  & - 
  & - &\wskip 
  & - &\wskip 
  & - 
  & - 
  \\
blocks-n50-s11 &\wskip
  & 158 
  & 197.4 (10/10) 
  & 172 &\wskip 
  & - 
  & - &\wskip 
  & - &\wskip 
  & - 
  & - 
  \\
blocks-n50-s12 &\wskip
  & 150 
  & 161 (10/10) 
  & 146 &\wskip 
  & - 
  & - &\wskip 
  & - &\wskip 
  & 200 
  & 236 
  \\
blocks-n50-s13 &\wskip
  & 136 
  & 155.4 (10/10) 
  & 134 &\wskip 
  & - 
  & - &\wskip 
  & - &\wskip 
  & 208 
  & 286 
  \\
blocks-n50-s14 &\wskip
  & 158 
  & 179 (10/10) 
  & 166 &\wskip 
  & - 
  & - &\wskip 
  & - &\wskip 
  & - 
  & - 
  \\
blocks-n50-s15 &\wskip
  & 170 
  & 219.8 (10/10) 
  & 190 &\wskip 
  & - 
  & - &\wskip 
  & - &\wskip 
  & - 
  & - 
  \\
blocks-n50-s16 &\wskip
  & 138 
  & 156.2 (10/10) 
  & 142 &\wskip 
  & - 
  & - &\wskip 
  & - &\wskip 
  & 154 
  & 154 
  \\
blocks-n50-s17 &\wskip
  & 168 
  & 207.8 (10/10) 
  & - &\wskip 
  & - 
  & - &\wskip 
  & - &\wskip 
  & 172 
  & 172 
  \\
blocks-n50-s18 &\wskip
  & 160 
  & 201.4 (10/10) 
  & - &\wskip 
  & - 
  & - &\wskip 
  & - &\wskip 
  & - 
  & - 
  \\
blocks-n50-s19 &\wskip
  & 148 
  & 197.4 (10/10) 
  & 152 &\wskip 
  & - 
  & - &\wskip 
  & - &\wskip 
  & - 
  & - 
  \\
blocks-n50-s20 &\wskip
  & 152 
  & 169.8 (10/10) 
  & 156 &\wskip 
  & - 
  & - &\wskip 
  & - &\wskip 
  & 154 
  & 154 
  \\
          \bottomrule
        \end{tabular}
      \end{scriptsize}
    \end{center}
              \caption{
    Results for both ASNet and the baseline planners on our evaluation set of (deterministic) Blocksworld problems.
    Each cell shows the length of the plan returned by the corresponding planner.
    For ASNets in stochastic execution mode (``PE''), 10 rollouts are performed,
    and we report both the mean cost of successful plans (first number) and
    the fraction of runs that reach the goal (second number).
    The first number in each problem name indicates the number of blocks, and
    the second identifies the seed used to generate the instance.
  }
  \label{tab:det-res-bw}
  \end{table}

  \begin{table}
    \begin{center}
      \begin{scriptsize}
        \begin{tabular}{@{}l@{}ccccccccccccc@{}}
        \toprule
        \multirow{2}{*}{Problem}\wskip&
          & \multicolumn{3}{c}{ASNet} \wskip&
          & \multicolumn{2}{c}{A*}\wskip&
          & \multicolumn{1}{c}{GBF}\wskip&
          & \multicolumn{2}{c}{LAMA}
          \\
          \cmidrule{3-5} \cmidrule{7-8} \cmidrule{10-10} \cmidrule{12-13}
          &\wskip & - & PE & No h.
          &\wskip & LM-cut & LM-count
          &\wskip & LM-cut
          &\wskip & -2011 & -first
          \\
          \midrule
mbw-b15-t1-s0 &\wskip
  & 52 
  & 59.2 (8/10) 
  & 54 &\wskip 
  & - 
  & - &\wskip 
  & 104 &\wskip 
  & 52 
  & 142 
  \\
mbw-b15-t3-s1 &\wskip
  & 38 
  & 50.7 (6/10) 
  & - &\wskip 
  & - 
  & 38 &\wskip 
  & 94 &\wskip 
  & 38 
  & 78 
  \\
mbw-b15-t3-s2 &\wskip
  & 42 
  & 57 (10/10) 
  & 46 &\wskip 
  & - 
  & 38 &\wskip 
  & 66 &\wskip 
  & 38 
  & 72 
  \\
mbw-b20-t1-s0 &\wskip
  & 68 
  & 87.7 (6/10) 
  & - &\wskip 
  & - 
  & - &\wskip 
  & 292 &\wskip 
  & 82 
  & 114 
  \\
mbw-b20-t4-s1 &\wskip
  & 60 
  & 70.9 (9/10) 
  & - &\wskip 
  & - 
  & - &\wskip 
  & 136 &\wskip 
  & 68 
  & 104 
  \\
mbw-b20-t4-s2 &\wskip
  & 56 
  & 70 (6/10) 
  & - &\wskip 
  & - 
  & - &\wskip 
  & 102 &\wskip 
  & 68 
  & 140 
  \\
mbw-b25-t1-s0 &\wskip
  & 98 
  & 135.4 (7/10) 
  & - &\wskip 
  & - 
  & - &\wskip 
  & - &\wskip 
  & 116 
  & 214 
  \\
mbw-b25-t2-s1 &\wskip
  & 90 
  & 106.9 (9/10) 
  & - &\wskip 
  & - 
  & - &\wskip 
  & - &\wskip 
  & 106 
  & 212 
  \\
mbw-b25-t5-s2 &\wskip
  & 66 
  & 101 (9/10) 
  & 64 &\wskip 
  & - 
  & - &\wskip 
  & - &\wskip 
  & - 
  & - 
  \\
mbw-b30-t1-s0 &\wskip
  & 172 
  & 150.8 (10/10) 
  & - &\wskip 
  & - 
  & - &\wskip 
  & - &\wskip 
  & - 
  & - 
  \\
mbw-b30-t5-s1 &\wskip
  & 258 
  & 122.8 (10/10) 
  & - &\wskip 
  & - 
  & - &\wskip 
  & - &\wskip 
  & - 
  & - 
  \\
mbw-b30-t9-s2 &\wskip
  & 74 
  & 99.5 (10/10) 
  & - &\wskip 
  & - 
  & - &\wskip 
  & - &\wskip 
  & 86 
  & 126 
  \\
mbw-b35-t1-s0 &\wskip
  & 162 
  & 186.7 (9/10) 
  & - &\wskip 
  & - 
  & - &\wskip 
  & - &\wskip 
  & - 
  & - 
  \\
mbw-b35-t5-s1 &\wskip
  & 138 
  & 165.4 (7/10) 
  & - &\wskip 
  & - 
  & - &\wskip 
  & - &\wskip 
  & 140 
  & 156 
  \\
mbw-b35-t8-s2 &\wskip
  & 100 
  & 123.6 (9/10) 
  & - &\wskip 
  & - 
  & - &\wskip 
  & - &\wskip 
  & - 
  & - 
  \\
mbw-b40-t1-s0 &\wskip
  & 178 
  & 208.6 (10/10) 
  & 158 &\wskip 
  & - 
  & - &\wskip 
  & - &\wskip 
  & - 
  & - 
  \\
mbw-b40-t10-s1 &\wskip
  & 110 
  & 133.2 (6/10) 
  & - &\wskip 
  & - 
  & - &\wskip 
  & - &\wskip 
  & - 
  & - 
  \\
mbw-b40-t6-s2 &\wskip
  & 108 
  & 162.3 (6/10) 
  & - &\wskip 
  & - 
  & - &\wskip 
  & - &\wskip 
  & - 
  & - 
  \\
mbw-b45-t1-s0 &\wskip
  & 204 
  & 269.2 (5/10) 
  & - &\wskip 
  & - 
  & - &\wskip 
  & - &\wskip 
  & - 
  & - 
  \\
mbw-b45-t2-s1 &\wskip
  & 182 
  & 230 (5/10) 
  & - &\wskip 
  & - 
  & - &\wskip 
  & - &\wskip 
  & - 
  & - 
  \\
mbw-b45-t6-s2 &\wskip
  & 156 
  & 220.2 (8/10) 
  & - &\wskip 
  & - 
  & - &\wskip 
  & - &\wskip 
  & - 
  & - 
  \\
mbw-b50-t1-s0 &\wskip
  & - 
  & 280 (2/10) 
  & 210 &\wskip 
  & - 
  & - &\wskip 
  & - &\wskip 
  & - 
  & - 
  \\
mbw-b50-t4-s1 &\wskip
  & 270 
  & 243.7 (6/10) 
  & - &\wskip 
  & - 
  & - &\wskip 
  & - &\wskip 
  & - 
  & - 
  \\
mbw-b50-t7-s2 &\wskip
  & 156 
  & 244.3 (6/10) 
  & - &\wskip 
  & - 
  & - &\wskip 
  & - &\wskip 
  & 180 
  & 208 
  \\
mbw-b55-t1-s0 &\wskip
  & - 
  & - 
  & - &\wskip 
  & - 
  & - &\wskip 
  & - &\wskip 
  & - 
  & - 
  \\
mbw-b55-t13-s1 &\wskip
  & 146 
  & 219 (9/10) 
  & 152 &\wskip 
  & - 
  & - &\wskip 
  & - &\wskip 
  & - 
  & - 
  \\
mbw-b55-t7-s2 &\wskip
  & 206 
  & - 
  & - &\wskip 
  & - 
  & - &\wskip 
  & - &\wskip 
  & - 
  & - 
  \\
mbw-b60-t1-s0 &\wskip
  & - 
  & - 
  & - &\wskip 
  & - 
  & - &\wskip 
  & - &\wskip 
  & - 
  & - 
  \\
mbw-b60-t7-s1 &\wskip
  & 184 
  & - 
  & 206 &\wskip 
  & - 
  & - &\wskip 
  & - &\wskip 
  & - 
  & - 
  \\
mbw-b60-t7-s2 &\wskip
  & - 
  & - 
  & - &\wskip 
  & - 
  & - &\wskip 
  & - &\wskip 
  & - 
  & - 
  \\
\bottomrule
        \end{tabular}
      \end{scriptsize}
    \end{center}
    \caption{
    Results for both ASNets and baseline planners on Matching Blocksworld.
    Although the domain is from the IPC 2008 learning track, the instances were custom-generated.
    Each instance name shows the number of blocks (e.g. \texttt{b40}), the number of towers in the initial state and goal (e.g. \texttt{t3}), and a number distinguishing different random seeds for the instance generator (e.g. \texttt{s0}).
  }
  \label{tab:det-res-mbw}
  \end{table}

  \begin{table}
    \begin{center}
      \begin{scriptsize}
        \begin{tabular}{@{}lccccccccccccc@{}}
        \toprule
        \multirow{2}{*}{Problem}\wskip&
          & \multicolumn{3}{c}{ASNet} \wskip&
          & \multicolumn{2}{c}{A*}\wskip&
          & \multicolumn{1}{c}{GBF}\wskip&
          & \multicolumn{2}{c}{LAMA}
          \\
          \cmidrule{3-5} \cmidrule{7-8} \cmidrule{10-10} \cmidrule{12-13}
          &\wskip & - & PE & No h.
          &\wskip & LM-cut & LM-count
          &\wskip & LM-cut
          &\wskip & -2011 & -first
          \\
          \midrule
gm-7x7-s70 &\wskip
  & 43 
  & 110.4 (10/10) 
  & - &\wskip 
  & - 
  & - &\wskip 
  & 98 &\wskip 
  & 45 
  & 98 
  \\
gm-7x7-s71 &\wskip
  & 29 
  & 29.5 (10/10) 
  & - &\wskip 
  & 28 
  & 28 &\wskip 
  & - &\wskip 
  & 28 
  & 100 
  \\
gm-7x7-s72 &\wskip
  & 29 
  & 28.7 (10/10) 
  & - &\wskip 
  & 28 
  & 28 &\wskip 
  & - &\wskip 
  & 28 
  & 155 
  \\
gm-8x8-s80 &\wskip
  & 42 
  & 45.8 (10/10) 
  & - &\wskip 
  & - 
  & - &\wskip 
  & - &\wskip 
  & 248 
  & 248 
  \\
gm-8x8-s81 &\wskip
  & 36 
  & 36.1 (10/10) 
  & - &\wskip 
  & - 
  & - &\wskip 
  & - &\wskip 
  & - 
  & - 
  \\
gm-8x8-s82 &\wskip
  & 40 
  & 40.2 (10/10) 
  & - &\wskip 
  & - 
  & - &\wskip 
  & - &\wskip 
  & 40 
  & 51 
  \\
gm-9x9-s90 &\wskip
  & 42 
  & 45.7 (10/10) 
  & - &\wskip 
  & - 
  & - &\wskip 
  & - &\wskip 
  & 42 
  & 101 
  \\
gm-9x9-s91 &\wskip
  & 39 
  & 38.3 (10/10) 
  & - &\wskip 
  & - 
  & - &\wskip 
  & - &\wskip 
  & 207 
  & 207 
  \\
gm-9x9-s92 &\wskip
  & 36 
  & 36.6 (8/10) 
  & - &\wskip 
  & 36 
  & - &\wskip 
  & - &\wskip 
  & 36 
  & 118 
  \\
gm-10x10-s100 &\wskip
  & 43 
  & 43.2 (10/10) 
  & - &\wskip 
  & - 
  & - &\wskip 
  & - &\wskip 
  & 259 
  & 259 
  \\
gm-10x10-s101 &\wskip
  & - 
  & 74.2 (4/10) 
  & - &\wskip 
  & - 
  & - &\wskip 
  & - &\wskip 
  & - 
  & - 
  \\
gm-10x10-s102 &\wskip
  & 38 
  & 38.6 (10/10) 
  & - &\wskip 
  & - 
  & - &\wskip 
  & - &\wskip 
  & 235 
  & 235 
  \\
gm-13x13-s130 &\wskip
  & 77 
  & 83 (2/10) 
  & - &\wskip 
  & - 
  & - &\wskip 
  & - &\wskip 
  & - 
  & - 
  \\
gm-13x13-s131 &\wskip
  & - 
  & 104.4 (8/10) 
  & - &\wskip 
  & - 
  & - &\wskip 
  & - &\wskip 
  & - 
  & - 
  \\
gm-13x13-s132 &\wskip
  & 64 
  & 65.1 (10/10) 
  & - &\wskip 
  & - 
  & - &\wskip 
  & - &\wskip 
  & - 
  & - 
  \\
gm-16x16-s160 &\wskip
  & 71 
  & 70.8 (10/10) 
  & - &\wskip 
  & - 
  & - &\wskip 
  & - &\wskip 
  & - 
  & - 
  \\
gm-16x16-s161 &\wskip
  & 83 
  & 85.6 (7/10) 
  & - &\wskip 
  & - 
  & - &\wskip 
  & - &\wskip 
  & - 
  & - 
  \\
gm-16x16-s162 &\wskip
  & 75 
  & 76.3 (10/10) 
  & - &\wskip 
  & - 
  & - &\wskip 
  & - &\wskip 
  & - 
  & - 
  \\
gm-19x19-s190 &\wskip
  & 89 
  & 90.2 (8/10) 
  & - &\wskip 
  & - 
  & - &\wskip 
  & - &\wskip 
  & - 
  & - 
  \\
gm-19x19-s191 &\wskip
  & 80 
  & 80.4 (10/10) 
  & - &\wskip 
  & - 
  & - &\wskip 
  & - &\wskip 
  & - 
  & - 
  \\
gm-19x19-s192 &\wskip
  & 103 
  & 111 (5/10) 
  & - &\wskip 
  & - 
  & - &\wskip 
  & - &\wskip 
  & - 
  & - 
  \\
\bottomrule
        \end{tabular}
      \end{scriptsize}
    \end{center}
    \caption{
    Results for both ASNet and the baseline planners on our suite of Gold Miner problems.
    Each problem was produced by supplying a unique random seed to the same generator used in the IPC 2008 Learning Track.
    The size of each grid (i.e.\@ width $\times$ height) is shown in each problem name.
  }
  \label{tab:det-res-gm}
  \end{table}


\newpage
\section{Domain Descriptions}\label{app:domain-descriptions}

This appendix contains complete verbal descriptions of each of the domains that we use for evaluation.
We have also included full (P)PDDL for those domains which do not appear in previous work or are otherwise helpful for understanding our experiments.

\subsection{Probabilistic Domains}

\paragraph{Probabilistic Blocksworld}
A simple probabilistic adaptation of the venerable Blocksworld domain.
Each problem in this domain includes $n$ blocks that can be stacked on top of one another to form towers.
The agent is equipped with a gripper that can lift up blocks and deposit them either on the table, or on top of some other block.
The aim is to move from the initial configuration of block towers to some specific goal configuration.
In the probabilistic setting, this task is complicated by the possibility of gripper failure: each time the gripper goes to pick up a block, or place a block on top of another block, there is a 25\% chance that the block held by the gripper will instead fall onto the table.
Because there are no dead ends, this domain can be solved by taking the actions recommended by a classical planner for a determinisation of the problem, and re-planning whenever a gripper action fails~\cite{little2007probabilistic}.
This domain is based on domains that have appeared in past rounds of the International Probabilistic Planning Competition~\cite{younes2004ppddl1}.
Our version differs from some past versions in that it lacks actions for moving \textit{towers} of blocks; instead, only a single block at a time can be moved, just as in deterministic Blocksworld.
All instances for this task (as well as Exploding and Deterministic Blocksworld) were generated by the algorithm from \citeA{slaney2001blocks}.

\paragraph{Exploding Blocksworld~\cite{younes2004ppddl1}}
A more challenging variant of deterministic Blocksworld which includes both avoidable and unavoidable dead ends.
Unlike Probabilistic Blocksworld, there is no chance of the gripper dropping a block onto the table unless commanded to do so.
Instead, the challenge comes from dealing with two additional attributes given to each block: namely, whether the block has been \textit{destroyed}, and whether the block has been \textit{detonated} (both of which are initially false).
Once a block has been destroyed, it cannot be picked up by the gripper anymore and is thus stuck in its position; once a block has been detonated, it \textit{can} still be moved and picked up by the gripper, but it cannot detonate again.
Whenever a block $b_1$ is placed on top of a block $b_2$, there is a 10\% chance that $b_1$ will detonate and destroy $b_2$, thereby preventing $b_2$ from being moved again.
When placing a block $b_1$ on the table, there is a 40\% chance that $b_1$ will detonate and destroy \textit{the table}, in which case no further blocks can be placed directly onto the table.
Because being unable to place additional blocks on the table can prevent the goal from being reached, optimal policies for exploding Blocksworld often involve intentionally detonating blocks by placing them on top of other blocks that are already in their goal positions.
This renders the detonated block inert and allows it to be placed on the table at no risk.
Our version of this domain first appeared in IPPC 2008~\cite{bryce2008ipc}, although we have modified it to remove a bug which allowed blocks to be stacked on top of themselves.

\paragraph{Triangle Tireworld~\cite{little2007probabilistic}}
This domain was described in the main text of the article.
A PPDDL description of the domain is given in \cref{fig:ttw-ppddl}.

\begin{figure}[ht]
  \begin{mdframed}
    \begin{small}
    \input{./code-samples/ttw.tex}
    \end{small}
  \end{mdframed}
\caption{PPDDL domain for Triangle Tireworld.}\label{fig:ttw-ppddl}
\end{figure}

\paragraph{CosaNostra Pizza~\cite{toyer2018action}}
This domain was described in the main text of the article.
A PPDDL description of the domain is given in \cref{fig:cn-ppddl}.

\begin{figure}[ht]
  \begin{mdframed}
    \begin{small}
    \input{./code-samples/cn.tex}
    \end{small}
  \end{mdframed}
\caption{PPDDL domain for CosaNostra Pizza.}\label{fig:cn-ppddl}
\end{figure}

\subsection{Deterministic Domains}

\paragraph{Deterministic Blocksworld}
Our deterministic Blocksworld variant is identical to the Probabilistic Blocksworld domain described in \cref{ssec:expts-prob-doms}, but with a 0\% probability that the gripper will drop a block.

\paragraph{Gold Miner~\cite{fern2011first}}
Each problem in this domain consists of an $n \times n$ grid in which some locations are filled with soft rock and some locations are filled with hard rock, and one location contains some gold (which the agent wishes to retrieve).
There are also laser cannons and bombs in some cells that the agent can pick up in order to destroy the rock.
A laser cannon can be used repeatedly and is able to destroy both hard and soft rock in a cell adjacent to the agent, but will also destroy any gold in that adjacent cell.
A bomb can only be used once, and can only destroy soft rock in an adjacent cell, but will never destroy gold.
The instances of this domain are generated in such a way that there is a simple generalised policy that solves all instance of this domain.
Specifically, the agent can get a laser cannon, blast connected paths between the gold location and a bomb location, then pick up the bomb and use it to remove the last piece of rock needed to get to the goal.
This domain appeared for the first time in the learning track of the 2008 International Planning Competition~\cite{fern2011first}.

\paragraph{Matching Blocksworld~\cite{fern2011first}}
%
Another variant of the Blocksworld domain in which there are two grippers with opposite polarities.
Each block is also assigned a polarity matching one of the grippers.
Either gripper can be used to move either type of block.
However, if a gripper of one polarity is used to move a block of a different polarity, then the block will be ``damaged'' so that no other block can be placed on top of it, although the damaged block itself can still be picked up and moved.
A generalised policy must be able to solve arbitrary Blocksworld instances \textit{and} avoid picking up blocks with mismatched grippers.
Like Gold Miner, this domain first appeared in the learning track of the 2008 International Planning Competition.

\section{Receptive Field Experiments}\label{app:expts-recept}

\begin{table}[ht]
  \begin{center}
  \begin{tabular}{@{}cccccc@{}}
    \toprule
    \multirow{2}{*}{\makecell{Proposition\\layers}} & \multicolumn{5}{c}{Chain length $K$}\\
    \cmidrule{2-6}
    & 1 & 2 & 3 & 4 & 5\\
    \midrule
    1 & 30/30 & 14/30 & 14/30 & 14/30 & 14/30\\
    2 & 30/30 & 30/30 & 14/30 & 14/30 & 14/30\\
    3 & 30/30 & 30/30 & 30/30 & 14/30 & 14/30\\
    4 & 30/30 & 30/30 & 30/30 & 30/30 & 14/30\\
    \bottomrule
  \end{tabular}
  \end{center}
  \caption{
    Coverage (expressed as a fraction of 30 runs) for ASNets of different depths on the receptive-field-limited class of \texttt{unreliable-robot} problems.
  }
  \label{tab:recept-field}
\end{table}

We experimentally verified the receptive field limitation mentioned in \cref{ssec:heur-inputs} by training ASNets of different depths on a collection of \texttt{unreliable-robot} problems with two location chains of varying length $K$.
Specifically, we used problems with $K=1,\ldots, 5$ locations to train an ASNet, and then tested the ASNet on those same problems.
Our training strategy was almost the same as in \cref{ssec:expts-main}.
The main difference is that for the deeper, harder-to-train policies (e.g. four proposition layers/five action layers), we disabled regularisers to make the network converge more quickly.
We also disabled skip connections, although they should not affect the results for this particular domain.
\cref{tab:recept-field} shows the results of our experiment.
As expected, an ASNet with $L \geq K$ proposition layers will succeed at a length-$K$ problem, but not at problems for which $K < L$.

\section{A Sparse Policy for CosaNostra Pizza}\label{app:interp-cn}

\begin{figure}[ht!]
\centering
\begin{footnotesize}
  \begin{tabularx}{0.95\linewidth}{r@{\thinspace}c@{\thinspace}l@{\thinspace}c@{\thinspace}l}
    \toprule
    \multicolumn{5}{X}{\textit{First action layer}.\vspace{4pt}}\\
    \qquad $\hidact{\mactn{leave-toll-booth}{?from,?to}}{1}$ & $=$ & $\hidact{\mactn{ltb}{?f,?t}}{1}$ & $=$ & $f(-1.31 \cdot \text{action-count}(\mactn{leave-toll-booth}{?from,?to})$\\
    & & & & \quad $+\: 0.28 \cdot \text{is-goal}(\mactn{deliverator-at}{?to}) + 3.22)$\\
    \qquad $\hidact{\mactn{pay-operator}{?loc}}{1}$ & $=$ & $\hidact{\mactn{po}{?l}}{1}$ & $=$ & $f(-1.10 \cdot \text{action-count}(\mactn{pay-operator}{?loc})$\\
    & & & & \quad $-\: 0.61 \cdot \mpropn{open}{?loc} + 1.29)$\\
    \midrule
    \multicolumn{5}{X}{\textit{First proposition layer}.\vspace{4pt}}\\
    \qquad $\hidprop{\mpropn{deliverator-at}{?loc}}{1}$ & $=$ & $\hidprop{\mpropn{da}{?l}}{1}$ & $=$ & $f(0.40 \cdot \pool(\hidact{\mactn{leave-toll-booth}{?loc,\cdot}}{1})$\\
    & & & & \quad $-\: 1.46 \cdot \pool(\hidact{\mactn{leave-toll-booth}{\cdot,?loc}}{1}$\\
    & & & & \quad $-\: 1.48 \cdot \pool(\hidact{\mactn{pay-operator}{?loc}}{1}) + 4.98)$\\
    \midrule
    \multicolumn{5}{X}{\textit{Second action layer}: omitted, all weights zero.\vspace{4pt}}\\
    \midrule
    \multicolumn{5}{X}{\textit{Second proposition layer}: merely ``scales up'' the first proposition layer using a skip connection.\vspace{4pt}}\\
    \qquad $\hidprop{\mpropn{deliverator-at}{?loc}}{2}$ & $=$ & $\hidprop{\mpropn{da}{?l}}{2}$ & $=$ & $f(3.60 \cdot \hidprop{\mpropn{deliverator-at}{?loc}}{1})$\\
    \midrule
    \multicolumn{5}{X}{\textit{Third action layer}.\vspace{4pt}}\\
    \qquad $\hidact{\mactn{leave-open-intersection}{?from,?to}}{3}$ & $=$ & $\hidact{\mactn{loi}{?f,?t}}{3}$ & $=$ & $0$\\
    \qquad $\hidact{\mactn{leave-toll-booth}{?from,?to}}{3}$ & $=$ & $\hidact{\mactn{ltb}{?f,?t}}{3}$ & $=$ & $1.25 \cdot \hidprop{\mpropn{deliverator-at}{?from}}{2} - 1.07 \cdot \hidprop{\mpropn{deliverator-at}{?to}}{2}$\\
    \qquad $\hidact{\mactn{load-pizza}{?loc}}{3}$ & $=$ & $\hidact{\mactn{lp}{?l}}{3}$ & $=$ & $-0.65 \cdot \hidprop{\mpropn{deliverator-at}{?loc}}{2} + 4.71$\\
    \qquad $\hidact{\mactn{pay-operator}{?loc}}{3}$ & $=$ & $\hidact{\mactn{po}{?l}}{3}$ & $=$ & $4.99$\\
    \qquad $\hidact{\mactn{unload-pizza}{?loc}}{3}$ & $=$ & $\hidact{\mactn{up}{?l}}{3}$ & $=$ & $0.66 \cdot \hidprop{\mpropn{deliverator-at}{?loc}}{2} - 4.74$\\
    \bottomrule
  \end{tabularx}
\end{footnotesize}
\caption{
  An easily-readable representation of a sparse ASNet trained for the CosaNostra Pizza domain.
  Weights have been rounded to two decimal places, and $f(\cdot)$ has been used to denote the ELU activation function.
  The middle column shows abbreviations used to refer to activations when discussing their semantics in the main text (e.g.\@ $\psi_{\mathrm{da}(\varname{?l})}^{1}$).
}
\label{fig:cn-sparse-eqns}
\end{figure}

\begin{figure}[ht!]
%
%
    \centering
    \begin{footnotesize}
    \begin{tikzpicture}
        \node[anchor=south west,inner sep=0] (image) at (0,0) {\includegraphics[width=6.85cm]{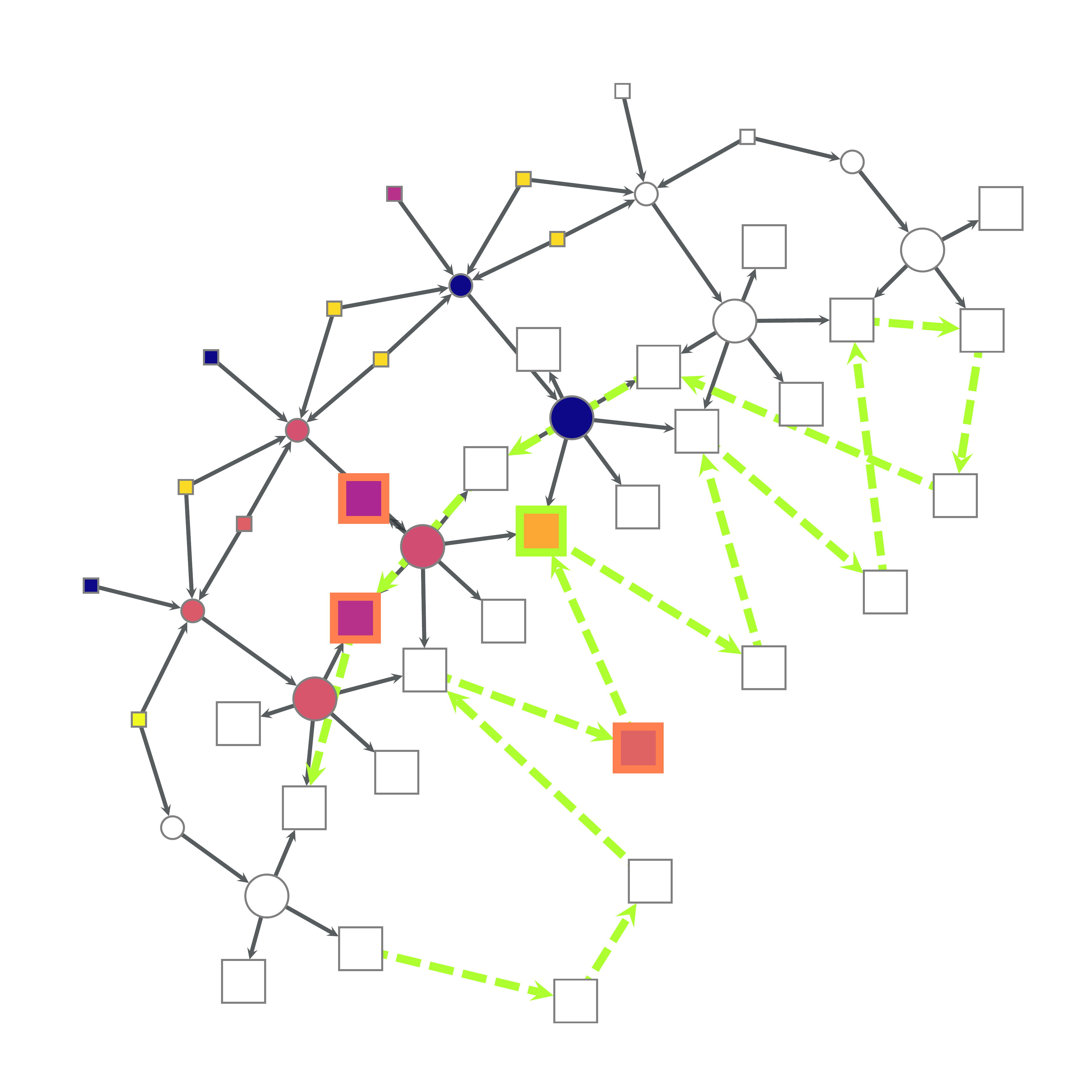}};
        %
        \draw [thick,dotted,->] (1.8,5.8) node [above,align=center] {$\hidact{\mactn{up}{b_1}}{3}$} to (2.25,3.95);
        \draw [thick,dotted,->] (0.9,5) node [above,align=center] {$\hidact{\mactn{ltb}{b_1,b_0}}{3}$} to (2,3.2);
        \draw [thick,dotted,->] (5.1,2.8) node [below right,align=center] {$\hidact{\mactn{ltb}{b_1,b_2}}{3}$} to (3.6,3.5);
        \draw [thick,dotted,->] (5,1) node [below right,align=center] {$\hidact{\mactn{po}{b_1}}{3}$} to (4.2,2.0);
        %
        \node [right] at (-0.1,6.8) {$\mactn{leave-toll-booth}{b_1,b_2}$ (step 6/16)};
        %
    \end{tikzpicture}
    \begin{tikzpicture}
        \node[anchor=south west,inner sep=0] (image) at (0,0) {\includegraphics[width=6.85cm]{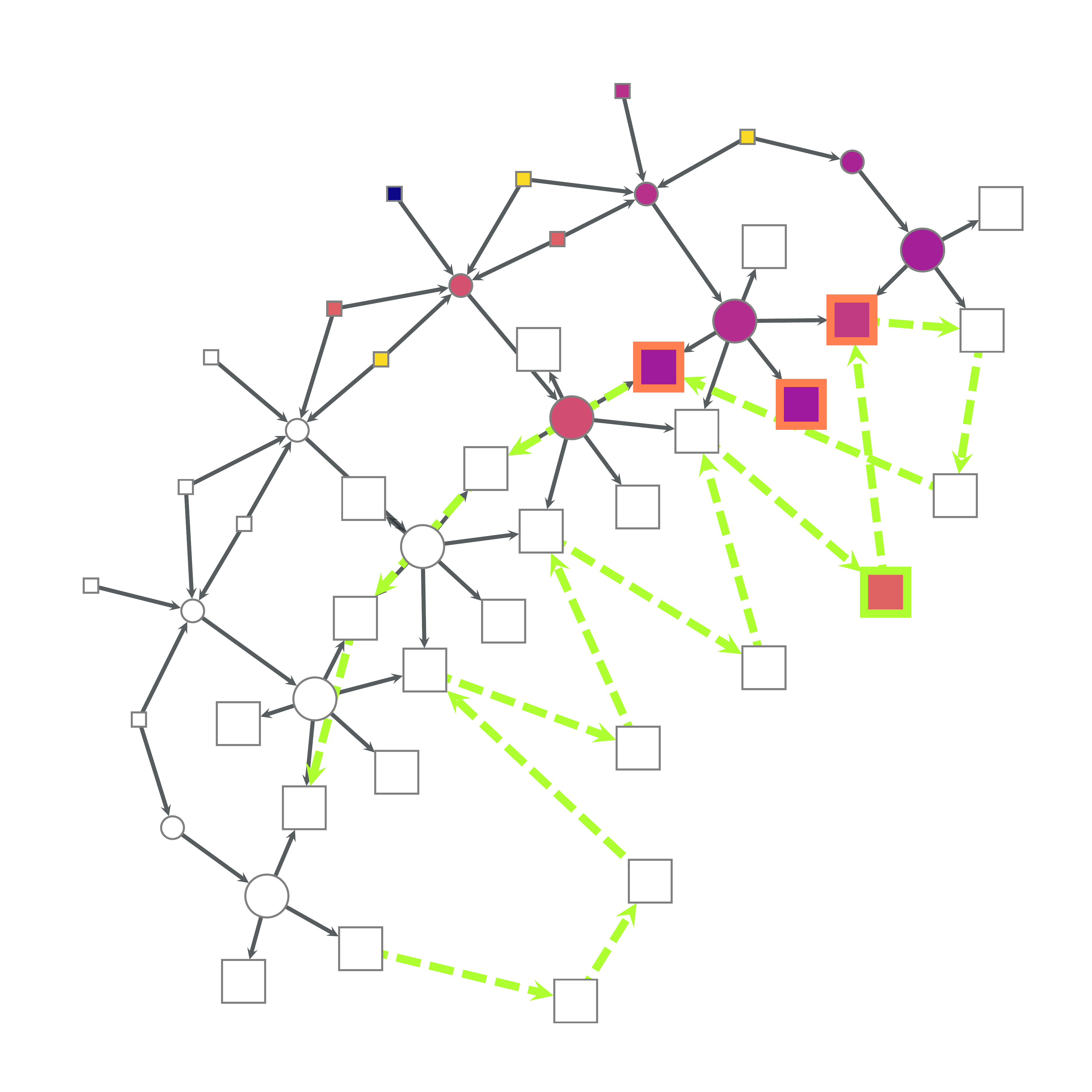}};
        %
        \draw [thick,dotted,->] (6.1,1.5) node [below,align=center] {$\hidact{\mactn{po}{b_3}}{3}$} to (5.65,2.95);
        \draw [thick,dotted,->] (5.2,2.3) node [below,align=center] {$\hidact{\mactn{up}{b_3}}{3}$} to (5.05,4.1);
        \draw [thick,dotted,->] (5.7,6) node [above,align=center] {$\hidact{\mactn{ltb}{b_3,home}}{3}$} to (5.35,5.05);
        \draw [thick,dotted,->] (2,5.7) node [above,align=center] {$\hidact{\mactn{ltb}{b_3,b_2}}{3}$} to (3.95,4.65);
        %
        \node [right] at (-0.1,6.8) {$\mactn{pay-operator}{b_3}$ (step 9/16)};
        %
    \end{tikzpicture}\\[-1.5em]
    \begin{tikzpicture}
      \draw[thin,dotted] (0,0) --(14,0);
    \end{tikzpicture}\\[0.5em]
    \begin{tikzpicture}
        \node[anchor=south west,inner sep=0] (image) at (0,0) {\includegraphics[width=6.85cm]{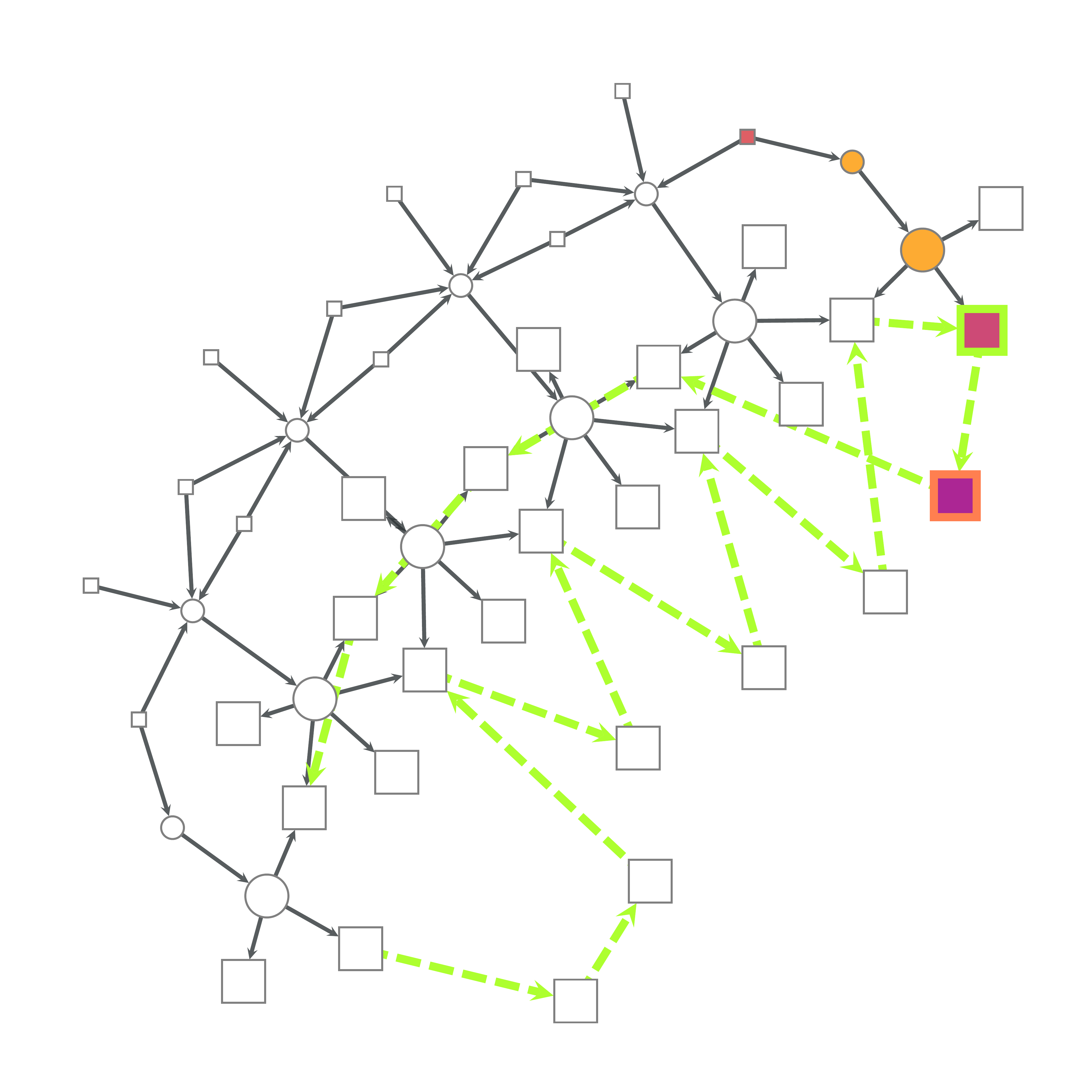}};
        %
        \draw [thick,dotted,->] (6.1,1) node [below,align=center] {$\hidact{\mactn{up}{home}}{3}$} to [out=75,in=295] (6.3,4.6);
        \draw [thick,dotted,->] (5.1,2) node [below,align=center] {$\hidact{\mactn{loi}{home,b_3}}{3}$} to [out=30,in=260] (6,3.55);
        %
        \node [right] at (-0.1,6.8) {$\mactn{unload-pizza}{home}$ (step 11/16)};
        %
    \end{tikzpicture}
    \begin{tikzpicture}
        \node[anchor=south west,inner sep=0] (image) at (0,0) {\includegraphics[width=6.85cm]{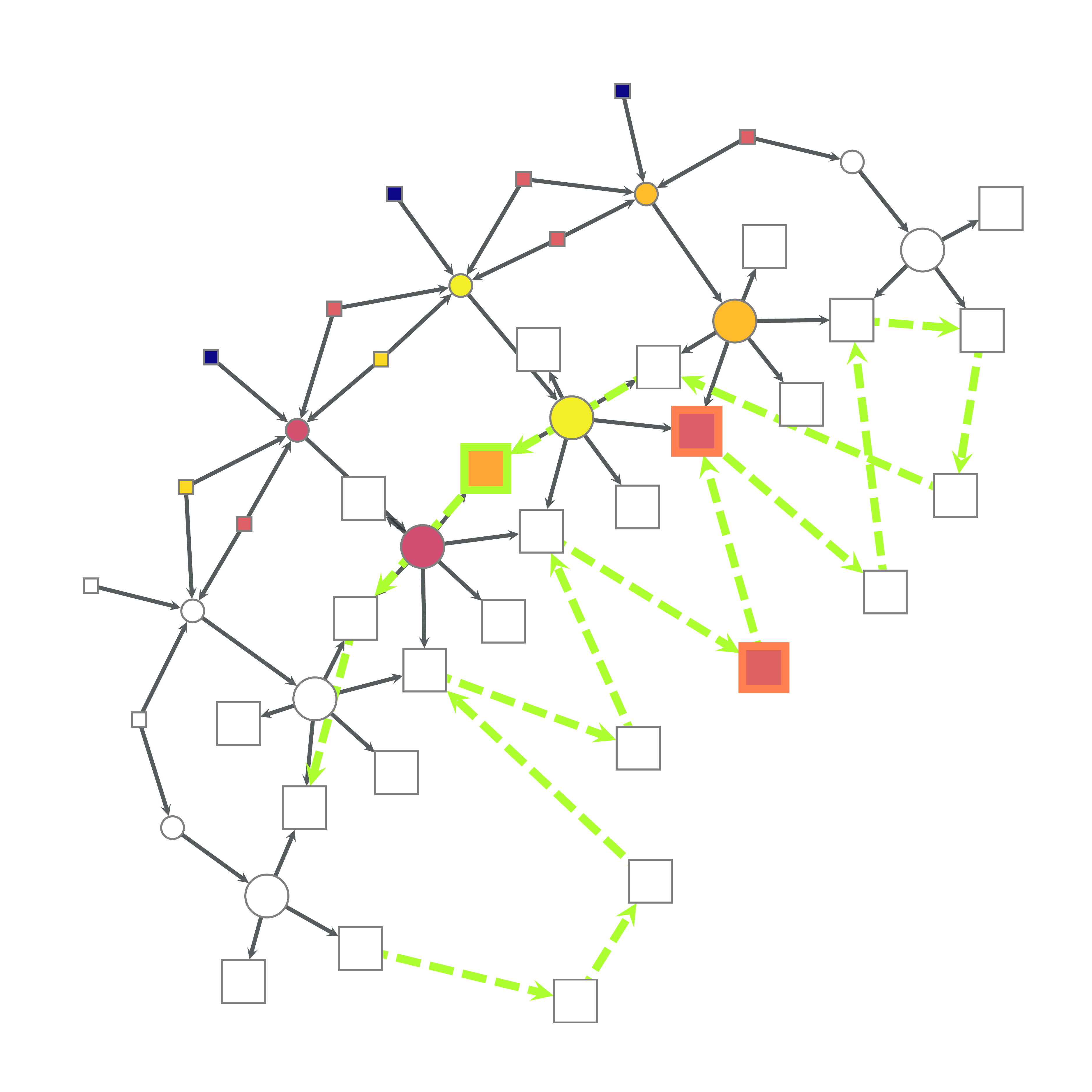}};
        %
        \draw [thick,dotted,->] (1.5,5.75) node [above,align=center] {$\hidact{\mactn{ltb}{b_2,b_1}}{3}$} to (2.9,4.1);
        \draw [thick,dotted,->] (4.8,1.25) node [below,align=center] {$\hidact{\mactn{po}{b_2}}{3}$} to (4.8,2.4);
        \draw [thick,dotted,->] (6,2) node [below,align=center] {$\hidact{\mactn{ltb}{b_2,b_3}}{3}$} to (4.6,3.9);
        %
        \node [right] at (-0.1,6.8) {$\mactn{leave-toll-booth}{b_2,b_1}$ (step 14/16)};
        %
    \end{tikzpicture}
    \includegraphics[width=0.8\linewidth]{figures/sparse-legend-horiz.pdf}
    \end{footnotesize}
    \vspace{-0.5em}
    \caption{
      Activations for the sparse CosaNostra Pizza policy given in \cref{fig:cn-sparse-eqns}.
      On the top row, the illustrated behaviours include moving in the correct direction when travelling from the shop to the customer, and paying tolls.
      The bottom row depicts unloading the pizza at the customer's home, and moving back from the customer to the shop.
      Each diagram includes labels for the final-layer modules which correspond to enabled actions.
    }\label{fig:cn-sparse-activations}
\end{figure}

Recall that in \cref{sec:understand}, we examined a sparse ASNet policy for Triangle Tireworld that was sufficiently compact to be written out as a small series of human-readable equations.
We repeated the same experiment for the CosaNostra Pizza domain introduced in \cref{ssec:expts-prob-doms}.
CosaNostra Pizza is a logistics problem in which a vehicle must pick up a pizza from its starting location (a pizza shop), travel though a chain of toll booths $b_0, b_1, \ldots, b_{K-1}$ leading from a shop to a customer, deliver a pizza to the customer, then return to the pizza shop.
This task is complicated by the behaviour of the toll booth operators: if the agent chooses not to pay at one of the toll booths on the way to the customer, then the toll collector may crush their car on the way back with 50\% probability.
The optimal policy is to pay all operators on the way to the customer, but not on the way back.
This policy reaches the goal with probability 1.

Our sparse network for CosaNostra Pizza is illustrated in \cref{fig:cn-sparse-eqns} (equations) and \cref{fig:cn-sparse-activations} (activations).
We obtained this network by training ten networks with an $\ell_1$ regularisation coefficient of 0.02 for one hour each, and then using the sparsest network that could successfully solve the test problems.
CosaNostra is a more complex domain than Triangle Tireworld, and a successful policy must solve several distinct sub-tasks: it must be able to pick up the pizza, alternate between paying toll booth operators and moving between booths, drop off the pizza, and travel back to the shop.
To keep our treatment concise, we will use \cref{fig:cn-sparse-eqns} to show that the ASNet behaves successfully at only one of these sub-tasks: choosing the correct direction to travel when moving the vehicle from the shop to the customer.
We invite the reader to try a similar approach to verify other desirable properties of the network (for instance: why does it not pick up pizza again immediately after dropping it off?).

Consider the actions chosen by an ASNet while it is moving the vehicle from the shop to the customer.
Specifically, say that it is at a booth $b_k$ in a chain $b_0, b_1, \ldots, b_{K-1}$, where $0 < k < K-1$ (i.e.\@ it is not at the first or last booth).
After paying the toll at booth $b_k$, the ASNet faces four choices: it can pay the toll booth operator again; it can unload the pizza; or it can take one of two movement actions, one of which leads to the customer and one of which leads back to the shop.
We will focus on the last two options: when does the ASNet choose the action that moves forward to booth $b_{k+1}$ over the action that moves back to booth $b_{k-1}$?
Formally, $\mactn{leave-toll-booth}{b_k,b_{k+1}}$ will be chosen over $\mactn{leave-toll-booth}{b_k,b_{k-1}}$ when the corresponding final-layer activations satisfy $\hidact{\mactn{ltb}{b_{k},b_{k+1}}}{3} > \hidact{\mactn{ltb}{b_{k},b_{k-1}}}{3}$ (recall our shorthand of using, e.g., $\mobjlist{ltb}$ in place of $\mobjlist{leave-toll-booth}$).
We can simplify this condition by substituting in a series of definitions from \cref{fig:cn-sparse-eqns}:
\begingroup
\allowdisplaybreaks
\begin{align*}
  \hidact{\mactn{ltb}{b_{k},b_{k+1}}}{3} &> \hidact{\mactn{ltb}{b_k,b_{k-1}}}{3} &\\[0.5em]
  \begin{aligned}
    &1.25 \cdot \hidprop{\mpropn{da}{b_k}}{2}\\
    &\quad -\: 1.07 \cdot \hidprop{\mpropn{da}{b_{k+1}}}{2}
  \end{aligned} &> \begin{aligned}
    &1.25 \cdot \hidprop{\mpropn{da}{b_k}}{2}\\
    &\quad -\: 1.07 \cdot \hidprop{\mpropn{da}{b_{k-1}}}{2}
  \end{aligned} & \text{(expand definition)}\\[0.5em]
  -1.07 \cdot \hidprop{\mpropn{da}{b_{k+1}}}{2} &> -1.07 \cdot \hidprop{\mpropn{da}{b_{k-1}}}{2} & \text{(simplify)}\\[0.5em]
  \hidprop{\mpropn{da}{b_{k+1}}}{2} &< \hidprop{\mpropn{da}{b_{k-1}}}{2} & \text{(flip sign)}\\[0.5em]
%
%
  \hidprop{\mpropn{da}{b_{k+1}}}{1} &< \hidprop{\mpropn{da}{b_{k-1}}}{1} & \text{(expand, remove $f(\cdot)$)}\\[0.5em]
  \begin{aligned}
    &0.40 \cdot \pool(\hidact{\mactn{ltb}{b_{k+1},\cdot}}{1}) \\
    &\quad -\: 1.46 \cdot \pool(\hidact{\mactn{ltb}{\cdot,b_{k+1}}}{1} \\
    &\quad -\: 1.48 \cdot \pool(\hidact{\mactn{po}{b_{k+1}}}{1})
  \end{aligned} &< \begin{aligned}
    &0.40 \cdot \pool(\hidact{\mactn{ltb}{b_{k-1},\cdot}}{1}) \\
    &\quad -\: 1.46 \cdot \pool(\hidact{\mactn{ltb}{\cdot,b_{k-1}}}{1} \\
    &\quad -\: 1.48 \cdot \pool(\hidact{\mactn{po}{b_{k-1}}}{1})
  \end{aligned} & \text{(expand, remove $f(\cdot)$)}
\end{align*}
\endgroup
Crucially we have used the fact that the ELU activation function, given by $f(x) = x$ when $x \geq 0$ and $f(x) = \exp(x) - 1$ when $x \leq 0$, is strictly increasing.
Consequently, we have $f(a) > f(b)$ iff $a > b$.

To help simplify the last inequality, recall from \cref{fig:cn-sparse-eqns} that the $\maschema{leave-toll-booth}$ modules in the first action layer are defined as
\[
  \hidact{\mactn{leave-toll-booth}{?from,?to}}{1} = f(-1.31 \cdot \text{action-count}(\mactn{leave-toll-booth}{?from,?to})~.
\]
This will take the value $f(-1.31) < 0$ if $\mactn{leave-toll-booth}{?from,?to}$ has ever been executed, and $f(0) = 0$ otherwise.
Observe that all of the max-pooling operations over $\hidact{\mactn{ltb}{?f,?t}}{1}$ modules will be pooling over a module for at least one $\maschema{leave-toll-booth}$ action that has not already been executed.
For instance, $\pool(\hidact{\mactn{ltb}{\cdot,b_{k-1}}}{1})$ pools over $\hidact{\mactn{ltb}{b_k,b_{k-1}}}{1}$, but $\mactn{leave-toll-booth}{b_k,b_{k-1}}$ will not have been executed yet (under normal execution of an otherwise-optimal policy).
Hence, we can replace those $\pool$ operations with zero values, and only consider the pooling operations over modules for $\maschema{pay-operator}$ ($\maschema{po}$) actions, giving us:
\begin{align*}
    -1.48 \cdot \pool(\hidact{\mactn{po}{b_{k+1}}}{1})
  &<
    -1.48 \cdot \pool(\hidact{\mactn{po}{b_{k-1}}}{1})\\
    \hidact{\mactn{po}{b_{k+1}}}{1}
  &>
    \hidact{\mactn{po}{b_{k-1}}}{1}\\
  \begin{aligned}
    &-1.10 \cdot \text{action-count}(\mactn{pay-operator}{b_{k+1}}) \\
    &\quad-\: 0.61 \cdot \mpropn{open}{b_{k+1}}
  \end{aligned}
  &>
  \begin{aligned}
    &-1.10 \cdot \text{action-count}(\mactn{pay-operator}{b_{k-1}}) \\
    &\quad-\: 0.61 \cdot \mpropn{open}{b_{k-1}}
  \end{aligned}\\
\end{align*}
In the second step we flipped the sign and then dropped the $\pool$ operator, which we are justified in doing because each toll booth has exactly one related $\maschema{pay-operator}$ action.
In the third step we substituted in the definition of $\hidact{\mactn{po}{?l}}{1}$, then dropped the activation and bias, since $u \mapsto f(u + c)$ is an increasing function of $u$ for any constant bias $c$.

In our scenario, the agent is travelling from the shop to the customer, and is between two toll booths $b_{k-1}$ and $b_{k+1}$.
At this point it will have already paid the toll operator at $b_{k-1}$, and so booth $b_{k-1}$ will be open and the action count for the corresponding action will be one.
It has not visited the next booth $b_{k+1}$, so $\mpropn{open}{b_{k+1}}$ will be false and the corresponding action count will be zero.
Substituting these values in the last inequality, we obtain
\[
    -1.10 \cdot 0 - 0.61 \cdot 0 = 0 > -1.10 \cdot 1 - 0.61 \cdot 1 = -2.81~,
\]
which is of course true.
This proves that the agent will correctly choose $\mactn{leave-toll-booth}{b_k, b_{k+1}}$ over $\mactn{leave-toll-booth}{b_k, b_{k-1}}$ when at booth $b_k$ somewhere in a chain $b_1, b_2, \ldots b_{K-1}$ (for $0 < k < K - 1$).

\vskip 0.2in
\bibliography{citations,fwt}
\bibliographystyle{theapa}

\end{document}